\documentclass[default,iicol]{sn-jnl}

\jyear{2022}

\theoremstyle{thmstyleone}

\theoremstyle{thmstyletwo}

\theoremstyle{thmstylethree}

\usepackage{subfig}
\graphicspath{{bib/figures/}}

\usepackage{float}

\usepackage{xcolor}
\usepackage{comment}

\usepackage{url}

\usepackage{amsmath, amssymb}
\usepackage{graphicx} 
\usepackage{relsize}

\usepackage{array}
\usepackage[flushleft]{threeparttable}
\usepackage{multirow}
\usepackage{bbm}

\usepackage{amssymb}
\usepackage{mathtools}

\DeclarePairedDelimiter\floor{\lfloor}{\rfloor}

\DeclareMathOperator*{\argmin}{argmin}
\usepackage{subfiles}

\begin{document}

\title{A Meta-level Analysis of Online Anomaly Detectors}

\author[1]{\fnm{Antonios} \sur{Ntroumpogiannis}}\email{droubo@csd.uoc.gr}

\author[2]{\fnm{Michail} \sur{Giannoulis}}\email{michail.giannoulis@uca.fr}

\author[1,3]{\fnm{Nikolaos} \sur{Myrtakis}}\email{myrtakis@csd.uoc.gr}

\author[3]{\fnm{Vassilis} \sur{Christophides}}\email{vassilis.christophides@ensea.fr}

\author[4]{\fnm{Eric} \sur{Simon}}\email{eric.simon@sap.com}

\author[1]{\fnm{Ioannis} \sur{Tsamardinos}}\email{tsamard.it@gmail.com}


\affil[1]{\orgname{Univ. of Crete}, \orgaddress{\street{Voutes Campus}, \city{Heraklion}, \postcode{70013}, \country{Greece}}}

\affil[2]{\orgname{Univ. Clermont Auvergne}, \orgaddress{\street{49 Bd François Mitterrand}, \city{Clermont-Ferrand}, \postcode{63000}, \country{France}}}

\affil[3]{\orgname{ENSEA, ETIS}, \orgaddress{\street{6 Av. du Ponceau}, \city{Cergy}, \postcode{95000}, \country{France}}}

\affil[4]{\orgname{SAP}, \orgaddress{\street{35 Rue d'Alsace}, \city{Levallois-Perret}, \postcode{92300}, \country{France}}}




\abstract{
Real-time detection of anomalies in streaming data is receiving increasing attention as it allows us to raise alerts, predict faults, and detect intrusions or threats across industries. Yet, little attention has been given to compare the effectiveness and efficiency of anomaly detectors for streaming data (i.e., of online algorithms). In this paper, we present a qualitative, synthetic overview of major online detectors from different algorithmic families (i.e., distance, density, tree or projection-based) and highlight their main ideas for constructing, updating and testing detection models. Then, we provide a thorough analysis of the results of a quantitative experimental evaluation of online detection algorithms along with their offline counterparts. The behavior of the detectors is correlated with the characteristics of different datasets (i.e., meta-features), thereby providing a meta-level analysis of their performance. Our study addresses several missing insights from the literature such as (a) how reliable are detectors against a random classifier and what dataset characteristics make them perform randomly; (b) to what extent online detectors approximate the performance of offline counterparts; (c) which sketch strategy and update primitives of detectors are best to detect anomalies visible only within a feature subspace of a dataset; (d) what are the tradeoffs between the effectiveness and the efficiency of detectors belonging to different algorithmic families; (e) which specific characteristics of datasets yield an online algorithm to outperform all others.
}


\maketitle

\section{Introduction}
\label{sec:introduction}
Abnormal data might be more interesting to study than the prevalent patterns \cite{Zimek2018,Aggarwal2013,Chandola2009}. \emph{Data anomalies} e.g., among measurements or observations may simply represent errors (called \emph{outliers}), but they can also indicate interesting phenomena (called \emph{novelties}), such as new incidents or faults of a system, intrusions to a computer network, frauds in credit cards transactions or even over-expressed genes of living things\footnote{In this paper, we use the terms outlier, novelty and anomaly detection interchangeably}. 

Recently, the \emph{real-time detection of anomalies} in streaming data has gained increasing attention \cite{Bailis2017,Sadik2014} as it allows to raise alerts, predict faults, detect intrusions and threats across industries. However, analyzing a sequence of samples\footnote{We call "sample" an element (i.e., observation, measurement) of a data stream.} arriving over time imposes unique constraints and challenges for machine learning models. In contrast to batch detectors, the full dataset is not available in advance and online detectors must learn incrementally as samples arrive \cite{Carbone2020}.In that respect, the order rather than the timestamp of the samples in a data stream is an important feature that models should take into account \cite{windowing_1,windowing_2}. Additional constraints on the detection setting may be imposed in practice. For instance, the high velocity of streams leaves little opportunity for labeling samples by experts. Also, a thorough tuning of several hyper-parameters \footnote{An hyper-parameter cannot be estimated from the data.} is challenging especially when data characteristics evolve over time \cite{best_offline_4}. For these reasons, in this paper we focus on unsupervised methods for detecting point anomalies and leave out of the scope of our study (semi-)supervised methods for shallow \cite{blazquez2021} or deep anomaly detection \cite{Pang2021}. In this respect, detection of range-based (i.e., subsequence or collective) anomalies based on temporal dependencies rather than independent abnormal samples \cite{Cook2020,Braei2020,Tatbul2018,Gupta2014,Chandola2012} is left as future work.

Existing unsupervised methods for detecting anomalies in multivariate datasets bypass the need of labeled samples by exploiting different \emph{anomalousness} criteria \cite{Wang2019,Zimek2018,Aggarwal2013,Chandola2009,high_dim_problem,Hodge2004} based on the divergence of statistical distributions, distance thresholds, density variance from nearest neighbors, isolation facility, etc. Indeed, recent online detection methods belong to different algorithmic families: (i) \emph{proximity-based} detectors including distance-threshold based like MCOD \cite{mcod} and CPOD \cite{cpod} or nearest-neighbor-based like LEAP \cite{leap}, inspired by KNN$_W$ \cite{knn} (ii) \emph{density-based} detectors either on the full feature space like STARE \cite{stare} and their offline counterpart LOF \cite{lof} or on feature subspaces like RS-Hash \cite{rshash} tracing its roots back to HICS \cite{hics}; (iii) \emph{tree-based} detectors such as HST \cite{hst} and RRCF \cite{rrcf} inspired by IF \cite{iforest} or OCRF \cite{ocrf}; and (iv) \emph{projection-based} detectors such as LODA \cite{loda} or XSTREAM \cite{xstream} featuring both online and offline versions.

Anomaly detection has been an active area of research over the past decades. Besides notorious surveys on anomaly detection approaches and methods  \cite{Hodge2004,Chandola2009,Chandola2012,Gupta2014,Aggarwal2013,Zimek2018,Wang2019,Cook2020}, various empirical studies have experimentally evaluated the effectiveness or the efficiency of detectors \cite{offline_benchmark_1,best_offline_2,best_offline_3,offline_benchmark_2,offline_benchmark_3,offline_benchmark_4,offline_benchmark_5,Choudhary2017}. However, the aforementioned works focus in their majority, on offline detectors. Online anomaly detection was listed in the future perspectives of the overview presented in \cite{Chandola2012} while \cite{Gupta2014} includes only the first steps in making  density-based detectors incremental, like LOF \cite{lof}. Regarding empirical studies, \cite{best_distance_based} focuses exclusively on the efficiency of online proximity-based detectors while \cite{Choudhary2017} compares stream clustering algorithms \cite{Silva2013} for anomaly detection. To the best of our knowledge, no previous work has compared qualitatively or quantitatively, over the same multi-dimensional datasets, distance-based (MCOD, CPOD), KNN-based (LEAP, KNN$_W$) and density-based detectors (STARE, RS-Hash, LOF) with tree-based (HST/F, RRCF, IF, OCRF) and projection-based detectors (XSTREAM, LODA). Additionally, reported experiments seldom report the tension between the effectiveness and the efficiency of the detection algorithms. To this end, we optimally tune the hyper-parameters of detectors per dataset rather than rely on the default configurations recommended by their inventors. Last, previous meta-learning analyses of detectors \cite{metaod,Vanschoren2019} do not consider meta-features related to anomalies visible only to a subset of the dataset feature space. 

In this context, several questions that are left unanswered regarding the performance of online anomaly detectors over data streams. First, previous studies do not assess the reliability of detectors' effectiveness against a random classifier, and do not either highlight the dataset characteristics (e.g., number of features) that make them perform randomly. Second, they do not indicate when online detectors can approximate the effectiveness of offline detectors and under which conditions (e.g., number of features irrelevant to the anomalies, anomaly ratio). Third, they do not indicate which is the best sketch strategy and update primitives of detectors (e.g., micro-clusters, random trees, histogram or chain-based density estimators) to detect anomalies that are only visible within a feature subspace of a dataset. Fourth, they do not analyze the tradeoffs between the effectiveness and the efficiency of detectors belonging to different algorithmic families. Last, they do not highlight the characteristics of datasets that make an online algorithm capable of outperforming all others. For instance, the statistically significant correlations between the relative performance of the best performing detectors and meta-features such as the number of samples or features in a dataset, the skewness of feature values, or the distance between the clusters of abnormal and normal samples. In summary, we make the following contributions:
\begin{itemize}
    \item {\bf Large Selection of Online Detectors from Different Algorithmic Families:} In Section~\ref{chapter:anomaly_detection_algorithms}, we introduce the \it{nine} online anomaly detectors included in our testbed, namely, MCOD \cite{mcod}, CPOD \cite{cpod}, LEAP \cite{leap}, STARE \cite{stare}, RS-Hash \cite{rshash}, HST \cite{hst}, RRCF \cite{rrcf}, LODA \cite{loda} and XSTREAM \cite{xstream} (for their offline counterparts, readers are referred to Appendix~\ref{app:offline}). We detail their scoring function, model creation and update primitives, as well as the involved analytical complexities. To ensure a common ground of comparison, we implemented a variation of HST with a forgetting mechanism similar to RRCF, called HSTF, as well as, a continuous scoring function for MCOD, CPOD and LEAP instead of their binary outcome.
    
    \item {\bf Fair Evaluation Environment for Online and Offline Anomaly Detectors over Multivariate Data:} In Section~\ref{chapter:benchmark_environment}, we describe the characteristics of abnormal and normal samples (e.g., anomaly ratio, dimensionality) in the \it{twenty-four} real and the \it{five} synthetic datasets included in our testbed that are widely used in previous empirical studies \cite{Wang2019, best_offline_1,best_offline_2,best_offline_3,best_distance_based}. We additionally consider in Section \ref{chapter:range-based} the recently proposed benchmark Exathlon \cite{Jacob2021} for explainable anomaly detection over repeated executions of two different Spark streaming applications, containing five different type of anomalies. To fairly compare the performance of detectors, we consider as \emph{evaluation metrics both Area Under the ROC Curve (AUC) and Average Precision (AP)} and explain their differences under edge cases. These metrics are computed for each algorithm under \emph{optimal evaluation conditions per dataset} (for optimal hyper-parameter values per dataset and sensitivity of algorithms to tuning, readers are referred to Appendix~\ref{app:hyperparameters}). 
    
    \item {\bf Thorough Evaluation of Detectors' Effectiveness:} In Section~\ref{chapter:effectivness}, we analyse the AUC and Mean AP of detectors over the 24 real datasets of our testbed (details are given in Appendix~\ref{sec:aucmapscores}) in order to reveal interesting patterns involving specific meta-features (i.e., Number of Samples/Features, Anomaly to Normal distance). In particular, we \emph{assess the reliability of the decisions} made by the detectors w.r.t. a random classifier, and \emph{rank in a statistically significant way} online and offline detectors according to their performance. Our analysis sheds light on \emph{how well online detectors approximate the performance of offline detectors}.
    \item {\bf Robustness of Detectors Against Increasing Dimensionality:}  In Section~\ref{chapter:robustness}, we assess the robustness of online and offline detectors against increasing data and anomaly subspace dimensionality using \it{twenty} synthetic datasets. Specifically, we investigate \emph{whether a particular algorithmic family (e.g., proximity, tree or projection-based) is able to discover anomalies that are only visible in a small subset of features (i.e., a subspace)}.
    \item {\bf Efficiency of Detectors and Trade-offs:} In Section~\ref{chapter:efficiency}, we report the execution time of training and updating detectors' models over the 24 real datasets of our testbed. We investigate the \emph{trade-off between the update time and the effectiveness of the 9 online detectors contrasted to the best overall performing detector}.
    \item {\bf Meta-learning Analysis of Leading Detectors:} In Section~\ref{chapter:metalearning}, we investigate which of the \emph{meta-features are statistically correlated with the relative effectiveness of the best performing detector} (detailed results per meta-feature are given in Appendix~\ref{app:meta}). This meta-analysis helps us to assess whether the best overall performing detector will also excel in a given dataset.
\end{itemize}

Finally, conclusions and future work are discussed in Section \ref{chapter:conclusion_future_work}. Detailed experimental results, as well as all employed hyper-parameter values are given in Appendices \ref{app:hyperparameters}, \ref{sec:aucmapscores} and \ref{app:meta}.

\section{Detection Algorithms} \label{chapter:anomaly_detection_algorithms}
In this section, we introduce the online anomaly detection algorithms considered in our experimental evaluation that belong to the distance, knn, tree or projection-based algorithmic families. The offline counterpart of these detectors is described in Appendix~\ref{app:offline}.

\begin{figure}
    \centering
    \subfloat[Sliding windows of size 7 and of slide 5.]{
    \includegraphics[width=0.5\linewidth]{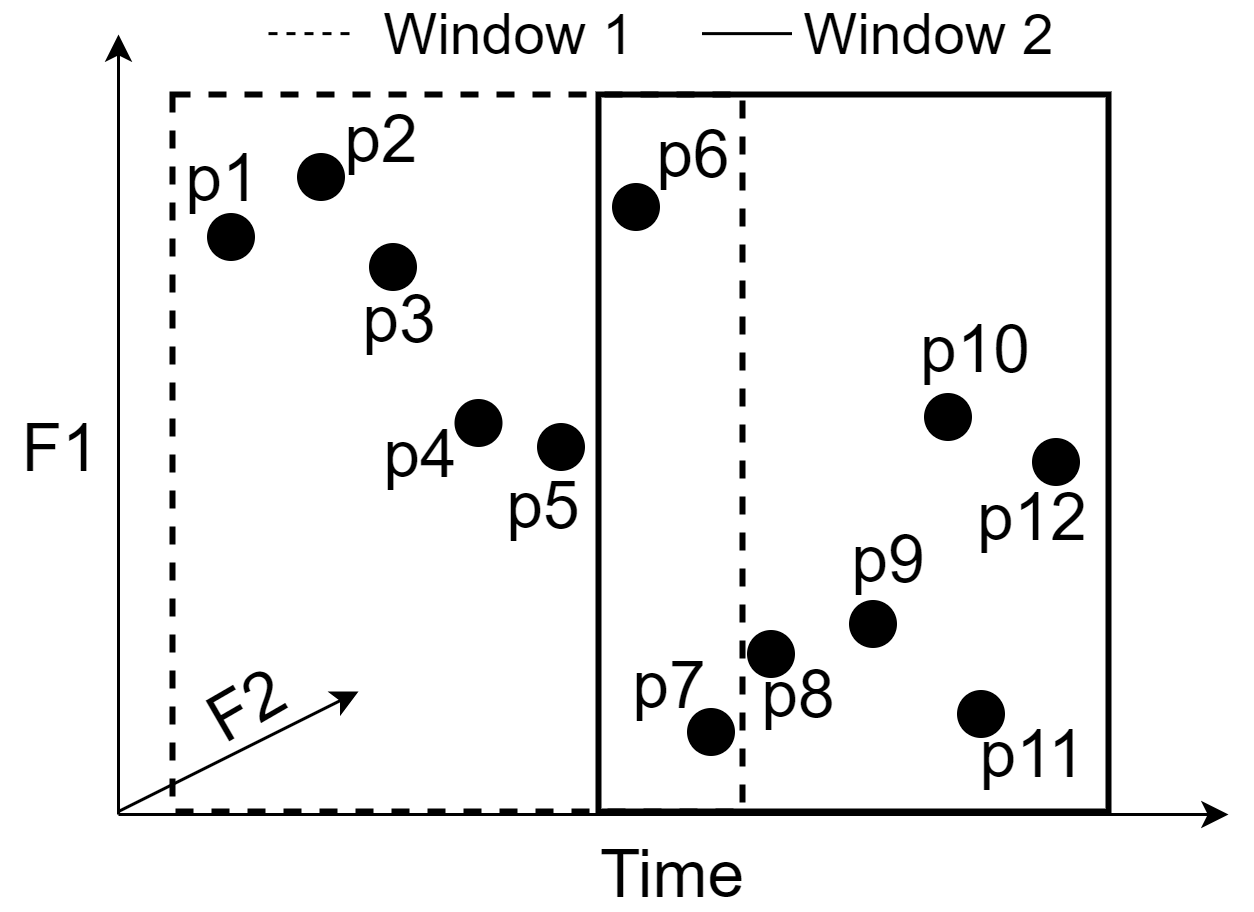}
    \label{fig:slidingWindowsEx}
    }~~
    \subfloat[Tumbling windows of size 6.]{\includegraphics[width=0.5\linewidth]{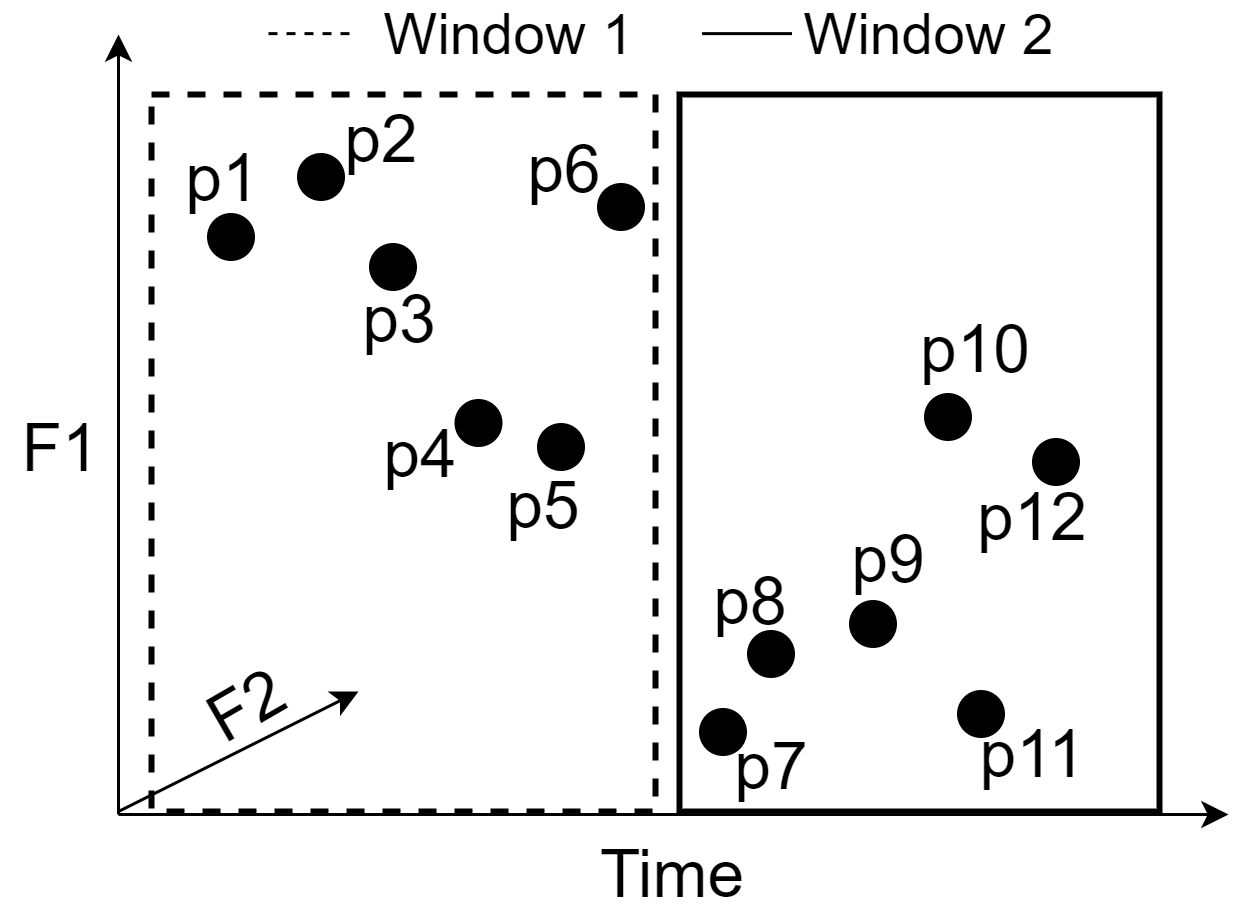}
    \label{fig:tumblingWindowEx}
    }
    \caption{Example of two sliding (left) and tumbling (right) windows in the time-varying feature space.}
    \label{fig:winEx}
\end{figure}

In contrast to offline detectors that model and score samples in one batch, online detectors continuously update their models for incrementally detecting anomalies in several windows of samples. The total number of samples in a window indicates the \textit{window size} while the number of samples that a window will be shifted over a data stream indicates the \textit{window slide}. Windows that overlap as they slide over a data stream are called \textit{sliding}, otherwise they are called \textit{tumbling} (i.e., window size = window slide), as illustrated in \ref{fig:winEx}.

Our testbed includes state-of-the-art online anomaly detection algorithms with publicly available implementations, as reported in the literature. CPOD \cite{cpod} was included because it outperforms other distance-based detectors (such as MCOD\cite{mcod} which is considered to be among the best detectors) in terms of runtime and memory usage\footnote{The recently proposed distance-based detector NETS \cite{nets} consumes less resources than MCOD but is not reporting any improvement in terms of effectiveness.}. HST \cite{hst} exhibits the best efficiency among tree-based detectors while RRCF\cite{rrcf} is the most effective detector of the same family. We additionally included two projection-based detectors: LODA\cite{loda} exhibiting a very low runtime and memory footprint and XSTREAM \cite{xstream} outperforming in terms of effectiveness other detectors on high-dimensional datasets. LEAP\cite{leap} has been proved to be three orders of magnitude faster than state-of-the-art KNN methods. STARE\cite{stare} outperforms other popular density-based detectors (\cite{kelos,na2018dilof}) in terms of execution time,  achieving comparable or higher accuracy. Finally, RS-HASH\cite{rshash} proved to be more efficient and effective than other subspace detectors like HiCS\cite{hics}. To the best of our knowledge, no previous study has compared both the effectiveness and efficiency of proximity-based, tree-based and projection-based detectors under stream and batch processing settings. In the next sub-sections, we provide the main operations and primitives of each detection algorithm, both theoretically and through a running example, and detail their train, update, forgetting and anomaly reporting procedures.

\subsection{Micro Cluster Outlier Detection (MCOD)} \label{online_detectors_mcod}
MCOD is a distance based detector \cite{mcod} that models neighboring regions of samples in a stream as micro-clusters (MC). MCOD requires to tune three hyper-parameters: a distance threshold $R$, a neighbor count threshold $K$ and a distance metric $\mathit{dist(\cdot, \cdot)}$ (e.g. Euclidean). Given a dataset $D$, MCOD identifies a sample $p$ to be a \textit{distance-based anomaly} if it has fewer than $K$ neighbors within distance $R$, otherwise $p$ is considered normal.

MCOD builds a set of micro-clusters (MC), to assess the normality of samples in every window. An MC is composed of at least $K$ + 1 samples and is centered on one sample. A sample belongs to at most one MC and the distance of any sample from the center of its MC is at most $R/2$. According to the  triangular inequality in the metric space, the  distance between every pair of samples in a MC is smaller than $R$. Therefore, every sample in a MC is considered normal. Samples that cannot be clustered, i.e., potential anomalies, are inserted in a list called \textit{PD}. The contents of \textit{PD} are processed as new samples in a window arrives and they can be either normal if $\mathit{R\textnormal{-Neigh}(p)} \geq K$ (see Eq. \ref{eq:rneighbors}) or abnormal, otherwise. The $\mathit{R\textnormal{-Neigh}(p)}$ indicates the number of neighbors of a sample $p$ in a radius $R$. Hence, we list the building blocks of MCOD: \\
\textbf{Training Phase.} MCOD does not have a training phase since its model operates over a single window by computing pair-wise distances.
\\
\textbf{Model Update.} A new sample $p$ can either: (i) be inserted into its nearest MC (see Eq. \ref{eq:nearestCluster}) if the distance from the center ($\mathit{mcc}$) of that MC is $\leq$ $R/2$; or (ii) form a new MC if it has at least $K$ neighbors in \textit{PD} list within a distance $\leq R/2$; or (iii) be inserted into \textit{PD}.
\\
\textbf{Forgetting Mechanism.} MCOD forgets all samples that have been processed in a current window before processing the next window. A forgotten sample can: (i) dissolve an MC if it contains less than $K$ + 1 points; or (ii) be removed from the \textit{PD} list.
\\
\textbf{Anomaly Report.} After processing new and forgotten samples, every sample with less than $K$ neighbors is reported as anomaly.
\vspace{-.3cm}
\begin{equation}
  \mathit{mcc_p^\star} = \argmin_{\mathit{mcc}} \mathit{dist(p, mcc)}.  
  \label{eq:nearestCluster}
\end{equation}
\vspace{-.5cm}
\begin{equation}
  \mathit{R\textnormal{-Neigh}(p)} = \vert\{p' \vert \mathit{dist(p, p') < R}\}\vert
  \label{eq:rneighbors}
\end{equation}

\begin{figure}
    \centering
    \subfloat[Sliding Window 1.]{
    \includegraphics[width=0.42\linewidth]{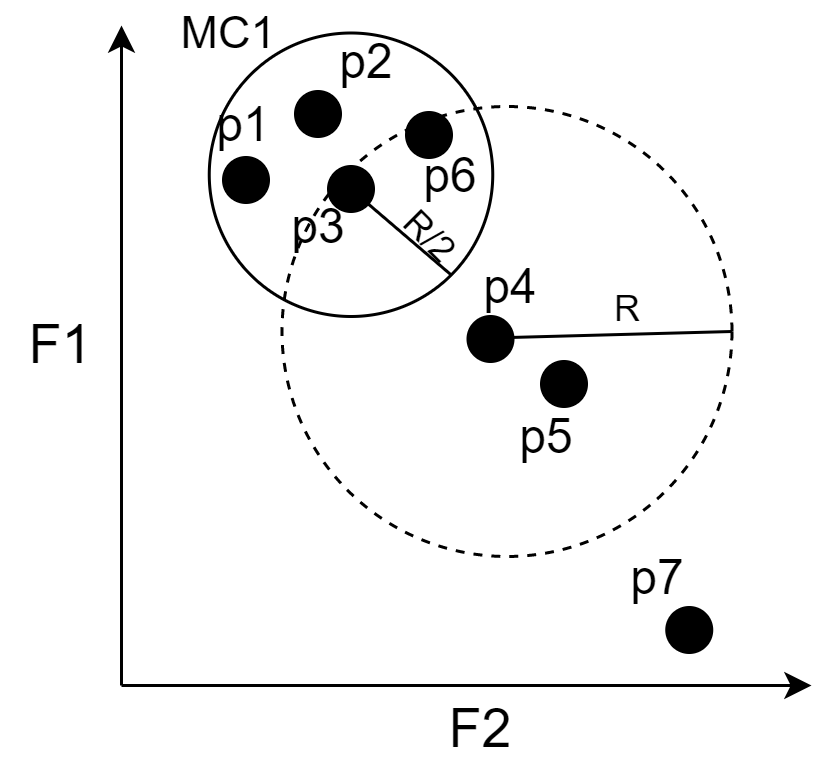}
    \label{fig:mcodSlidWin1fspace}
    }\qquad
    \subfloat[Sliding Window 2.]{\includegraphics[width=0.42\linewidth]{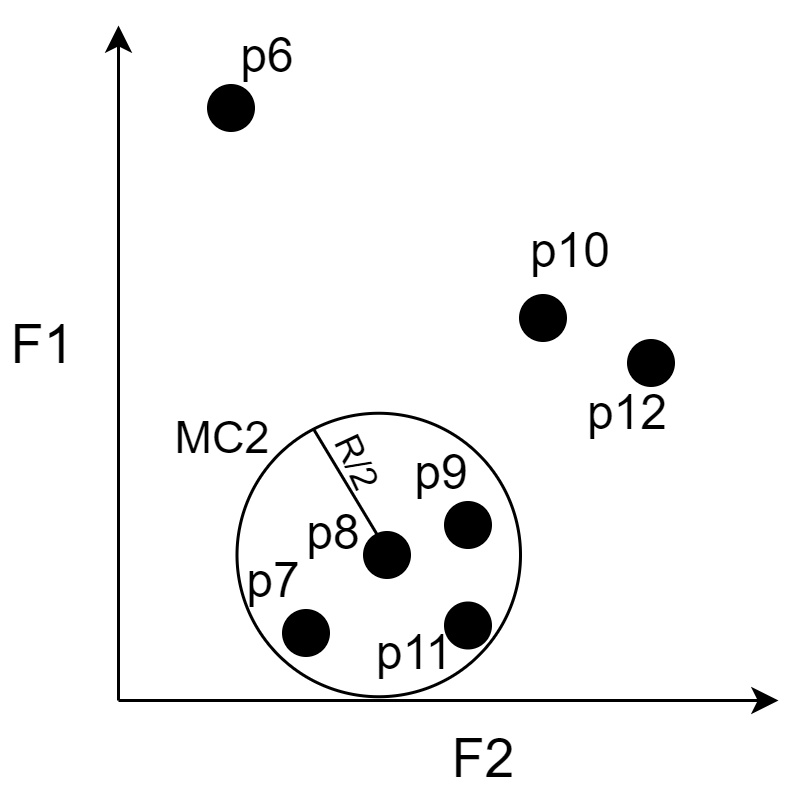}
    \label{fig:mcodSlidWin2fspace}
    }
    \caption{MCOD running on the sliding windows of Figure \ref{fig:slidingWindowsEx} in the feature space with neighbor count threshold $K=3$ and radius threshold $R$.}
    \label{fig:mcodSlidWindowsFspace}
\end{figure}

A running example inspired from the MCOD paper \cite{mcod} is presented in Figure \ref{fig:mcodSlidWindowsFspace} over the two sliding windows depicted in Figure \ref{fig:slidingWindowsEx}. Samples have two features, namely \{\textit{F1, F2}\}. Using the first window (Figure \ref{fig:mcodSlidWin1fspace}), MCOD builds a micro-cluster $\mathit{MC_1}$ containing the samples $p_1, p_2, p_3, p_6$, which are considered as normal. The \textit{PD} list contains the samples that do not belong to $\mathit{MC_1}$: $p_4, p_5, p_7$. Sample $p_4$ is  normal as it has $K=3$ neighbors within distance $R$. Samples $p_5, p_7$ are anomalies since they have less than $K=3$ neighbors within distance $R$. Next, all samples of the first window are forgotten, so $\mathit{MC_1}$ is dissolved and \textit{PD} is emptied. MCOD then processes the second window (Figure \ref{fig:mcodSlidWin2fspace}) and forms the micro-cluster $\mathit{MC_2}$. Observe that $p_7$ is now a normal sample, while $p_6$ is an anomaly because its preceding neighbors have been forgotten.

MCOD originally provides a binary label $l$ as outcome depicted in Eq. \ref{eq:mcodLabel} $\mathit{MCOD_l}$: 0 for normal and 1 for abnormal samples. However, to homogenize the  comparison of the outcome with other algorithms, we need a continuous scoring function of samples. We therefore use the function $\mathit{MCOD_s}$ that gives a score $s$ depicted in Eq. \ref{eq:mcodScore}. 
Intuitively, the more a sample lies in a sparse region, i.e., it has few or no neighbors within distance $R$, or it is far away from its nearest MC, the higher it should be scored. For instance, $p_7$ and $p_3$ in Figure \ref{fig:mcodSlidWin1fspace} will respectively get the highest and lowest score. Also, $p_5$ has a higher score than $p_4$ which has a higher score than $p_6$. 
\vspace{-.5cm}

\begin{equation}
    \thickmuskip=0mu
  \scalebox{0.9}{$\mathit{MCOD_l(p)}$} = \mathit{dist(p, mcc_p^\star)} < \frac{R}{2}  \vee \mathit{\scalebox{0.8}{R\textnormal{-Neigh}}(p)} \geq K
\label{eq:mcodLabel}
\end{equation}

\vspace{-.5cm}
\begin{equation}
  \mathit{MCOD_s(p) = \frac{1}{R\textnormal{-Neigh}(p)} \cdot dist(p, mcc_p^\star)}
  \label{eq:mcodScore}
\end{equation}

The main advantage of MCOD is that it effectively prunes pair-wise distance computations, providing a more efficient neighbor search through the creation of micro-clusters. Every sample that remains in an MC when the window slides is considered as normal without further checks. Thus, to classify a new sample it suffices to compute distances only w.r.t. the centers of MC and the samples in the \textit{PD} list. MCOD has linear time complexity on average $\Theta((1-c) w  log((1-c)  w)+K  log(K)))$ and linear space complexity $O(c  w+(1-c)  K  w)$ \cite{best_distance_based}, where $c$ is the number of clusters, $K$ is the number of nearest neighbors and $w$ is the window size. However, depending on the data characteristics, i.e., if the data are very sparse, no clusters may be formed resulting to $c=0$. Thus, for the worst case, MCOD has linearithmic time complexity $O(w \mathit{log} (w) + K\mathit{log}(K))$ The hyper-parameters of MCOD require careful tuning (see Figure \ref{fig:std_online_hyperparameters} in Appendix \ref{app:hyperparameters}). Specifically, a very low $R$ and high $K$ values on a dataset with several sparse areas may lead MCOD to create very few micro-clusters, identifying most samples as anomalies. The aforementioned behavior will also degrade the efficiency of MCOD as it must compute the distances with respect to more samples in the \textit{PD} list, which is inefficient for neighbor searches. In the opposite case, with a very high $R$ and low $K$ values in a dataset having several dense areas, MCOD may identify most samples as normal and create a large number of micro-clusters.

\subsection{Core Point-based Outlier Detection (CPOD)}
\label{sec:cpod}

CPOD is a distance-based detector \cite{cpod} that models neighboring regions of samples in a stream using core points. CPOD requires to tune two hyper-parameters: a distance threshold $R$ and a neighbor count threshold $K$. Given a dataset $D$, CPOD identifies a sample $p$ to be a \emph{distance-based} anomaly if it has fewer than $K$ neighbors within distance $R$, otherwise $p$ is considered normal.

CPOD shares the same definition of anomalies as MCOD, but it optimizes neighbor's search by looking at the close neighborhood of few special samples called \emph{core points} rather than the neighborhood of the nearest micro-cluster center. The same technique is also used to address a limitation of MCOD when every pair of points has a distance greater than $R/2$. In this degenerate case, no micro-cluster is formed resulting in quadratic neighbor search and a poor efficiency especially with streams. A core point is a special sample that supports multi-distance indexing as it stores its Euclidean distances to other samples in multiple ranges within each slide. It permits to both quickly identify normal samples and reduce neighbor search spaces for anomaly candidates. A core point has the following two properties: (i) the distance between any pair of core points is greater than $R$; (ii) each sample $p \in D$ is linked to at least one core point $c$, having distance less than $R$ from $c$. Note that in contrast to the micro-cluster centers of MCOD, core points do not require to have at least $K$ neighbors to be formed. Each core point $c$ stores every neighbor sample $p$ in a map $E$ for different radius values $k \in \{0,1,2,3\}$, ranging from $R/2$ to $2R$: \\
$ E_k(c) = \{p \in D \vert ~ kR/2 < \mathit{dist(c,p)} \leq (k+1)R/2\}.$
To calculate the neighbors of a sample $p$ within distance $R$, CPOD leverages the distance from its corresponding core point(s) to reduce the computations, matching exactly one of the four cases below, where $c^\star$ is the closest core from a core set $C$ to $p$:
\begin{enumerate}
    \item $\bigcup_{k = 0,1,2}E_k(c^\star)$, if $\mathit{dist(c^\star,p)} \leq R/2$
    \item $\bigcup_{k = 0,1,2,3}E_k(c^\star)$, if $R/2 < \mathit{dist(c^\star,p)} \leq R$
    \item $\bigcup_{k = 0,1}E_k(c_i \in C$, if $R < \mathit{dist(c^\star, p)} \leq 2R$
    \item $0$, if $2R < \mathit{dist(c_i, p)}, \forall c_i \in C$
\end{enumerate}

\noindent
From the aforementioned cases, exactly one can hold for a particular sample $p$. To better grasp the neighbor search procedure, we explain the first case, where $\mathit{dist(c^\star,p)} \leq R/2$. CPOD operates via a prefilter approach, called minimal probing. First, it automatically considers the samples within $R/2$ ($k=0$ in $E_k$) as neighbors of $p$ due to triangular inequality. If the neighbors in $R/2$ do not exceed the neighbor threshold, then $k$ is increased $k=1$ and the search is expanding to the range $(R/2, R]$. If the threshold is satisfied, the search stops and $p$ is declared as normal, otherwise the search continuous to higher ranges. The same logic is followed for the rest cases.
In the following, we report the building blocks of CPOD:
\\\textbf{Training Phase.} CPOD does not have a training phase since its model computes the core points for every new slide, starting from the first window.
\\\textbf{Model Update.} A new sample $p$ can either: (i) be linked to exactly one core, if $\mathit{dist(c^\star, p)} \leq R$; (ii) to multiple cores if $R < \mathit{dist(c_,p)} \leq 2R$; (iii) form a core point or (iv) to not correspond to any core point if $dist(c_i,p) > 2R, \forall c_i \in C$.
\\\textbf{Forgetting Mechanism.} with a new slide, CPOD forgets all the expired samples, i.e., samples that are just removed from the current window,
before processing the next window slide. The neighbor count of the active samples, i.e., samples that remain in the current window, is decreased and the expired samples are removed from $E$ maps of their cores.
\\\textbf{Anomaly Report.} CPOD spots samples as anomaly candidates if they have a distance $\leq R$ from their cores and less than $K$ neighbors, expanding the neighbor search in higher ranges. Every sample with less than $K$ neighbors after CPOD's expanded search is reported as anomaly.

Now, we present a running example of CPOD based on the Figure \ref{fig:mcodSlidWindowsFspace} we used to explain MCOD. For this example with $K=3$, we consider the cores $c_2, c_4$ to be core points in the first slide, having the same values as $p_2$ and $p_4$ respectively. Thus, the neighbors of each core in distance $[0, 2R]$ are $E(c_2): \{p_2, p_1, p_6, p_4, p_5\}$ and $E(c_4)$ contains all samples, where $E$ denotes all the neighbors in different ranges.
Every sample in the solid circle falls in the first case, presented previous neighbor-search procedure; it has distance $\leq R/2$ from $c_2$ and it will be immediately identified as normal having 3 neighbors. Sample $p_5$ also fall in the first case, there are not enough neighbors in distance $\leq R/2$, it is considered an anomaly candidate and the search will expand to higher radius ranges from its closest core $c_4$. Expanding the search to $R$, no new neighbors are found  so the search stops and $p_5$ is reported as anomaly. The sample $p_7$ falls in the third case, having distance $R < dist(c_4, p_7) \leq 2R$ from its closest core $c_4$, so CPOD searches neighbors in radius $R/2$ to $R$ in every core, labelling $p_7$ as anomaly since only $p_5$ found as its neighbor. In the next slide, considering only $c_{10}$ as core taking the values of $p_{10}$, CPOD labels $p_7$ as anomaly (unlike MCOD). This is because $R < \mathit{dist(c_{10}, p_7)} \leq 2R$, thus the neighbors search is performed in radius $<R$ of each core, resulting to samples $p_{12}, p_8, p_9$ as possible neighbors, missing $p_{11}$ where $dist(c_{10}, p_{11})>R$. From the explored samples, only $p_8, p_9$ are neighbors of $p_7$ so it is labelled as anomaly. Moreover, $p_6$ is also an anomaly since $\mathit{dis(c_{10}, p_6)}>2R$, thus it has no neighbors in radius $R$. The rest samples are labelled as normal.

CPOD originally provides a binary anomaly outcome. However, our evaluation metrics require an ordered outcome and thus we report the score of a sample $p$ as the inverse number of its neighbor count 
$\mathit{CPOD_s(p)} = 1 /(\mathit{\vert N_{CPOD}(p) \vert} + \epsilon)$. Note that lower scores denote greater anomalousness.

CPOD has linear time complexity $O(N_c~w + N_f~N_r)$, where $N_c$ is the total number of cores, $w$ is the number of samples in a window, $N_f$ is number of the anomaly candidates and $N_r$ the  neighbor search time for these candidates. Note that in a very sparse dataset with many isolated samples, i.e., many samples have distance $> R$ and $\leq 2R$ from every core, the neighbor search will use each core, searching on their $R/2$ and $R$ radius. This could result an overhead if many cores have been formed. Therefore, $R$ and $K$ should be carefully selected.

\subsection{Lifespan-Aware Probing (LEAP)}
\label{sec:leap}
LEAP is a distance-based detector \cite{leap} that encompasses two different anomaly semantics, namely \emph{distance-threshold}-based and \emph{nearest-neighbor}-based\footnote{nearest neighbors are distinguished according to maximum and average distance}. 
LEAP requires to tune two hyper-parameters: a distance threshold $R$ and a neighbor count threshold $K$. Given a dataset $D$, LEAP identifies a sample $p$ to be a \emph{distance-based} anomaly if it has fewer than $K$ neighbors within distance $R$, otherwise $p$ is considered normal. 

LEAP shares the same definition of anomalies as MCOD, but it mitigates expensive range queries with adequate indexing: rather than storing in the same index structure all samples of a window, a separate smaller index is maintained for each slide. 
For a given window, a slide index $s_i$ references a sample that appears in sliding window $i$ and not in sliding window $i+1$. 
For instance, for a window of size $w=9$ and slide $s=3$, three index structures are created because three sliding windows cover the samples of that window. LEAP maintains an evidence list $evi$ for every sample $p$:
$p.evi = \bigcup_{i=t,..t-s} \{q \in s_i \vert \mathit{dist(p, q)} \leq R\}
$, containing the neighbors of $p$ in each slide in reverse chronological order, starting from the newest slide $s_t$ up to $s_{t-s}$, where $s$ is the slide size, adopting the minimal probing principle, i.e., if more than $K$ neighbors are found till the current slide, the search stops. LEAP maintains also a trigger list:
$tr = \{q \in w \vert q.evi \setminus s_{t-s} \neq q.evi\}$, containing the samples that are going to lose neighbors in the next window slide. 
Subsequently, we report the building blocks of LEAP:
\noindent
\\\textbf{Training Phase.} LEAP does not have a training phase since it performs the neighbor search for every new slide, starting from the first window.
\\\textbf{Model Update.} A new sample $p$ leads to re-probing the neighbors of each sample in \emph{tr} list and potentially it can let a sample(s) to be removed from \emph{tr} list.
\\\textbf{Forgetting Mechanism.} when a slide expires: (i) its index is discarded; (ii) the expired neighbors of $p$ in \emph{p.evi} list are removed and (iii) the samples in the \emph{tr} list are re-evaluated.
\\\textbf{Anomaly Report.} For a sample $p$, the newest samples are examined first and in case the neighbors are fewer than $K$, the probing continuous. If $p$ has less than $K$ neighbor after each slide of a window is probed, $p$ is reported as anomaly.

Now, we present a running example of LEAP based on the Figure \ref{fig:mcodSlidWindowsFspace} we used to explain MCOD. For this example with $K = 3$, $w=7$ and slide $s=5$, LEAP will create two indexes $s_1$ and $s_2$, respectively indexing the samples $p_1,...p_5$ and samples $p_6, p_7$. LEAP will build an evidence list for each sample. For $p1$, it will start searching for neighbors in $s_2$ and then in $s_1$ resulting to $p_1.evi = \{p_6, p_2, p_3\}$. For the first window $p_7$ and $p_5$ are reported as anomalies and the rest samples as normal. The samples belonging to the trigger list in the first window are $tr = \{p_6, p_7\}$ as they will lose neighbors in the next slide. When the window slides, the samples in the trigger list will be evaluated first. Now, $p_6$ is labelled as anomaly while $p_7$ as normal considering its succeeding neighbors. for this window, each sample inside the solid circle is considered normal and the anomalies are the samples $p_6, p_{10}, p_{12}$. 

LEAP originally provides a binary anomaly outcome. However, our evaluation metrics require an ordered outcome and thus we report the score of a sample $p$ as the inverse number of its neighbor count in radius $R$: $\mathit{LEAP_s(p)} = 1/R\textnormal{-Neighbors}(p)$. Note that lower scores denote greater anomalousness.

In the worse case, i.e., when the minimal probing principle fails, LEAP requires quadratic time-complexity $O(w^2)$, where $w$ is the number of samples in a window. However, authors mention that more advanced data structures can be utilized to reduce the neighbor search complexity.

\subsection{Half Space Trees (HST)}
\label{sec:hst}

HST \cite{hst} is a \emph{tree-based} anomaly detector that learns a sketch of a data stream using an ensemble of half-space trees (HS-Tree). A HS-Tree is a full binary tree in which all leaves are at the same depth. HST requires to tune two hyper-parameters: the maximum depth $h$ of each HS-tree and the number of HS-trees $T$ of the ensemble.

Each half-space tree is built using a random perturbation of the original feature space, called \textit{workspace}, where an internal node of a tree represents  a selected feature. HST selects a feature \textit{randomly and uniformly}. Then, the \textit{splitting value} is randomly picked in the half-way of the \emph{work range} of a selected feature. The work range (wr) is defined in Eq. \ref{eq:hstWorkRangeOrig} and differs per feature $F$. Note that $v_F$ is a random value between the maximum and the minimum value of a feature, $v_F \in [F_{min}, F_{max}]$. Note that HST uses all the window samples to construct the trees in contrast to batch tree-based detectors such as IF \cite{iforest} and OCRF \cite{ocrf} that rely on bootstrapping to induce diversity among the constructed trees. HST ensures this diversification by selecting randomly the splitting value $v_F$. The $v_F$ value lets HST to construct wide value ranges, which is a crude estimate of the range of the unseen samples. We should also stress that HST assumes that the data is scaled such that values of features are bounded in [0, 1]. However, such normalization would have altered the nature of other tree-based algorithms such as RRCF \cite{rrcf} that utilizes the different value ranges to select the most prominent split feature at each step. Therefore, to ensure that the algorithms are compared on an equal basis, we modified the work range function (see Eq. \ref{eq:hstWorkRangeMod}) so that the data range does not have to be restricted. 
\vspace{-.3cm}
\begin{equation}
\label{eq:hstWorkRangeOrig}
    \mathit{wr(F)} = v_F \pm 2 \cdot \mathit{max}(v_F, 1-v_F)
\end{equation}
\vspace{-.5cm}
\begin{equation}
\label{eq:hstWorkRangeMod}
    \mathit{wr'(F)} = v_F \pm 2 \cdot \mathit{max}(v_F - F_{min}, F_{max}-v_F)
\end{equation}
HST assigns an anomaly score to each sample. The score of a sample $p$ in a specific tree $t \in \mathit{HS-Trees}$ is defined as:
\vspace{-.3cm}
\begin{equation}
\label{eq:hstTreeScore}
    \mathit{score}(p, t) = \mathit{Node.r} \cdot 2^\mathit{Node.h},
\end{equation}
where $\mathit{Node.r}$ is the mass and $\mathit{Node.h}$ is the height of a leaf node respectively in a tree $t$. The lower the score a sample obtains, the more anomalous it is considered. Then, HST assigns a total score to each sample $p$ which is the sum of scores obtained from the constituent trees $\mathit{HS-Trees}$:
\vspace{-.3cm}
\begin{equation}
\label{eq:hstTotalScore}
    \mathit{HST}(p) = \sum_{t \in \mathit{HS-Trees}} \mathit{score}(p, t).
\end{equation}
A significant limitation of HST is that as new samples are processed, the mass profiles of the corresponding leaf nodes can only increase; in other words, HST can never forget. To overcome this limitation, we have implemented a HST variation with a simple \emph{forgetting mechanism} inspired by RRCF \cite{rrcf}, noted as \textbf{HSTF}. Subsequently we list the building blocks of HST/F: \\
\textbf{Training Phase.} During training HST builds an ensemble of half-space trees that learns the sketch of the data stream. The mass profile of a leaf is the number of samples that end up to that subspace. The lower the mass, the more sparse the region is considered. The structure of the trees formed during training remain unaltered during the update phase and only the mass profiles are updated.
\\
\textbf{Model Update.} During updating, HST uses two alternating tumbling windows: a latest window which is not full yet and a reference window preceding the reference window. The mass profile of the reference window is computed and samples in latest window falling in low mass leaves (partitions) are considered anomalous. When the latest window is full, the mass profile of a leaf is updated by adding the number of samples that fell into that partition.
\\
\textbf{Forgetting Mechanism.} Our HSTF variation forgets the samples of the oldest windows by decreasing the mass profiles of the corresponding leaves. The number of samples that the model must remember is specified by a \emph{forgetting threshold} $f$ i.e., a third hyper-parameter w.r.t. the original HST. The lower the value of $f$, the faster a high mass profile may become sparse. As an old sample is deleted only after the insertion of a new sample, $f$ should be a multiplicative of the window size $w$. Therefore, when $f$ is exceeded, the mass profile of the $w$ oldest leaves (i.e., not been updated by the latest window) in each tree will be decreased by one. 
\\
\textbf{Anomaly Report.} As it relies on tumbling windows, HST/F reports the anomaly scores for each sample after processing entirely the latest window according to Eq. \ref{eq:hstTotalScore}.
\begin{figure}
    \centering
    \subfloat[Partitions (left) and model (right) on tumbling window 1.]{
    \includegraphics[width=0.82\linewidth]{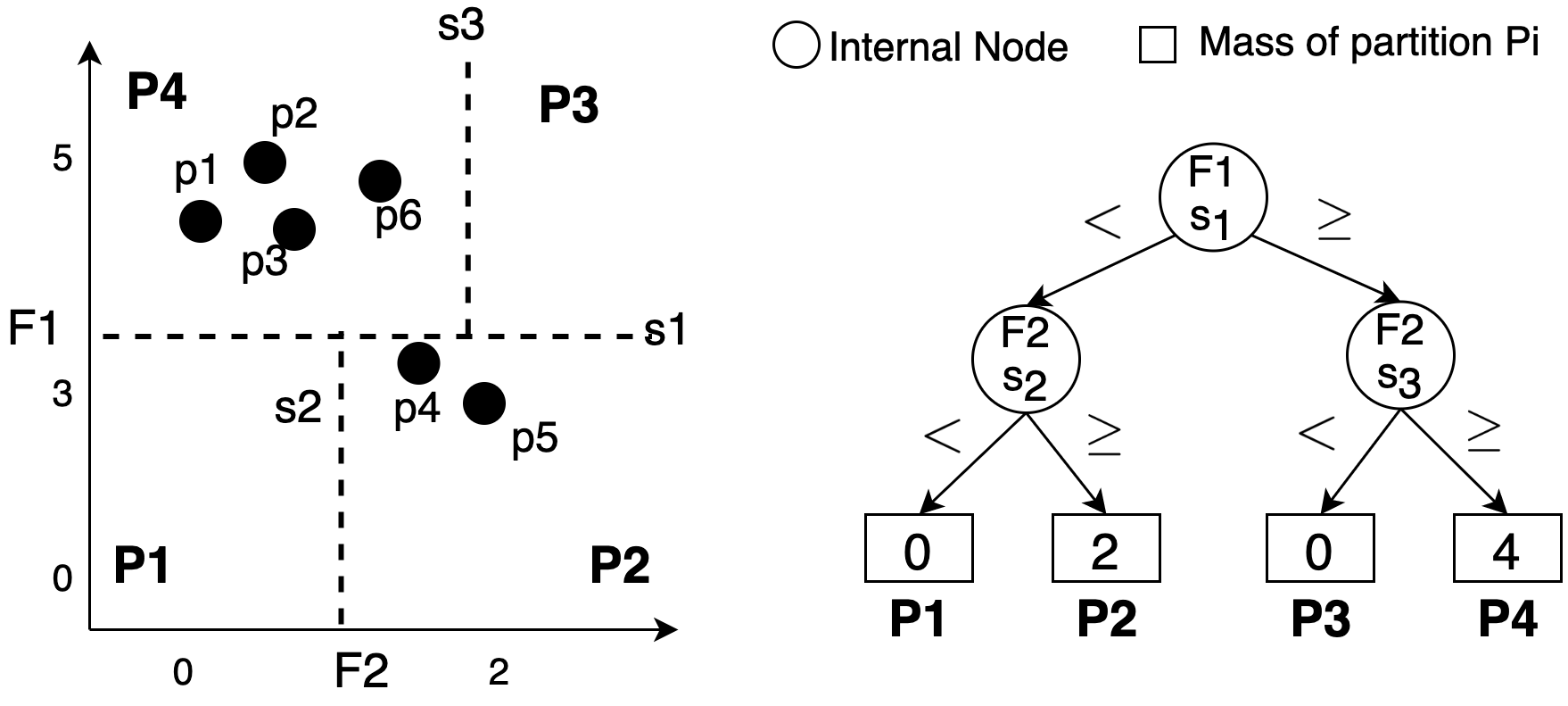}
    \label{fig:hstTumbWin1fspace}
    }\qquad
    \subfloat[Partitions (left) and model (right) on tumbling window 2.]{\includegraphics[width=0.82\linewidth]{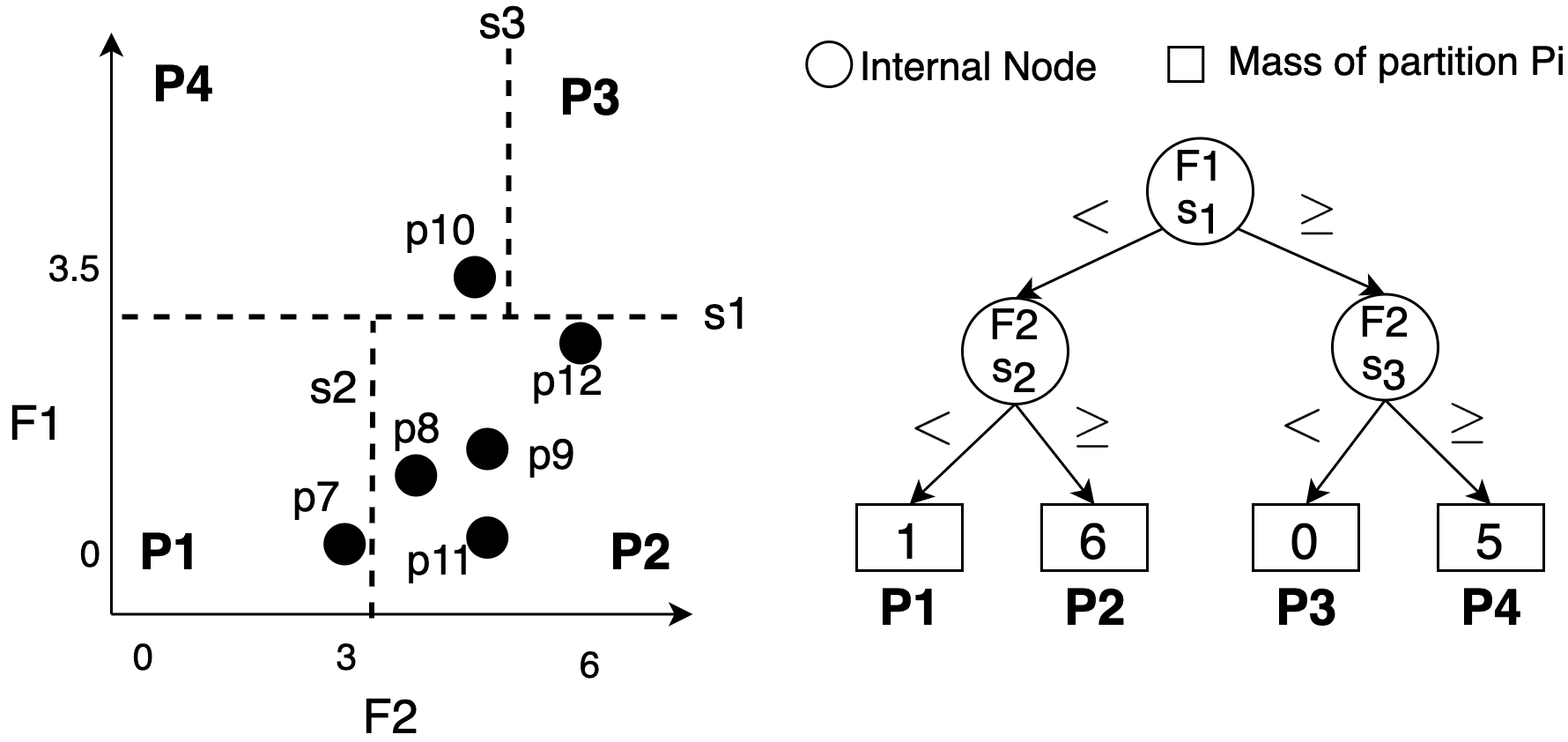}
    \label{fig:hstTumbWin2fspace}
    }
    \caption{HST partitions (left) and constructed model (right) when running on the tumbling windows of Figure \ref{fig:tumblingWindowEx} in the feature space with $T=1$ trees and max height $h=2$.}
    \label{fig:hstTumbWindowsFspace}
\end{figure}%
A running example of HST/F is illustrated in Figure \ref{fig:hstTumbWindowsFspace}. In this example we assume that the max height $h=2$ and only one tree $T=1$ is built. First, the reference window is constructed on the first six samples (see Figure \ref{fig:hstTumbWin1fspace}), leading to 4 partitions based on a random splitting value $s \in \mathit{wr}'(F)$ of a randomly selected feature $F$ at each step. For the second window (see Figure \ref{fig:hstTumbWin2fspace}), we assess the abnormality of each new sample based on the mass of the partition that it falls into, computed from the preceding (reference) window. Therefore, sorting the six samples of the second window in \emph{descending} score value order (indicating increasing anomaloussness) yields : $\langle p_{10}, \{p_8, p_{11}, p_9, p_{12}\}, p_{7} \rangle$, where $\{\cdot\}$ indicates a tie. Sample $p_7$ is the most abnormal sample because it falls into partition $P_1$ with 0 mass. After the scores of the new samples have been computed, the latest window becomes the reference window and the mass of each partition is updated accordingly. An interesting case is the sample $p_{10}$ that is the most normal among the six samples in the second window because it falls into partition $P4$ with mass 4 (see Figure \ref{fig:hstTumbWindowsFspace}). The aforementioned behavior shows that if a plethora of samples are concentrated in a partition in the beginning of the stream but very few samples fall into that partition as the stream evolves, HST will assign high scores to those samples leading to potential false negatives. The suggested forgetting mechanism of HSTF can reduce this effect by decreasing the mass profiles of such partitions.

HST requires linear time $O(T (2^{h + 1} - 1))$ for model construction\footnote{$2^{h+1} - 1$ is the number of nodes in a perfect binary tree} and linear time $O(t h w)$ for model update, where $w$ is the window size. In the worst case each sample may end up to a different leaf. Thus, all points in a window may update all mass profiles in a different tree traversal. Therefore, complexities are amortized constant when $h$, $t$ and $w$ are set. Note that the forgetting threshold does not change the complexity of the original algorithm.

\subsection{Robust Random Cut Forest (RRCF)}
\label{sec:rrcf}

RRCF is a tree-based detector \cite{rrcf} used by the AWS Data Analytics Engine~\footnote{\url{https://aws.amazon.com/kinesis/}} that learns a sketch of a data stream using an ensemble of Robust Random Cut Trees (RRCT). A RRCT is a full binary tree used to calculate the \emph{collusive displacement} (CoDisp) of a sample. CoDisp measures the differential effect of adding/removing a particular sample from a RRCT. RRCF requires to tune three hyper-parameters: the maximum number of samples \textit{Max Samples} that are used to build a tree during training; the maximum number $f$ of leaf nodes to forget after updates; and the number of trees $T$ of the ensemble. RRCF differs from HST in three aspects: (i) it prioritizes features with higher value range; (ii) it uses a forgetting mechanism to delete old samples and (iii) the anomalies are reported instantly, i.e., before the current window is being processed completely. Subsequently we list the building blocks of RRCF: \\
\textbf{Training  Phase.} RRCF trains the trees of the ensemble by subsampling without replacement few initial sliding windows. \textit{Max Samples} are used to build a tree.
An internal tree node represents a \emph{splitting feature} that is selected \emph{proportionally} to its normalized value range. Features with larger value spaces may contain extreme values and therefore, anomalies. Each internal node has a \emph{splitting value} which is selected randomly and uniformly from the range of the selected feature. With HST, the splitting value essentially partitions samples into smaller subspaces. Every internal node keeps a \emph{bounding box} that stores the value range of the feature at a specific depth. A leave node contains a sample along with its arrival time in the stream and the number of replicas in case that many samples end up in the same leaf. The construction of a tree stops when every sample in the training set is isolated from the remainder of the training data, i.e., falls in a leaf.
\\
\textbf{Model  Update.} RRCF updates incrementally the the trees using sliding windows. When a new sample $p$ traverses the internal nodes of a tree, if the feature values of $p$ exceed the bounding box of the last internal node in the path then a new node is built, otherwise $p$ ends up in the same leaf as another sample, increasing its replica counter.
\\
\textbf{Forgetting Mechanism.} RRCF provides a time-decaying mechanism to forget old samples. When the number of leaves exceed the forgetting threshold $f$, the oldest samples per insertion time are deleted and the tree is restructured accordingly.
\\
\textbf{Anomaly Report.} After the insertion of a new sample in the model, its anomaly score is \emph{immediately computed} unlike HST that requires to process an entire window. RRCF uses an anomalousness criterion called \emph{collusive displacement} (CoDisp). To compute CoDisp, the displacement of a node $n_p$ that sample $p$ traversed through in a tree $t \in $ \textit{RRC-Trees} is computed as:
\vspace{-.3cm}
\[
    \mathit{Disp(n_p,t)} = \frac{\textnormal{number of samples beneath } \mathit{sibling_{n_p}}}{\textnormal{number of samples beneath } {n_p}}.
\]
CoDisp extends the notion of Disp by accounting for duplicates and near-duplicates, called \emph{colluders}, that can mask the presence of anomalies. Given a path of nodes $P$ starting from a leaf node $l$ to the node before the root $r$ of a tree $t \in \textnormal{RRC-Trees}$, the CoDisp of $n_p$ is computed as the average maximal displacement over the traversal path of $p$ across all trees:
$1/T \sum_t \mathit{max(\{Disp(n_i,t) \vert n_i \in P\})}.
$

Intuitively, CoDisp measures the change in the model complexity incurred by the insertion or deletion of $p$. The model complexity here can be represented as the sum of depths for all samples in the tree. Therefore, a \emph{tree-based anomaly} is defined as a sample that significantly increases the depth for a set of samples, when it is included in the tree. 

\begin{figure}
    \centering
    \subfloat[Training on Sliding Window 1.]{
    \includegraphics[width=1\linewidth]{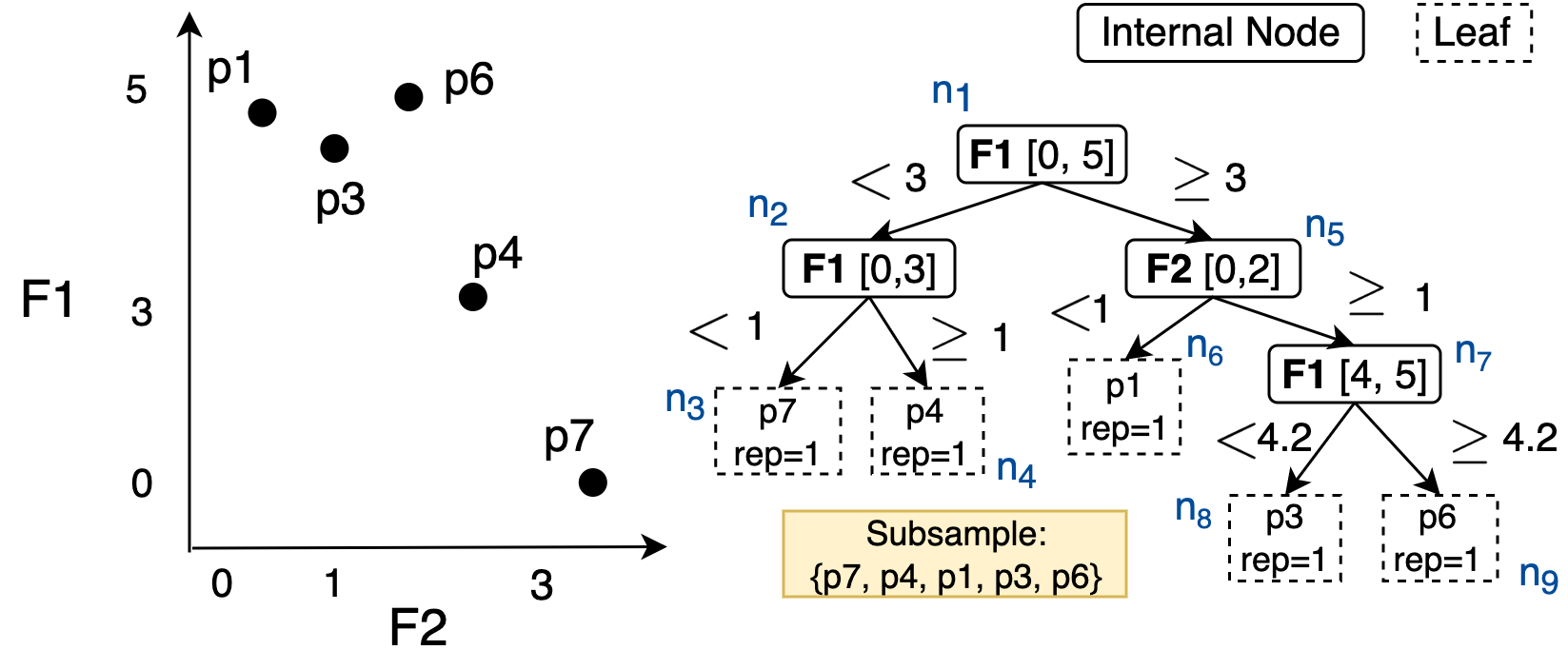}
    \label{fig:rrcfSlidWin1fspace}
    }\qquad \vspace{0.1cm}
    \subfloat[Update RRCF by inserting the samples $p_7....p_{10}$ on the arrival of Sliding Window 2.]{\includegraphics[width=1\linewidth]{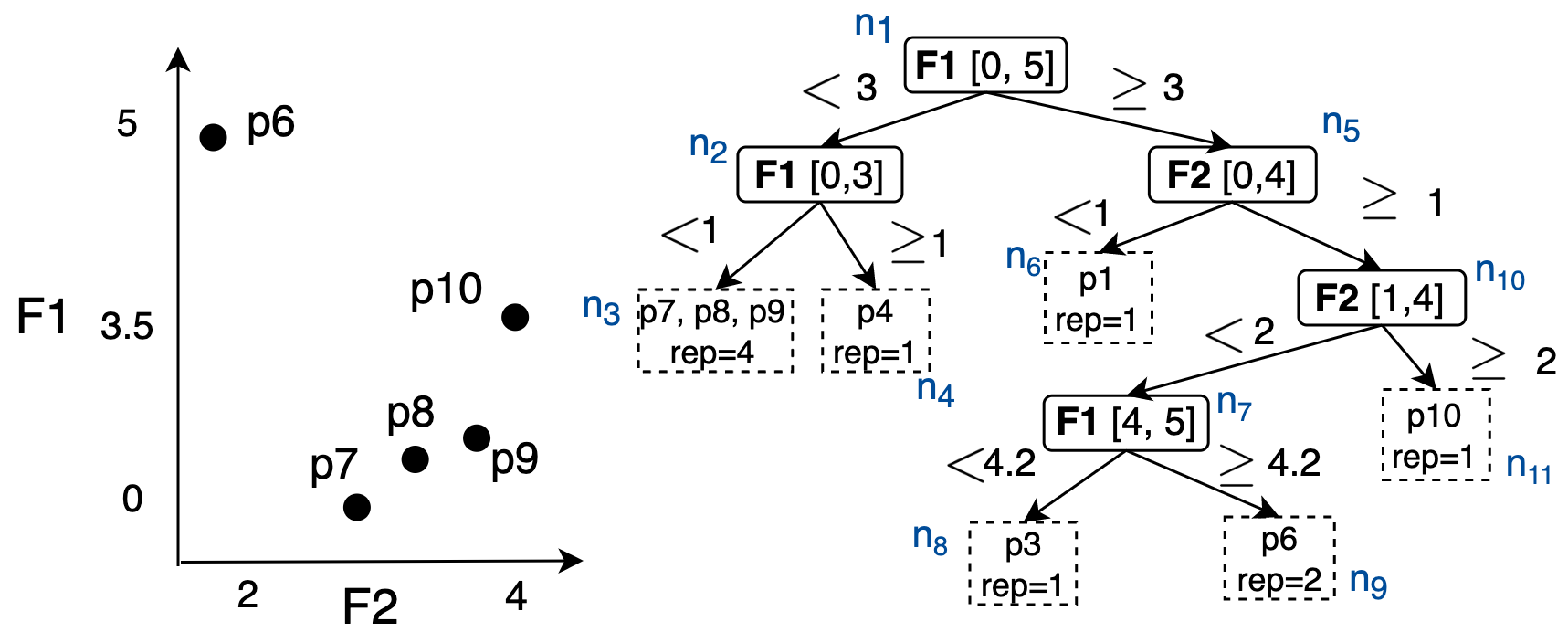}
    \label{fig:rrcfSlidWin2fspace}
    }
    \caption{RRCF model built on sliding Windows of Figure \ref{fig:slidingWindowsEx} in the feature space with $T=1$ and \textit{Max Samples = 5}}
    \label{fig:rrcfModelRunningEx}
\end{figure}

A running example of RRCF is illustrated in Figure \ref{fig:rrcfModelRunningEx}. The model is initially built (training phase) using the first sliding window in Figure \ref{fig:rrcfSlidWin1fspace}. We assign to each node a unique id $n_i$. Every internal node represents the selected feature along with its bounding box (value range). The feature $F1$ is selected at the first step as it has higher value range ($[0,5]$) than $F2$ ($[0,3]$); we depict the splitting value on the edges of each node. When the first window is finished, each sample is isolated in a leaf along with a replica counter (rep). The scoring and update procedures are preformed using the next sliding window in Figure \ref{fig:rrcfSlidWin2fspace}. To keep a neat tree visualization, we present the model up to sample $p_{10}$. Given that a sample is scored only when it is inserted to the tree altering its structure, we compute the CoDisp analytically only for $p_{10}$. Observe that $p_{10}$ exceeds the bounding box of $F2$ and thus a new node is created in depth 3. Therefore, we compute the CoDisp for the nodes $n_{11}, n_{10}$ and $n_5$ resulting in the following values of Disp: $\{3/1, 1/4, 5/5\}$; the CoDisp is the maximum value which is 3. We also report the scores of the rest samples upon their insertion: $p_6$=0.5, $p_7$=1.33, $p_8$=1, $p_9$=0.8. Compared to Disp, CoDisp captures the effect of the deletion at tree-level instead of a leaf-level. This helps recognizing clustered anomalies, i.e., a sample masked by its anomalous neighborhood.

RRCF requires linear time to construct a forest $O(t (2 n - 1))$ \footnote{$2 n - 1$ is the number of nodes of a full binary tree with n leaves.} and logarithmic updating time $O(t  \log(n))$ \cite{rrcf}. In the worst case each sample ends up to a different leaf at the maximal height which requires to update the entire subtree till the root. The CoDisp requires linearithmic time $O(t \log(n) w)$ to update the tree structure for every sample in the window $w$.

\subsection{Lightweight Online Detector of Anomalies (LODA)}
\label{sec:loda}
LODA \cite{loda} is a projection-based detector that constructs an ensemble of $k$ one-dimensional histogram density estimators using sparse random projections. The significant advantage of LODA is that it is hyper-parameter free. Specifically, the number of histograms $k$ can be estimated by measuring the reduction of variance after adding another histogram \cite{Somol2013OnSR} and the number of bins $b$ can be estimated via the method of Birgé and Rozenholc \cite{Birge2006}. LODA can operate in batch (noted as L-B for brevity) or online mode (noted as L-S for brevity) that continuously updates the histograms as the stream evolves. In batch mode, the two hyper-parameters are robustly estimated using all available samples while in online mode, hyper-parameters are estimated using only the training samples. Subsequently, we list the building blocks of LODA in streaming mode (L-S). 
\\
\textbf{Training  Phase.} During training L-S constructs a one-dimensional histogram $j$ as follows. First, a projection vector $w_j$ is built with coefficients $\sim N(0, \mathbf{1}_d)$ where $\sqrt{d}$ of them are selected uniformly at random to be replaced with zeros. The non-zero coefficients denote a subspace of features used to build the histogram $j$. Second, L-S relies on online histograms \cite{BenHaim2010ASP} to approximate the distribution of data by using a set of pairs $H_j = {(z_{1j}, m_{1j}), ...,(z_{bj}, m_{bj})}$, where $z_{ij} = w_{j}^{T} x_i$ is the projection of the $i$-th sample in the $j$-th histogram and $m_{ij}$ is total number of the samples falling into the same projection. When the number of bins exceeds the estimated threshold $b$, the two nearest projections $z_{ij}$, $z_{lj}$ are merged to form the new pair: $(\frac{z_{ij}\cdot m_{ij} + z_{lj}\cdot m_{ij}}{m_{ij} + m_{lj}}, m_{ij} + m_{lj})$. Finally, two additional pairs for the minimum and maximum projections are added: $H_j \leftarrow H_j \cup \{(z_{min}, 0), (z_{max}, 0)\}$.
\\
\textbf{Model Update.} L-S operates in tumbling windows using one of the following modes: (a) under a continuous histograms mode \cite{BenHaim2010ASP} each sample is first scored and then inserted to the histograms; (b) under a two alternating histograms mode samples are inserted to the model after all been scored i.e., upon the completion of the new window similarly to HST \cite{hst}. The updating process is the same as for building the initial histograms while their minimum and maximum bounds may be updated by new samples. Note that the number of histograms does not change during updates.
\\
\textbf{Forgetting Mechanism.} Similar to the original implementation of HST, L-S does not employ a forgetting mechanism. Therefore, the frequency of the bins can only increase as the stream evolves. 
\\
\textbf{Anomaly Report.} To compute the score of a sample $x_{i'}$, the projection $z_{i'j}$ for a histogram $j$ is computed and the two nearest projections (if exist) $z_{ij} < z_{i'j} <  z_{i+1j}$ are used to calculate the anomaly score:
\vspace{-.5cm}
\begin{equation}
\hat{p}(x_{i'}) = \frac{1}{k} \sum_{j=1}^k \frac{z_{ij}\cdot m_{ij} + z_{i+1j}\cdot m_{i+1j}}{2 M_j (z_{i+1j} - z_{ij})},
\end{equation}
where $M_j = \sum_{i=1}^b m_{ij}$. If a sample falls in a sparse region, it receives a lower score indicating more anomalousness. Note that if $\not \exists ~ i: z_{ij} < z_{i'j} <  z_{i+1j}$, the score cannot be computed.

\begin{figure}
    \centering
    \includegraphics[width=1.0\linewidth]{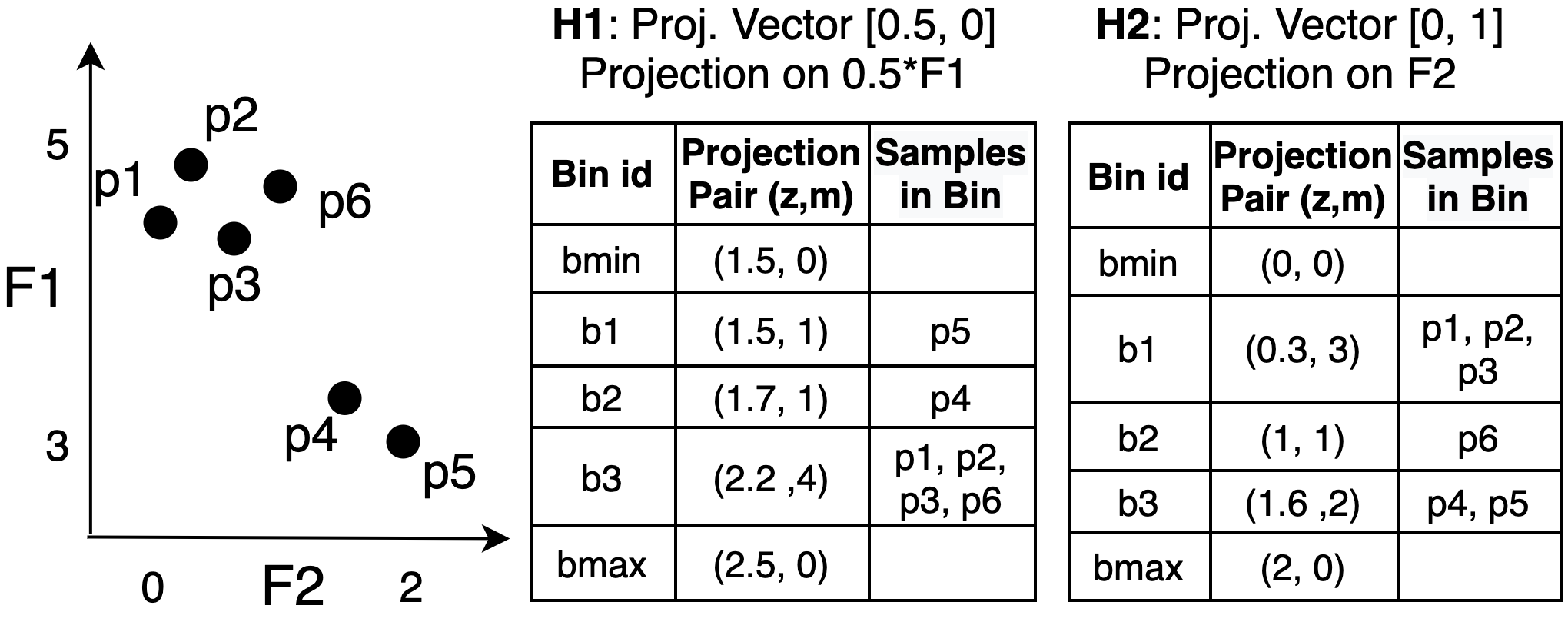}
    \caption{LODA-streaming model trained on tumbling window 1 of Figure \ref{fig:tumblingWindowEx} in the feature space with $b=3$ bins and $k=2$ histograms.}
    \label{fig:lodaModelRunningEx}
\end{figure}

A running example of L-S trained on the first six samples of tumbling window 1 of Figure \ref{fig:tumblingWindowEx} is depicted in Figure \ref{fig:lodaModelRunningEx}. L-S estimates the number of bins $b=3$ (not considering the min/max bins) and the number of histograms $k=2$. Since there are two histograms, there are two one-dimensional random projections. For $H_1$ the feature $F1$ is selected with projection vector $w_1 = [0.5, 0]$ and for $H_2$ the $F2$ is selected with projection vector $w_2 = [0, 1]$. Then the samples of second window of Figure \ref{fig:tumblingWindowEx} arrive (see Figure \ref{fig:hstTumbWin2fspace} for the value ranges). The first sample to be scored is $p_7 = (3,0)$~\footnote{We consider the coordinates of each point as ($F2, F1)$}. However, the projections of $p_7$ are $\langle 0,0 \rangle$ in $H_1$ and $\langle 0,3 \rangle$ in $H_2$ which are outside the min/max bounds and thus its anomaly score cannot be computed. Then, $p_7$ is inserted creating the bins $b'_{ H_1} = (0, 1)$ and $b'_{H_2} = (3, 1)$. As the number of bins are $4 > 3$ (not considering the min/max bins), the $b_{1, H_1}$ and $b_{2, H_1}$ are merged into $b_{12, H_1} = (1.6, 2)$ and the $b_{2, H_2}$ and $b_{3, H_2}$ are merged into $b_{23, H_2} = (1.3, 3)$. Finally the min/max bounds are updated: $b_{\mathit{min}, H_1} = (0,0)$ and $b_{\mathit{max}, H_2} = (3,0)$ for $H_2$.The next sample is $p_8 = (3.5, 1)$ projected as $\langle 0.5,0 \rangle$ in $H_1$ and $\langle 0,3.5 \rangle$ in $H_2$. Therefore, it can only be scored in $H_1$, being between $b'_{H_1}=0< 0.5 < 1.6=b_{12, H_1}$. Thus $\hat{p}_8 = (0 \cdot 1 + 1.6 \cdot 2) / (2\cdot7\cdot(1.6-0)) = 0.28$.

Given $k$ histograms with $b$ bins each, the time complexity of L-S is $O(Nk(d^{-\frac{1}{2}} + b))$ for training, where $N$ is the number of samples, and $O(w k)$ for updating and scoring, where $w$ is the window size.

\subsection{XSTREAM}\label{sec:xstream}

XSTREAM \cite{xstream} is a projection-based  detector that constructs an ensemble of half-space chains (HSC) serving as density estimators. Unlike all previous detectors it is able to cope with feature-evolving data streams. XSTREAM requires to tune three hyper-parameters: The number of random projections $K$, the number of HSC $M$ and the depth of HSC $D$. XSTREAM can operate in batch (noted as X-B for brevity) or online mode (noted as X-S for brevity). Both rely on sparse random projections of samples over a subset of features. In each random projection, $1/3$ of the features are set as non-zero, weighted by $\{-\sqrt{3/K}, \sqrt{3/K}\}$ with equal probability. Then, each sample $p$ from $\mathbb{R}^d$ is projected to $\mathbb{R}^K$ to obtain a new projected sample $\langle r_1^T p, ..., r_K^T p\rangle$, where $d$ denotes the dimensionality of the dataset and $r_i$ a random projection. Subsequently, we list the building blocks of X-S. 
\\
\textbf{Training Phase.} During training, a fraction of samples are kept to estimate the value range $\Delta_{F'_i} = (\mathit{max}_{F'_i} - \mathit{min}_{F'_i}) / 2$ of each constructed feature $F'_i \in \mathbf{F'} = \{F'_1,..., F'_K\}$. At each level of a HSC, a feature $F'_i$ is selected randomly with replacement from $\mathbf{F'}$. The value of $F'_i$ selected at level $l$ in the bin-vector $\mathbf{z}_l$ of $p$, is computed using the following equations:
\vspace{-.3cm}
\begin{equation}
  z_{F'_i, l}=\begin{cases}
    (p_{F'_i} + s_{F'_i}) / \Delta_{F'_i}, & \text{if $o(F'_i, l) = 1$}.\\
    (2z_{F'_i} - s_{F'_i}) / \Delta_{F'_i}, & \text{if $o(F'_i, l) > 1$}.
  \end{cases}
\end{equation}
\noindent
where $p_{F'_i}$ is the value of $F'_i$ of a projected sample $p$,  $s_{F'_i}$ is a random shift drawn from $\mathit{Uniform}(0, \Delta_{F'_i})$ and $o(F'_i, l)$ is the frequency that a feature has been sampled till the $l$-th level. The bin vector at $l$ level is computed as $\mathbf{z}_l = \{z_{F'_i, l} \vert F'_i \in F'\}$. To maintain the bin-count at every level for a chain $c$, X-S relies on \emph{count-min-sketch} ($cms$) structure, noted as $\mathbf{H}_c = \{H_{c,l} \vert  l=1...D\}$. For a specific level $l$ of a HSC $c$, $H_{c,l}$ is updated as: 
\vspace{-.3cm}
\begin{equation}
\label{eq:xstream-cms}
  H_{c,l}=\begin{cases}
     H_{c,l}[\floor{\mathbf{z}_l}] +1, & \text{if $\floor{\mathbf{z}_l} \in H_{c,l}$}.\\
    1, & \textnormal{otherwise}.
  \end{cases}
\end{equation}
\noindent
\textbf{Model Update.} Like HST \cite{hst}, X-S operates with two alternating tumbling windows. Two $cms$ structures are maintained $\mathbf{H}_\mathit{ref}$ and $\mathbf{H}_\mathit{cur}$ for the reference and current window respectively. The samples of the current window are scored using the bin-counts of $\mathbf{H}_\mathit{ref}$, and the bin-counts of $\mathbf{H}_\mathit{cur}$ are updated as in training. When all samples of the current window have been scored, the $\mathbf{H}_\mathit{ref}$ is replaced by $\mathbf{H}_\mathit{cur}$ and the counts of $\mathbf{H}_\mathit{cur}$ are set to zero. This technique lets X-S handles drifts in data distribution between two consecutive windows.
\\
\textbf{Forgetting Mechanism.} When all samples of the current window are inserted into $\mathbf{H}_\mathit{cur}$, the bin-counts learned in the reference window are replaced by the ones of the current window. Therefore, whenever the window slides, X-S forgets together all samples of the reference window.
\\
\textbf{Anomaly Report.} The anomaly score for a projected sample is the minimum bin-count across all levels of a chain, averaged out across all HSC:
    $\frac{1}{M} \sum_{c \in C} \mathit{min}_l~ 2^lH_{c,l}[\floor{\mathbf{z}_l}]$ ,
where C is the set of the $M$ chains. The intuition behind the scoring function is to measure the anomalousness of a sample across the different feature granularities and report the score that corresponds to the lowest density this sample is located at.

\begin{figure}
    \centering
    \includegraphics[width=1.0\linewidth]{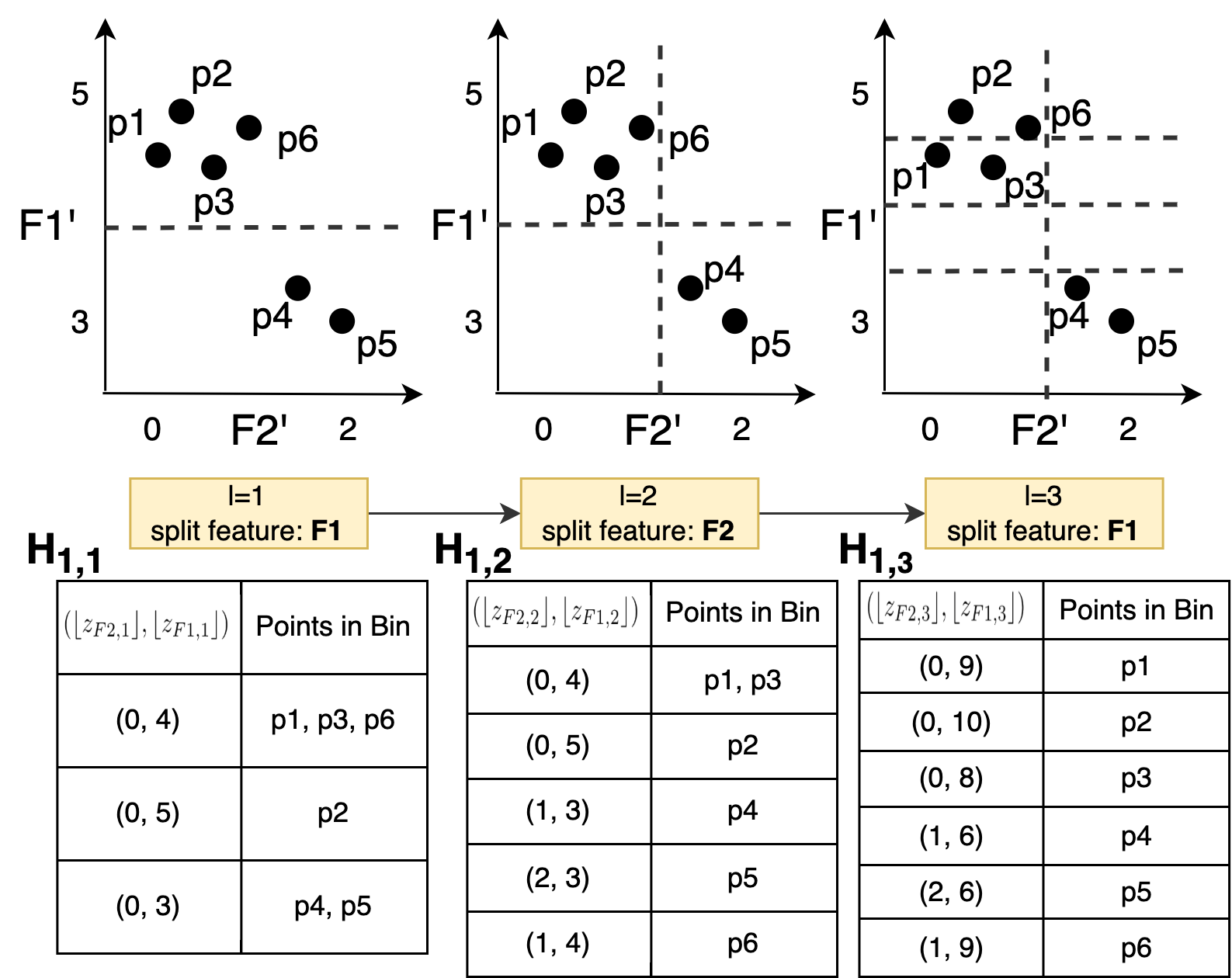}
    \caption{XSTREAM model trained on tumbling window 1 of Figure \ref{fig:tumblingWindowEx} in the feature space with depth $D=3$, $M=1$ chain and $K=2$ projections. For simplicity we assume that the projections are identical to the original feature values and the random shift $\mathbf{s} = (0,0)$ for each feature.}
    \label{fig:xstreamModelRunningEx}
\end{figure}

A running example is illustrated in Figure \ref{fig:xstreamModelRunningEx}. The first tumbling window of samples $\{p_1,...,p_6\}$ (see Figure \ref{fig:tumblingWindowEx}) serves as the reference window, and is used to build the initial $cms$ structures $H_{1,l}$ for $l=1,...3$ ($M=1$  and $D=3$). First, samples are projected using two random projections $r_1, r_2$: the $F'_1$ is comprised by $r_1 = [1, 0]$ using only the values of $F_1$ and $F'_2$ is comprised by $r_2 = [0, 1]$ using only the values $F_2$. The  halved value range $\Delta_{F'_2}, \Delta_{F'_1}$ of each feature is 1: $\mathbf{\Delta} = (1,1)$. Subsequently, we compute the bin vector of sample $p_6$ across the chain levels. In $l=1$, the feature $F1$ is selected and $p_6$ has the bin vector $\mathbf{z}_1 = (0, 4.7)$. In level 2, $F'_2$ is selected leading to $\mathbf{z}_2 = (1, 4.7)$ and in the last level $F'_1$ is selected again yielding $\mathbf{z}_3 = (1, 9.4)$. Therefore, the discretized bin vectors are $\mathbf{z}_1 = (0, 4), \mathbf{z}_2 = (1, 4), \mathbf{z}_3 = (1, 9)$. Since there is distribution shift in the second tumbling window, all samples get a zero score, which is the lowest possible density. For the projected sample $p_1$, the bin-counts are: 3 ($l=1$), 2 ($l=2$), 1 ($l=3$), and its anomaly score is 1. 

X-S has linear time complexity $O(N  K  m  D  M)$ to construct the $cms$ structure, where $N$ is the number of training samples, $K$ is the number of projections, $m$ is the number of fixed-size hash tables to approximate the bin counts, and $M$ is the number of HSC of depth $D$. The time complexity to update the $cms$ structure is $O(w K  m  D  M)$ with $w$ being the window size.

\subsection{Randomized Subspace Hashing (RS-Hash)}
\label{sec:rs-hash}

RS-Hash \cite{rshash} is a density-based detector that operates in subspaces. It constructs an ensemble of histograms on feature subspaces, serving as density estimators as in XSTREAM and LODA. RS-Hash requires to tune three hyper-parameters: the number of hash functions $h$, the sub-sample size $s$ and the number of repetitions $m$. 

The main idea is to repeatedly construct grid-based histograms on sub-samples and combine the obtained scores in an ensemble fashion. Each histogram is built on a sparse, randomly chosen subspace of the original feature space. The features of a subspace with dimensionality $r$ are sampled uniformly at random from $(1+0.5 \cdot log_{max(2,1/f)}(s), log_{max(2,1/f)}(s))$, where $f$ is a locality sampled uniformly at random from $(1/\sqrt{s}, 1-1/\sqrt{s})$. Unlike XSTREAM and LODA, RS-Hash assumes that the $r$ features have equal weight and the histograms are constructed on the original sample values of these features rather than their inner product with the selected subspace. After selecting the subspace features, each sample is normalized using min-max normalization and histograms are constructed using a \emph{count-min-sketch (cms)} structure, as in XSTREAM. In total, $h$ histograms are built for a particular subspace. The process is repeated $m$ times, with a different sub-sample hashed on a different subspace. Subsequently, we report the building blocks of RS-Hash:
\noindent
\\\textbf{Training Phase.} During training, a fraction of samples are kept to construct the initial histograms. We denote a histogram $j$ of a sub-sample $i$ as $H_{ij}$ that uses a particular hash function. To maintain the bin-count, RS-Hash leverages the cms structure that hashes the values of a sample $p$ as in Eq. \ref{eq:xstream-cms} of XSTREAM. The difference is that the values of $p$ are normalized but not projected as in XSTREAM.
\\\textbf{Model Update.} When a new sample $p$ arrives, RS-Hash first scores $p$ and then the histograms' counts are updated.
\\\textbf{Forgetting Mechanism.} Whenever the window slides, RS-Hash forgets all the expired samples by reducing the counts of the corresponding hash buckets\footnote{in our implementation we consider the simple forgetting mechanism rather than the time-decaying mechanism}.
\\\textbf{Anomaly Report.} To compute the score of a new sample $p$, the non-used features in a subspace are encoded as -1 and the remaining ones are normalized. The anomaly score is the minimum bin-count across the histograms of a sub-sample, averaged over the different sub-samples:
\vspace{-.3cm}
\begin{equation}
\label{eq:rs-hash-score}
\mathit{Score(p)} = \frac{1}{m} \sum_{i=1}^m \mathit{log_2(}\mathit{min_{j}}H_{ij}[p] +1).
\end{equation}
We reuse the running example of LODA in Figure \ref{fig:lodaModelRunningEx}, where the first window of Figure \ref{fig:tumblingWindowEx} is used for training RS-Hash. We set $w=2$ resulting to two histograms and $m=1$. We select four samples uniformly at random: $p_1, p_3, p_5, p_6$ and for Thus, we take the histograms $H_{11}=\{b_1:[p_1, p_3, p_6], b_2:[p_5]\}, b_3:[]$, $H_{12}= \{b_1:[p_1, p_3], b_2:[p_6], b_3:[p_5]\}$, where $b_i$ denotes the bucket $i$; the buckets may differ due to different hash functions. After training, the samples of window 2 of Figure \ref{fig:slidingWindowsEx} arrive. We assume that the first sample $p_7$ falls in $b_3$ in both hash tables. The minimum count is 1 from $H_{11}$ so it receives a score of 0 (see Eq \ref{eq:rs-hash-score}). Then, it increases the count of $b_3$ by 1 in both hash tables. The scoring continuous for the remaining samples respectively. Recall that each point is first scored and then the cms structure is updated. After each sample is scored, the forgetting mechanism will be activated, reducing the counts in the corresponding buckets.

RS-Hash has linear time complexity for training $O(shm)$ including the cms construction, and $O(whm)$ for updating and scoring, where $s$ is the subsample size, $h$ the number of hash functions, $m$ the repetitions and $w$ is the window size.

\subsection{STAtionary REgion skipping (STARE)}
\label{sec:stare}

STARE \cite{stare} is a density-based detector that identifies the top-n local anomalies in sliding windows. STARE requires three hyper-parameters to be tuned: The number $k$ of nearest kernel centers $\theta_k$, the diagonal length of a grid cell $\theta_R$ and the error allowance threshold $\gamma$.

STARE relies on a kernel density estimation (KDE) to compute the density around each sample. In contrast to other KDE-based methods, STARE does not globally update the samples' densities for every window slide. Instead, it optimizes density estimation based on the observation that data distributions
in many regions hardly change across window slides - a notion called \emph{stationary region skipping}. STARE operates in three phases: (i) \emph{Data distribution approximation}, (ii) \emph{Cumulative net-change-based skip} and (iii) \emph{Top-n Anomaly Detection}. In the first phase, STARE divides the space into $d$-dimensional grid cells of diagonal length $\theta_R$, where $d$ is the data dimensionality. Each grid $c$ has a kernel center that represents the samples that fall into a grid while empty grids are not considered. STARE stores a weight distribution grid $\mathbb{G}$ with the number of samples in each $c$. In the second phase, STARE examines the changes in $\mathbb{G}$, denoted as net-weight distribution grid $\Delta\mathbb{G}$, between the current and previous slide to avoid updating the local densities of samples in stationary regions. Note that some regions may change slightly, e.g., only one sample is removed from each grid; in this case, STARE relies on a threshold $\gamma$ to skip updates of slightly changed regions. Higher $\gamma$ values\footnote{Authors report that the value $\gamma=0.01$ is optimal for the most datasets.} increase the error in density estimation, sacrificing accuracy for efficiency. In the last phase, STARE first searches in candidate cells, that are guaranteed to contain the top-n anomalies. To do that, density bounds are computed for each cell. For the candidate cells, STARE performs a point-level detection to the candidate cells. Subsequently, we report the building blocks of STARE:
\begin{figure}[t!]
    \centering
    \subfloat[Sliding Window 1.]{
    \includegraphics[width=0.43\linewidth]{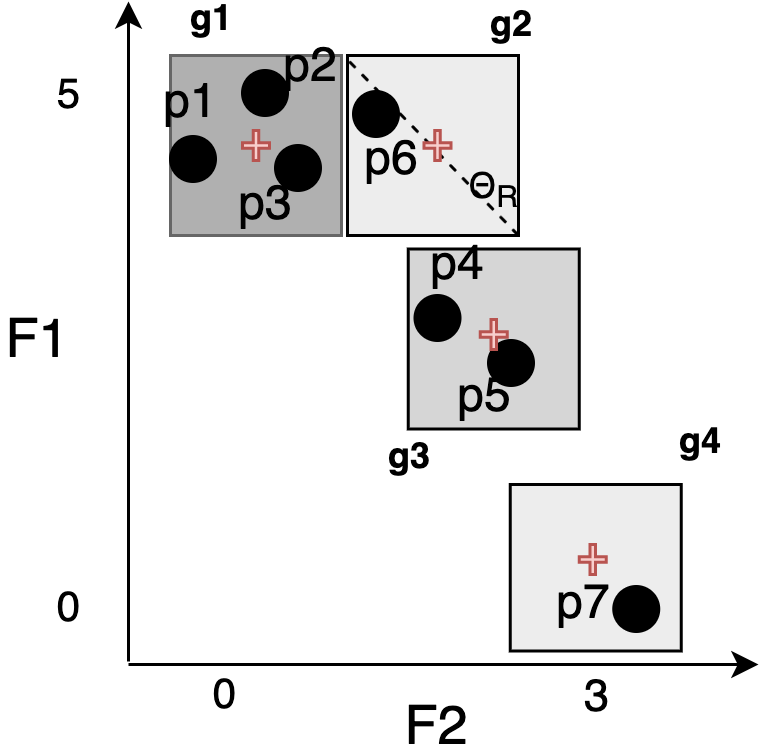}
    \label{fig:stareSlide1}
    }\qquad
    \subfloat[Sliding Window 2.]{\includegraphics[width=0.43\linewidth]{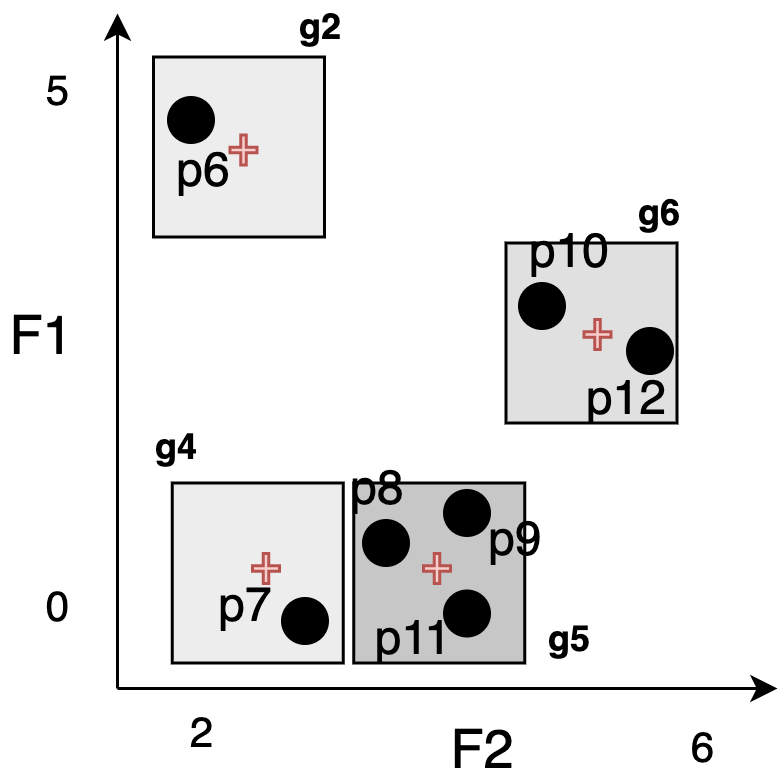}
    \label{fig:stareSlide2}
    }
    \caption{STARE running example on sliding windows of Figure \ref{fig:slidingWindowsEx}. A square is a grid cell $g_i$ with a kernel center (red cross), representing specific samples.}
    \label{fig:stareSlides}
\end{figure}
\noindent
\\\textbf{Training Phase.} STARE uses the first window to partition the data space into grid cells, to store their centers and to compute the densities for all samples.
\\\textbf{Model Update.} Each new sample is indexed to a grid cell, updating its sample count and therefore its weight distribution.
\\\textbf{Forgetting Mechanism.} When a slide expires, the weight distribution of each affected grid cell is updated by reducing its sample count.
\\\textbf{Anomaly Report.} The density of a sample $p$ is calculated as the weighted average density over kernel centers near to $p$:
$\mathcal{D}(p) = \sum_{i=1}^{\theta_k} \frac{w_i}{\sum_{j=1}^{\theta_k}} \prod_{l=1}^d \mathcal{K}_{h^l}(\mathit{dist}(p^l, kc_i^l))$, where $\theta_k$ is the distance to the k-th nearest kernel center $kc_i$ of $p$, $K_h$ is the kernel function taking as input the distance between $p$ and $kc_i$ in a univariate fashion for a dimension $l$. The score of a sample $p$ is given by $\mathcal{S}(p) = (\mu - \mathcal{D}(p)) / \sigma$, where $\mu, \sigma$ are the mean and standard deviation of the local densities at the $\theta_k$ nearest kernel centers of $p$. The anomaly score ranges from $-\infty$ to $+\infty$, where high values indicate lower density, i.e., more anomaloussness.

A running example is depicted in Figure \ref{fig:stareSlides}. In the first slide of Figure \ref{fig:stareSlide1}, STARE forms the non-empty grid cells and computes the density for each sample. In this window, samples are ranked in decreasing order of their score as follows: $p_4 > p_5 > p_7 > p_6 > p_2 > p_1 > p_3$. The first two samples have greater score than $p_7$ as they fall near to a dense region; thus they are assessed to be anomalous. On the next slide of Figure \ref{fig:stareSlide2} the samples of grid cells $g_1, g_3$ are expired and two new cells are formed $g_5, g_6$. Assuming $\theta_k = 2$, only the two nearest kernel centers will be examined for each sample. The $\gamma$ threshold requires at least four samples to get expired in a cell in order to re-compute the samples' density in the cell. In this window, the weight distribution has changed from the previous, affecting the cells $g_2$ and $g_4$; thus the net-change mechanism is activated. However, since only three samples expired in $g_1$ (the nearest kernel to $g_2$), the density of $p_6$ will not be updated, as well as $p_7$. The new ranking will be $p_{6} > p_{10} > p_{12} > p_7 > p_8 > p_9 > p_{11}$. Observe that $p_7$ should have now received the highest score as it is slightly far from a dense area while $p_6$ should have received lower score as it lies on a sparse area far from dense areas after the update. Of course, with a different choice of $\gamma$ this behaviour can be avoided. 

STARE has time complexity $O(w + N^2_G)$ for training in the first window and $O(w + rN^2_G)$ for updating in subsequent slides, where $w$ is the window size, $N_G$ is the non-empty grid cells and $r$ is the ratio of changed grid cells between two consecutive windows.

\begingroup
\def\arraystretch{1.5}

\begin{table*}[ht]
\caption{Qualitative comparison of online detectors.}
\label{table:online_detectors_comparison}
\resizebox{\textwidth}{!}{%
\begin{tabular}{|c|c|c|c|c|c|c|c|}
\hline
       & \textbf{Split Feature} & \textbf{Split Value} & \textbf{Scoring} & \textbf{Window Type} & \textbf{Model Update} & \textbf{Training Time} & \textbf{Updating Time}        \\ \hline
\textbf{HST/F}   & Uniformly                               & Mid Value                             & Mass Profile                      & Tumbling                              & Mass Profile                           & $O(t    (2^{h + 1} - 1))$               & $O(t h  w)$                                    \\ \hline
\textbf{RRCF}    & Proportional to FVR                     & Uniformly                             & Co. Displacement                  & Sliding                               & Reconstruction                         & $O(t    (2    n - 1))$                  & $O(t  \log(n) w)$                                \\ \hline
\textbf{MCOD}    & -                                       & -                                     & Micro Cluster                     & Sliding                               & Reconstruction                         & -                                       & $O(w\mathit{log}(w) + K\mathit{log}(K))$ \\ \hline
\textbf{XSTREAM} & Uniformly                               & Uniformly                             & Bin-Count                         & Tumbling                              & Reconstruction                         & $O(N  K  m  D  M)$                      & $O(w K  m  D  M)$                                \\ \hline
\textbf{LODA}    & -                                       & -                                     & Density                           & Tumbling                              & Reconstruction                         & $O( Nk(d^{-\frac{1}{2}} + b))$          & $O(w k)$                                         \\ \hline
\textbf{CPOD}    & -                                       & -                                     & Cores                             & Sliding                               & Reconstruction                         & -                                       & $O(N_c~w + N_f~N_r)$                           \\ \hline
\textbf{LEAP}    & -                                       & -                                     & Minimal Probing                   & Sliding                               & Reconstruction                         & -                                       & $O(w^2)$                                       \\ \hline
\textbf{RS-HASH} & -                                       & -                                     & Bin-Count                         & Tumbling                              & Reconstruction                         & $O(s ~ h ~ m)$                          & $O(w ~ h ~ m)$                                 \\ \hline
\textbf{STARE}   &      -                                   &          -                             &         KDE                          & Sliding                                      & Reconstruction                                       &           $O(w + N^2_G)$                               &          $O(w + rN^2_G)$                                       \\ \hline
\end{tabular}%
}
\end{table*}
\endgroup
~

\subsection{Summary}
Table \ref{table:online_detectors_comparison} summarizes the main characteristics of the online detectors included in our benchmark. Online detectors are essentially incremental versions of offline detectors that assess anomalousness of samples using similar criteria. According to their authors, HST and RRCF are tree-based online detectors inspired by IF. MCOD \cite{mcod} and CPOD \cite{cpod} are distance-threshold based while LEAP \cite{leap} is a nearest-neighbor-based inspired by KNN$_W$. STARE \cite{stare} is a \emph{density-based} detector on the full feature space as LOF \cite{lof}, while RS-Hash \cite{rshash} working on feature subspaces trace its roots back to HICS \cite{hics}. Offline detectors (detailed in Appendix~\ref{app:offline}) constitute essentially the \textit{baselines} for the effectiveness of online detectors. 

Regarding the time-complexity for the \emph{model update} of the employed online detectors, according to Table \ref{table:online_detectors_comparison}, they are divided into linear (XSTREAM, LODA, RS-HASH, CPOD, HST/F), linearithmic (MCOD, RRCF) and quadratic (LEAP, STARE) at worst case. Note that some of the reported complexities may differ from the analytical complexities reported by other works such as MCOD’s \cite{best_distance_based}. This is because we took into account the data characteristics. For instance, in MCOD if each pair of points have a distance greater than $R/2$, i.e., the data are sparse, or the hyperparameter $K$ is set to a large value, then no micro-cluster will be formed, resulting in linearithmic neighbor search. In CPOD, if the data are sparse, many cores will be formed; if many samples are far from their core, in range $(R, 2R]$, then each core will be explored, resulting to computational overhead. In LEAP, when the minimal probing principle fails, the algorithm has quadratic complexity, which can be reduced with advanced data structures. In STARE, if the data distributions between many consecutive windows  differ more than the predefined threshold $\gamma$, the algorithm will have quadratic complexity, as no region will be stationary. The data characteristics will determine the ranking of the algorithms w.r.t. execution time.

\section{Experimental Environment} \label{chapter:benchmark_environment}

Our experimental evaluation relies on datasets widely used in previous empirical studies \cite{Wang2019, best_offline_1, best_offline_2, best_offline_3,best_distance_based}. These set-based datasets contaminated with point anomalies \cite{Chandola2009} exhibit different characteristics of abnormal and normal samples (e.g., anomaly ratio, dimensionality) and are useful for an unbiased comparison of offline and online algorithms over all possible order of arrival of samples in a data stream. We additionally consider the recently proposed Exathlon \cite{Jacob2021} for explainable anomaly detection over times series that overcomes several limitations of previously used benchmarks for temporal data \cite{Wu2021}. These sequence datasets contaminated with interval anomalies \cite{Chandola2012} can be used to evaluate online detectors only for the specific order implied by the timestamps of their samples.

We have implemented in Java the tree-based online algorithms (HST/F and RRCF) and integrated in our testbed the original implementation in Java of MCOD \footnote{\url{https://github.com/Waikato/moa}},  LEAP\footnote{\url{https://infolab.usc.edu/Luan/Outlier/CountBasedWindow/DODDS/src/outlierdetection/}}, STARE\footnote{\url{https://github.com/kaist-dmlab/STARE}}, CPOD\footnote{\url{https://github.com/tranvanluan2/cpod}} as well as in C++(Online) of XSTREAM \cite{xstream} \footnote{\url{https://github.com/cmuxstream/cmuxstream-core}}, Matlab(Online)/Python(Offline) LODA \cite{loda} \footnote{Online version: \url{http://agents.fel.cvut.cz/stegodata/tools/}
\\ Offline Version: \url{https://github.com/yzhao062/pyod}}.We also extended the Python implementation of RS-Hash available at github\footnote{\url{https://github.com/bedanta01/Subspace-Outlier-Detection}}. We finally rely on third-party implementations in Python of KNN$_W$, LOF and IF from the scikit-learn library version 0.21.2 \footnote{\url{https://scikit-learn.org/}} and the original implementation of OCRF \cite{ocrf} \footnote{\url{https://github.com/ngoix/OCRF}}. We used Java Version 1.8, Python Version 3.7.4, and Sci-kit Learn Version 0.21.3. Experiments were conducted on 2-core Intel i7-7500U at 2.7 GHz, with 8 GB of RAM on Windows 10. The platform of the experimental evaluation environment, as well as the scripts used for the analysis of the result can be found on the Gitgub repository: \url{https://github.com/droubo/meta-level-analysis-of-anomaly-detectors}.

\subsection{Datasets}
\label{datasets}

The bulk of our evaluation lies on real data which in their majority are contaminated with synthetically generated anomalies. To experimentally compare detectors w.r.t. particular factors such as anomaly ratio and dimensionality, window size and speed, etc., we used representative datasets with both synthetic abnormal and normal data.  

In unlabeled datasets, point anomalies are synthetically inserted using different methods. In datasets used in classification problems, as anomalies are considered the samples belonging to the \textit{minority} class. In datasets used in clustering problems, anomalies are inserted samples far away from regions of high \textit{density}. In other datasets, anomalies are \textit{implanted} by simply adding noise to the values of their features. Anomalies are additionally characterized as \textit{subspace} when they are visible only to a subset of the dataset's feature space, and as \textit{fullspace} otherwise. Note that our experimental evaluation highlights the behavior of detectors for subspace anomalies that have been less studied in the literature. 

To simulate a stream of samples from a batch dataset (real or synthetic) we generate a sequence of windows of a given size. To guarantee a smooth detection difficulty across all windows, their content is obtained by shuffling normal and abnormal samples. Moreover, anomalies are stratified in windows using a step related to the anomaly ratio of the dataset. Thus, we guarantee that if $\mathit{step} \, \leq \, \mathit{window} \, \mathit{size}$, then each window is going to have a similar number of anomalies. We finally partition windows into training and testing sequences. In this way, we can run a given detector over a specific dataset under multiple possible sample orderings and report its average effectiveness. 

To select the window size for the experiments concerning point anomalies, we relied on prior work on point anomaly detection \cite{rrcf,mcod,hst,loda}. Instead of selecting a fixed window size in advance as in \cite{hst,loda}, we examined three sizes comprising 64, 128 and 256 samples. Regarding the algorithms that operate using sliding windows, we used stride=1 \cite{rrcf}. To ensure a fair comparison of detectors, we investigated the window size that favors the majority of the detectors. In Figure \ref{fig:window_sizes} (see Appendix), we found that a window of 128 samples results in better median MAP performance for the majority of the detectors. As a matter of fact, for many detectors the MAP using the three window sizes is almost identical, due to the stratification of anomalies performed in each window per dataset. We should stress that for AUC we observed the same trend. Hence, in subsequent experiments concerning point anomalies, we decided to use a window of 128 samples.

\subsubsection{Real Datasets}
\label{sec:real_world_datasets}
Table \ref{table:datasets} depicts the characteristics of the real world and synthetic datasets used in our experiments. Real world datasets have been originally introduced in the UCI \cite{uci} and OpenML \cite{open_ml} repositories, for multi-class classification. In our comparison, we use the unnormalized versions of these datasets\footnote{After removing null values and categorical features not treated by our anomaly detectors.} that are made available for the anomaly detection problem from GLOSS \cite{gloss}, ODDS, DATAHUB, DAMI \cite{best_offline_3} and XSTREAM \cite{xstream}. In the majority of these datasets, the anomaly ratio is low (no more than 9\%) with the exception of \textit{Electricity} in which anomalies and normal samples are almost balanced. It is worth also describing how anomalies are implanted in GLOSS. The authors pick randomly 5\% of the samples transforming each one to an anomaly by replacing a randomly picked feature subset with the values of a sample from a different class; the size of each feature subset was chosen uniformly from [$2, \mathit{max}(2, 0.1 \cdot d))$], where $d$ is the dataset's dimensionality. The aforementioned anomaly generation process results in scattered fullspace anomalies.

\begingroup
\def\arraystretch{1.5}
\begin{table}[h]
\centering
\caption{Real datasets with fullspace anomalies and Synthetic datasets with subspace anomalies.}
\label{table:datasets}
\scalebox{0.7}{
\begin{tabular}{|l|c|c|c|}
\hline
\textbf{Name}                          & \multicolumn{1}{l|}{\textbf{\#Samples}} & \multicolumn{1}{l|}{\textbf{\#Features}} & \multicolumn{1}{l|}{\textbf{AR}} \\ \hline
\multicolumn{4}{c}{\textbf{Real Datasets}} \\
\hline
\textbf{http  (ODDS)}                  & 567498                                  & 3                                        & 0.4                             \\ \hline
\textbf{smtp  (ODDS)}                  & 95156                                   & 3                                        & 0.3                             \\ \hline
\textbf{wilt  (ODDS)}                  & 4839                                    & 5                                        & 5.4                             \\ \hline
\textbf{adult  (ODDS)}                 & 48842                                   & 6                                        & 23.9                            \\ \hline
\textbf{Diabetes  (GLOSS)}             & 768                                     & 8                                        & 5.1\\ \hline
\textbf{electricity  (DATAHUB)}   & 10000                                   & 8                                        & 57.5                          \\ \hline
\textbf{pima  (ODDS)}                  & 768                                     & 8                                        & 34.9                            \\ \hline
\textbf{breast-w  (GLOSS)}             & 699                                     & 9                                        & 5                            \\ \hline
\textbf{forestcover  (ODDS)}           & 286048                                  & 10                                       & 1                               \\ \hline
\textbf{magic-telescope  (XSTREAM)}    & 13283                                   & 10                                       & 7.2                             \\ \hline
\textbf{PageBlocks  (DAMI)}            & 5171                                    & 10                                       & 5                               \\ \hline
\textbf{pendigits  (GLOSS)}            & 10992                                   & 16                                       & 5                             \\ \hline
\textbf{Cardiotocography  (DAMI)}      & 1742                                    & 21                                       & 5                               \\ \hline
\textbf{ALOI  (DAMI)}                  & 49534                                   & 27                                       & 3                               \\ \hline
\textbf{AnnThyroid  (GLOSS)}           & 3772                                    & 29                                       & 5                              \\ \hline
\textbf{Hypothyroid}                   & 3772                                    & 29                                       & 5                               \\ \hline
\textbf{WDBC  (DAMI)}                  & 569                                     & 30                                       & 2.7                             \\ \hline
\textbf{Ionosphere  (GLOSS)}           & 351                                     & 34                                       & 5.1                              \\ \hline
\textbf{MNIST}                         & 7603                                    & 100                                      & 9                               \\ \hline
\textbf{Arrhythmia  (GLOSS)}           & 452                                     & 279                                      & 5.1                             \\ \hline
\textbf{madelon  (XSTREAM)}            & 1430                                    & 500                                      & 9.1                             \\ \hline
\textbf{isolet  (XSTREAM)}             & 4886                                    & 617                                      & 8                               \\ \hline
\textbf{letter-recognition  (XSTREAM)} & 4586                                    & 617                                      & 8.5                             \\ \hline
\textbf{InternetAds  (DAMI)}           & 1630                                    & 1555                                     & 2                               \\ \hline
\multicolumn{4}{c}{\textbf{Synthetic Datasets}} \\
\hline
HiCS20$_{(2,3,4,5)SD}$ & 875 & 20 & 2.0 \\
\hline
HiCS40$_{(2,3,4,5)SD}$ & 875 & 40 & 2.0\\
\hline
HiCS60$_{(2,3,4,5)SD}$ & 875 & 60 & 2.0 \\
\hline
HiCS80$_{(2,3,4,5)SD}$ & 875 & 80 & 2.0 \\
\hline
HiCS100$_{(2,3,4,5)SD}$ & 875 & 100 & 2.0\\
\hline
\end{tabular}
}
\end{table}
\endgroup

\subsubsection{Synthetic Datasets}
\label{sec:synthetic_datasets}
Table \ref{table:datasets} depicts the five synthetic datasets we used in our experiments. These datasets have been introduced to evaluate the HiCS subspace anomaly detector \footnote{\url{https://www.ipd.kit.edu/~muellere/HiCS/}} and exhibit a high variability in the number of features (20D-100D) while the number of samples is constant (875). They are normalized and contaminated with subspace anomalies of varying dimensionality (2D, 3D, 4D, 5D) as indicated by the subscripts of their names. 

The peculiarity of the HiCS datasets is that they are contaminated by point anomalies that lie far away from dense regions of normal samples with highly correlated features. LOF has been applied exhaustively on each of 2D, 3D, 4D and 5D feature subspaces to score subspace anomalies. In our experiments, we focus on the 100D HiCS dataset and removed anomalies that are visible to more than one subspace. Then, we generated five dataset variations with 20, 40, 60, 80 and 100 dimensions that contain exactly the same subspace anomalies of a given dimensionality (2D, 3D, 4D or 5D). The objective is to increase the ratio of irrelevant features as we increase the dimensionality of the datasets. In the 20D HiCS all features belong to at least one of the subspaces anomalies are visible i.e., they are \textit{relevant} to the subspace anomalies of the dataset. As the same anomalies are contained in all datasets, in the 100D dataset we have 20 relevant and 80 irrelevant features. We have finally included 18 anomalies per dataset leading to an anomaly ratio of 2\%.

\subsubsection{Exathlon Time Series}
\label{sec:exathlon}

The Exathlon benchmark \cite{Jacob2021} contains repeated executions of 10 different Spark streaming applications (2.3 million samples in total). The dataset is splitted across 93 different files, called Traces, with recordings of a Spark streaming application run (2,283 features) and are grouped by application (9 traces per application on average). Out of these traces we have focused on 34, called disturbed traces, that contain both normal and abnormal samples. They are split into 5 categories based on the type of anomalies they contain \footnote{There are also some cases of unknown anomalies they may appear in any of those categories.} in realistic system health monitoring settings: (i) bursty input traces, (ii) bursty input until crash, (ii) stalled input traces, (iv) CPU contention and (v) process failure traces. Unlike to the previously explained datasets where anomalies are scattered throughout all windows, we are interested in evaluating the effectiveness of online detectors to spot point anomalies that occur in a specific time frame, called \emph{range-based} anomalies. We should stress that in these datasets, the anomalies are produced in specific time-frames under the same conditions and therefore they are clustered. In our experiments, we have used sequence data acquired only from 2 applications (5 total datasets, one for each type of anomaly), as experiments with time series acquired from all applications prove to be very time-consuming.

\subsubsection{Dataset Profiling} \label{sec:metafeatures}

To be able to explain why the effectiveness of detectors is different even on the same dataset, meta-features are extracted from real and synthetic datasets using the python library PyMFE \cite{pymfe}. In addition to general and statistical meta-features explored in previous meta-learning works \cite{metaod,Vanschoren2019}, we  introduce meta-features related to both fullspace and subspace anomalies. 

\begin{table*}[!t]
\centering
\caption{Set of statistical and general dataset meta-features}
\label{table:metafeatures}
\scalebox{0.8}{
\begin{tabular}{|l|c|}
\hline
\multicolumn{2}{@{}c}{\textbf{General}}                                                             \\ \hline
G1. Nr. Samples    & $n$                                                                  \\
G2. Nr.Features     & $p$                                                                \\
G3. Anomaly ratio     & $\frac{\#anomalies}{\#inliers + \#anomalies}$                                                                         \\ \hline

\multicolumn{2}{@{}c}{\textbf{Fullspace related}}                                                   \\ \hline
FR1. Distance between inliers' and anomalies' center of mass    & $\sqrt{inlier_{i,j} - anomaly_{i,j}}$                     \\
FR2. Nr. of canonical correlation between each feature and class  &  $\#can\_corr(feature_i, class) $              \\
FR3. Nr. of features normally distributed    &  $\#norm\_dist(feature_i)   $        \\
FR4. Anomaly to Normal Distance (AND) & $\frac{DistCM - AvgMAD}{AvgMedian}$ \\ \hline
\multicolumn{2}{@{}c}{\textbf{Subspace Related}}                                                   \\ \hline
S1. Mean Canonical correlations   & $\frac{\sum_{i=0}^p can\_corr(feature_i, class)}{p} $                                               \\
S2. Roy's largest root   & $F_{(2_{v1}+2),(2_{v2}+2)} = \frac{2_{v1}+2}{2_{v2}+2} * fmax$                                                           \\
S3. Pillai’s trace    & $ \sum_{j=1}^{s} theta_{j} = tr(!t(E+!t)^{-1}) $                                                              \\
S4. Lawly-Hotelling trace   & $ T^2_g= e \sum_{j=1}^{s} f_{j} $                                                        \\
S5. Wilks’ Lambda value  & $L_{p,!t,e} = \frac{|E|}{|E + !t|} = \prod_{j=1}^p (1-theta_j)$
\\ 
\hline

\multicolumn{2}{@{}c}{\textbf{Value Space}}                                                     \\ \hline
F1. Mean/SD absolute value of the covariance of distinct dataset feature pairs & $\frac{cov(feature_i, feature_j)}{\#pairs}, i \neq j$\\
F2. Mean/SD eigenvalues of covariance matrix &                                    \\
F3. Mean/SD IQR   &     $q75 - q25$                                                          \\
F4. Mean/SD Kurtosis &    $\frac{\mu_4}{\mu_2} $                                                          \\
F5. Mean/SD MAD &     $median(X - \Tilde{X})$                                                           \\
F6. Mean/SD MAX Value  &  $max_X$                                                        \\
F7. Mean/SD MEAN Value & $\mu$                                                       \\
F8. Mean/SD Median Value & $\Tilde{X}$                                                       \\
F9. Mean/SD MIN Value  &      $min_X$                                                    \\
F10. Mean/SD Range of values   &     $max_X - min_X$                                              \\
F11. Mean/SD Standard Deviation of values  & $\sigma$                                   \\
F12. Mean/SD Skewness  &   $\frac{\mu_3^2}{\mu_2^2} $                                                     \\
F13. Mean/SD Sparsity &  $\frac{\#unique values}{\#samples}$                                                             \\
F14. Mean/SD Variance &  $\sigma^2$ \\
F15. Nr. of Statistical Outliers (3 STD) & \#samples 3 std away\\
F16. Nr of distinct highly correlated pairs of features & $\#corr(feature_i, feature_j), i \neq j$\\ \hline
\end{tabular}
}
\end{table*}

More precisely, our meta-level analysis of experimental results explores the following sub-categories of meta-features (see Table~\ref{table:metafeatures}): (1) \emph{General} reporting the number of samples/features and the anomaly ratio of datasets; (2) \emph{Fullspace} characterizing anomalies in the entire feature space of datasets such as the distance between the center of mass of abnormal and normal classes, the number of pairwise correlated features, the number of features whose values are normally distributed, or the difficulty in the separation of abnormal from normal samples called \emph{Anomaly To Normal Distance} (AND); (3) \emph{Subspace} indicating the existence of anomalies in subspaces of the features of the dataset. In case of subspace anomalies, relevant features have a higher correlation with the target variable (i.e., the anomaly class) in comparison to irrelevant ones according to different test statistics like Wilks’ \cite{wilkslambda}, Pillai’s,  Roy’s \cite{roys}, Lawley-Hotelling. (4) \emph{Value space} characterising the skewness of the different feature distributions by taking the average/std value of a statistic such as maximum/minimum. 

\begin{figure}[!t] 
    \centering
  \begin{minipage}[b]{0.5\linewidth}
    \centering
    \includegraphics[width=1\linewidth]{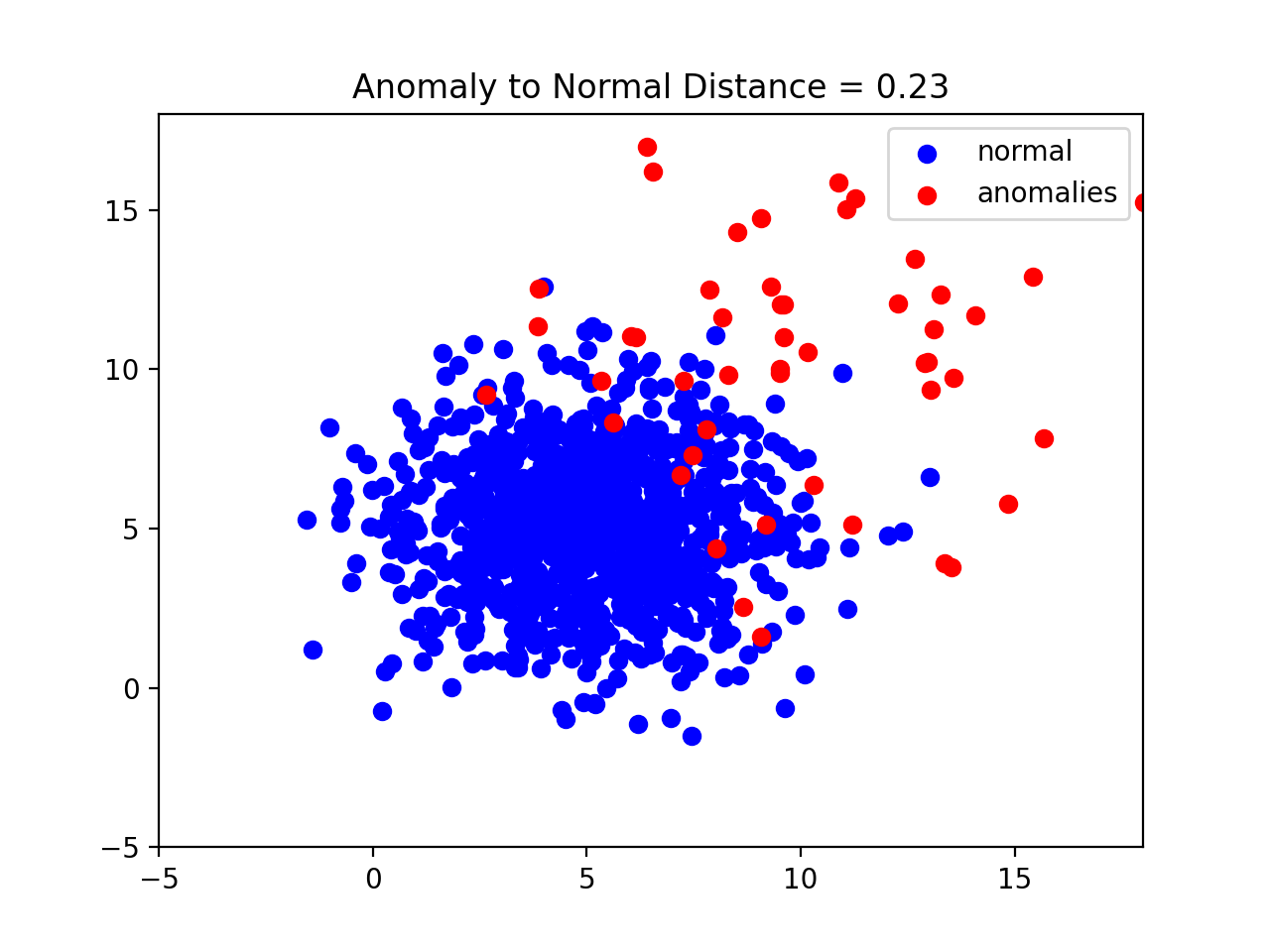}
  \end{minipage}
  \begin{minipage}[b]{0.5\linewidth}
    \centering
    \includegraphics[width=1\linewidth]{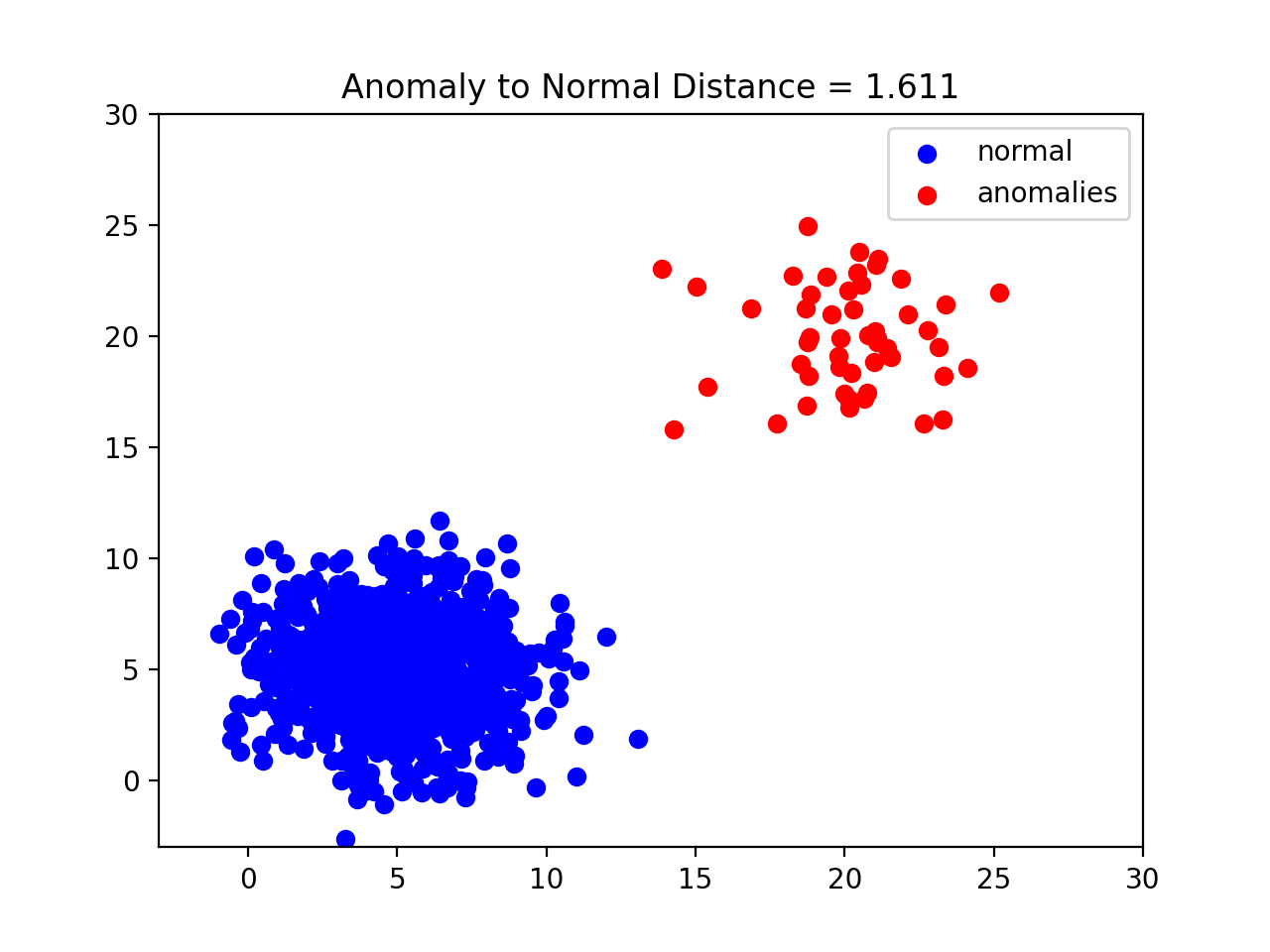}
  \end{minipage} 
     \caption{High Anomaly to Normal Distance (AND) value  indicates readily separated anomalies.}
     \label{fig:and_example}
\end{figure}

In the sequel, we detail the AND meta-feature introduced in this work:
\vspace{-.3cm}
\begin{equation}
\mathit{DistCM} = \frac{\sum_{i=0}^{n}{\mathit{med(N_i)} - \mathit{med(A_i)}}}{n}\\
\end{equation}
\vspace{-.3cm}
\begin{equation}
\mathit{AND} = \frac{\mathit{DistCM - AvgMAD}}{\mathit{AvgMed}}, 
\label{eq:madvsdist}
\end{equation}
where n is the number of features, $\mathit{AvgMAD}$ is the average features' Median Absolute Deviation, and the $\mathit{AvgMed}$ is the average features' median value. Equation~\ref{eq:madvsdist} reveals the normalized difference between the average MAD and the distance between the normal and abnormal center of mass. The intuition behind this formula is how far Anomalies lie from Normal samples, w.r.t to the average Median Absolute Deviation of the distribution. As we can see in Figure~\ref{fig:and_example}, the lower the value of $\mathit{AND}$ is, the more difficult it is to separate abnormal from normal samples. The above meta-features can be distinguished as \emph{Predictive} or \emph{Explanatory}. The former can be computed from the dataset without knowing the target (e.g., dataset dimensionality) while the latter require knowledge of the target (e.g., anomaly ratio). 

\subsection{Evaluation protocols and metrics}
\label{sec:evaluationmetrics} 

We are interested in assessing the effectiveness of anomaly detectors in separating abnormal from normal samples on the basis of the real-valued scores they produce. Unlike offline detectors working with one large window, online detectors incrementally model and score samples in a stream in several windows. 

To ensure a common ground for comparison between detectors’ effectiveness across all datasets of our testbed, we rely on widely used metrics for evaluating the detection of top-$k$ point anomalies \cite{best_offline_3,stare,rrcf,xstream,loda}. Range-based metrics capturing the positional overlap of the discovered anomalous subsequences with the ground-truth \cite{Tatbul2018} are valuable when evaluating shallow \cite{blazquez2021} or deep \cite{Pang2021} anomaly detection methods over time series and it is left as future work.

We consider an evaluation protocol inspired from \textit{Forward Chaining Cross Validation} \cite{rolling_cv} used in time series analytics. It essentially computes an evaluation metric per \emph{window} in the test partition. This evaluation protocol captures the variance of detectors' effectiveness across all windows as their model gets updated in a streaming fashion. The effectiveness of score-based detectors can be assessed by two metrics; Area Under the Receiver Operating Characteristics Curve (ROC AUC)  and Average Precision (AP) \cite{ap}. 

\noindent
AUC is a 2D plot representing the tradeoff between the false positive rate \textit{FPR} (in x-axis) and true positive rate \textit{TPR} (in y-axis), for different score thresholds. The higher the AUC ROC the greater the probability of a detector to classify correctly the samples to abnomal and normal classes. The 0.5 value indicates a random classification and 1.0 value indicates perfect classification.

AP is used to measure explicitly whether anomalies obtain a higher score than normal samples. The higher the AP the less overlap exist between the score distributions of abnomal and normal classes. The 0.0 (1.0) value indicates that all normal samples (anomalies) scored higher than true anomalies (normal). In contrast to AUC, the value that indicates random classification varies per dataset. In fact, it is equal to the percentage of the positive class \cite{ap_more_informative_1}, which in our case is the anomaly ratio of each dataset. More formally, AP is defined using Precision at $k$ ($P@k$) as follows:

\begin{itemize}
    \item \textbf{Precision at $k$} ($P@k$): given a dataset D consisting of n items and a set of anomalies $A \subset D$, P@k is defined as the proportion of the true anomalies to the top $k$ potential anomalies ranked by the detection method:
    $P@k = \frac{\vert\{x \in A\vert rank(x) \leq k\}\vert}{k}.$
    \item \textbf{Average precision} (AP): instead of evaluating the precision individually, this measurement refers to the mean of precision scores over all possible positions:
     $\frac{1}{\vert A\vert} \sum_{k=1}^{n} P@k \cdot \mathbbm{1}[x_k \in A].$
\end{itemize}

\noindent
Given that our stream generation method stratifies anomalies per windows, we set for online detectors $k$ equal to the window size and for offline ones $k$ is equal to the total number of samples. Statistical methods could be alternatively used to estimate $k$ \cite{YangRF19}. In this way, we guarantee that each window contains at least one ground truth anomaly (i.e., $A \neq \emptyset$ in every window). We should stress that the final AUC and AP given an online anomaly detector are computed on each window; thus the final metric value is the average computed over the constituent windows, leading to the Mean Average Precision (\emph{MAP}).

\begin{figure}[!t] 
    \centering
    \includegraphics[width=1\linewidth]{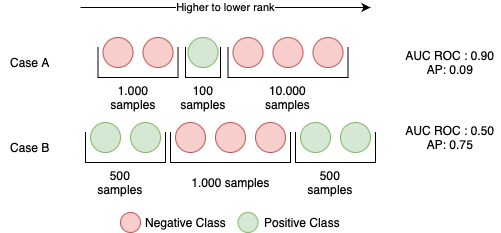}
    \caption{Exemplifying the differences of detectors' scoring functions in AUC ROC vs AP.
    }
    \label{fig:metrics_example}
\end{figure}

\noindent
The two metrics may yield different results given a certain order of samples. In order to demonstrate two extreme cases, we consider the example of Figure~\ref{fig:metrics_example}. Case A, exemplifies a low AP despite having high AUC ROC. When the majority of the normal samples (negative class) are scored in the lowest positions, AUC penalizes less the detector produced such order than AP. Case B exemplifies the opposite case: there is a higher confidence in the positive class on the higher rankings. AP can be more enlightening, as we are interested on ranking anomalies higher than normal samples given that AUC ROC on highly imbalanced datasets (i.e., with low anomaly ratio) can be misleading \cite{ap_more_informative_1, ap_more_informative_2,aucroc_value_loss_1,aucroc_value_loss_3}.

In order to rank the detectors according to their MAP or AUC scores, we used the AutoRank library in python \cite{autorank}, which also statistically compares the ranks of each detector using the non-parametric Friedman test, as well as the post-hoc Nemenyi test for paired comparison. 

\section{Effectiveness of Detectors} 
\label{chapter:effectivness}

The objective of this series of experiments is to answer three main questions: (a) \emph{what is the reliability of decisions made by anomaly detectors compared to a random classifier?} (b) \emph{how online and offline detectors could be ranked according to their performance on real datasets?} and (c) \emph{to what extend online detectors approximate the performance of offline detectors in identifying sub/full space anomalies?} 

To answer these questions, we benchmark ten online detectors \{HST, HSTF, RRCF, MCOD, XSTREAM (X-S), LODA (L-S), RS-HASH, STARE, LEAP, CPOD\} and six offline detectors \{LOF, KNN$_W$, iForest (IF), XSTREAM (X-B), OCRF, LODA (L-B)\} using twenty-four real datasets (see Section~\ref{datasets}). The AUC and MAP scores of online and offline detectors per dataset are reported in Tables~\ref{table:auc_scores} and \ref{table:map_scores}, respectively (see Appendix~\ref{sec:aucmapscores}). These scores are obtained by averaging over 30 independent runs (with different sampled data streams) for HST/F, RRCF, X-S, and L-S (non deterministic detectors) and 1 run for MCOD, RS-HASH, STARE, LEAP, CPOD (deterministic detectors). For the optimal hyper-parameters of each detector per dataset readers are referred to Appendix~\ref{app:hyperparameters}. After preliminary experiments with a subset of our datasets, we chose 128 as the optimal window size for the vast majority of the detectors (only RS-HASH is favored by larger sized windows see Figure~\ref{fig:window_sizes}). Varying window sizes per algorithm and dataset will complicate a statistical sound comparison of algorithms over the two performance metrics (MAP and AUC) across all datasets of our testbed. In the sequel, we will present a synthetic overview of the main findings after analysing the AUC and MAP scores.

\subsection{Random Detection}
\label{sec:random}

\begin{figure}[!t]
    \centering
    \includegraphics[width=1.1\linewidth]{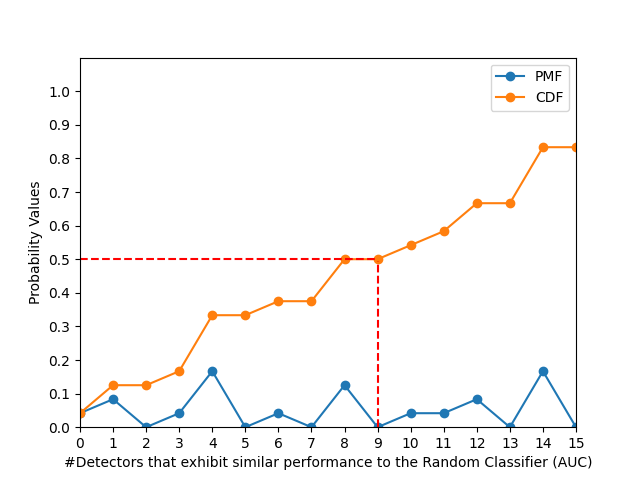}
    \caption{CDF and PMF of online and offline detectors with a performance comparable to the Random Classifier: More than a half of the detectors fail in 50\% of the datasets.}
    \label{fig:cdf}
\end{figure}

In a first step, we are interested in \emph{assessing the reliability of the performance exhibited by the different online and offline detectors in real datasets}. To this end, we compare their AUC scores in real datasets (see Table \ref{table:auc_scores}) with the performance of a \emph{Random Classifier} described in Section~\ref{sec:evaluationmetrics}. In particular, we are looking for dataset characteristics that make detectors to perform randomly. Figure~\ref{fig:cdf} depicts the CDF and PMF plots of the distribution of the number of detectors that exhibit similar performance to the Random Classifier per dataset  i.e., with AUC ROC $ < 0.6$. The AUC ROC essentially represents the probability of a random anomaly to be scored higher than a normal sample, thus it provides a natural comparison with the performance of the random binary classifier: when AUC=0.5, then the classifier is not able to distinguish between positive and negative class points.
As we can see, 9 out of 16 anomaly detectors are close to the Random Classifier in half of the datasets (see the dotted red line). This observation clearly indicates the serious limitations of existing unsupervised detectors. Interestingly, on datasets that contain implanted anomalies, 11.3 detectors fail to outperform on average the random classifier whereas on datasets where the minority class is used as the anomaly class, 7.6 detectors fail on average. This highlights that implanted anomalies are more challenging for the detectors than anomalies simulated by the minority class. This is due to the fact that in implanted datasets, anomalies are scattered in the full feature space and their detection requires knowledge of the data distribution drawn from the entire dataset rather than from individual windows. As we can observe in Table \ref{table:map_scores} in Appendix B, batch detectors outperform streaming detectors in five out of six dataset with implanted anomalies. On the only dataset with real anomalies (Electricity), 12 detectors fail but this is mainly attributed to the high anomaly ratio of this dataset (57.5\%). In fact, the number of detectors that a exhibit random performance is correlated, with a \emph{pvalue of 0.039} and a \emph{correlation value of 0.422}, with the anomaly ratio of the datasets. Consequently, the higher the anomaly ratio, the more likely it is for the detectors to behave randomly. LOF and KNN are the detectors that fail in the least amount of datasets compared to the rest (7 and 8 datasets respectively). Among online detectors X-S and MCOD have the least random behavior (both fail on 11 datasets out of 24). It is also worth mentioning that no algorithmic family seem to fail in more datasets compared to the rest.

To capture the degree of difficulty caused by the different types of anomalies contaminating each dataset, we consider the meta-feature \emph{Anomaly To Normal Distance} (AND) described in Section~\ref{sec:metafeatures}. Over the 14 datasets with the lowest \emph{AND} value, 11.3 detectors fail on average. Over the remaining 10 datasets with the highest \emph{AND} value, 6.2 detectors fail on average. Clearly, the higher the distance of abnormal from normal samples w.r.t. the average mean absolute deviation, the greater the number of detectors scoring close to the Random Classifier. Another meta-feature significantly correlated with the number of detectors exhibiting a performance close the Random Classifier, is the number of distinct highly \emph{correlated pairs of features}, with a {\emph{pvalue = 0.005} and \emph{correlation value of -0.546}}. The more correlated pairs of features in a dataset, the less detectors perform similarly to the Random Classifier.

\subsection{Ranking Online detectors}
\label{sec:online_ranking}

We are turning now our attention to the \emph{performance comparison of online anomaly detectors}. 

Table~\ref{table:online_ap} summarizes the total number of wins of online detectors along with the average difference of its MAP (AUC ROC) score from the wining detector in each dataset. According to MAP metric, four detectors (X-S, MCOD, L-S and RS-HASH) tie in the first place winning in 4 out of 24 datasets respectively. LEAP follows, with 3 wins while HST and HSTF have 2 wins each. CPOD and STARE do not achieve any win. HST, in datasets outperformed by others, exhibits a performance very close to the leader (compared to the rest of the detectors) with a  difference of 15.8\% on average. Despite its forgetting mechanism, HSTF have a higher average difference from the leader (18.7\%) compare to HST, while having the same amount of wins. RS-Hash, STARE and LEAP win all together in 29\% of the datasets, but they exhibit the highest average difference (22.5\% - 28.8\%) from leader in the remaining ones. Note that 3 of the X-S wins are observed in datasets (3 out of 6) with a number of samples lower than 800. Also note that, projection-based detectors (X-S and L-S) lead in 4 out of 6 datasets that contain implanted anomalies (MCOD wins in the rest).

\begin{table}[!t]
\centering
\caption{Number of online detectors' wins and average difference from the winner (ADW) on real datasets: X-S dominates in both metrics}
\label{table:online_ap}
\begin{tabular}{|l|c|c||c|c|}
\hline
& \multicolumn{2}{c||}{MAP}  & \multicolumn{2}{c|}{AUC ROC}  \\ \hline
\textbf{Detector} & \textbf{\#Wins} & \textbf{ADW} & \textbf{\#Wins} & \textbf{ADW} \\ \hline
\textbf{X-S}   & 4 & 19.2\% & 9 & 3.6\% \\ \hline
\textbf{HST}  & 2  & 15.8\%  & 2  & 7.9\% \\ \hline
\textbf{HSTF}  & 2  & 18.7\% & 2  & 8.2\% \\ \hline
\textbf{RRCF}  & 1  & 20.2\% & 1  & 11.8\% \\ \hline
\textbf{MCOD}  & 4  & 19.5\% & 2  & 7.9\% \\ \hline
\textbf{L-S}  & 4  & 18.8\%  & 1  & 9.3\% \\ \hline
{\textbf{RS-Hash}}  & 4  & 22.5\%  & 4  & 9.5\% \\ \hline
{\textbf{STARE}}  & 0  & 28.8\%  & 1  & 14.3\% \\ \hline
{\textbf{LEAP}}  & 3 & 24.1\%  & 2  & 14.2\% \\ \hline
{\textbf{CPOD}}  & 0  & 25.7\%  & 0  & 14.4\% \\ \hline
\end{tabular}
\end{table}

Table~\ref{table:online_ap} also summarizes the total number of wins of each online detector based on the AUC ROC scores. According to this metric, X-S gets 5 more wins while L-S and MCOD have 3 and 2 less wins respectively compared to MAP. STARE obtains its first win compared to MAP. We observe that X-S achieves 4 more wins in datasets where most detectors exhibit similar performance to the Random Classifier. In this case, the performance difference between the first and the second leading detector becomes significant compared to MAP. Also, X-S wins in 5 out of 8 datasets with number of samples greater than 10.000. In other datasets, we observe that according to AUC ROC detectors may not perform so well as when considering MAP (see Section~\ref{sec:evaluationmetrics}). More precisely, L-S on InternetAds and X-S on Pima succeed to score more anomalies higher than the other detectors, but failed on the lower rankings.

\begin{figure}[!t]
    \subfloat[MAP]{
        \includegraphics[width=1\linewidth]{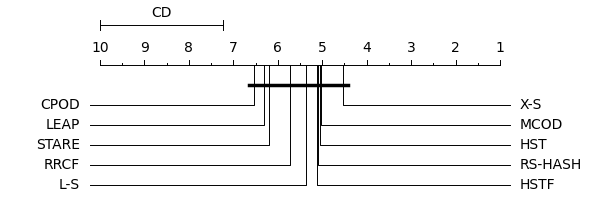}
        \label{fig:ap_online_rank}
    }
    \hfill
    \subfloat[AUC ROC]{
        \includegraphics[width=1\linewidth]{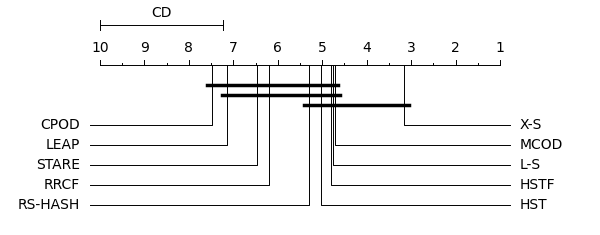}
        \label{fig:auc_online_rank}
    }
    \caption{{Online detectors' ranking: X-S dominates on both MAP and AUC.}}
    \label{fig:online_ranking}
\end{figure}

To determine if there are any significant difference between the average ranks of the detectors, we use the non parametric Friedman test  \cite{friedman}. We reject the null hypothesis with a significance level of 5\% that all detectors' performances are equal. Next, we use the post-hoc Nemenyi test, in order to compare the detectors in pairs. There is a significant difference when the difference between the average ranks of two detectors is higher than a critical distance (CD) of 2.57 (for 10 detectors on 24 datasets at a significance level of 0.05). Figure \ref{fig:ap_online_rank} depicts the statistically significant ranking of detectors' performance in the 24 datasets of our testbed according to their MAP. X-S is ranked first, followed by MCOD, while having not statistically significant difference with any of the remaining detectors. HST/F, MCOD and RS-HASH have similar average ranks (around 5). HST surpasses HSTF due to the lower average performance difference from the leader, despite having the same number of wins. L-S is ranked sixth followed by RRCF. STARE and CPOD and LEAP are the last three detectors with very little differences in their average ranks.

This overall performance picture does not drastically change when we consider AUC ROC. As we can see in Figure \ref{fig:auc_online_rank}, X-S now has a statistical significant difference with the last 4 detectors (CPOD, LEAP, STARE, RRCF). X-S's great performance on AUC ROC, indicates that it consistently ranks most of normal samples lower than anomalies. The following 5 detectors (L-S, HST/F, MCOD and RS-HASH) are ranked close, which can be mainly attributed to the class imbalance of the datasets, as AUC ROC provides less information compared to MAP (see Section~\ref{sec:evaluationmetrics}). L-S is now ranked three places higher compared to MAP, as it becomes the third best performing detector leaving RS-HASH in the sixth place.

In a nutshell, X-S exhibits the best overall performance when considering both metrics, having a statistical significant difference with half of the detectors using AUC ROC. L-S is performing better w.r.t. AUC ROC rather than MAP, while the additional forgetting mechanism on HST, does not significantly boost its performance.  The rest of the detectors remain relatively stable w.r.t. both metrics.
We should note that the optimized variants of MCOD like CPOD or LEAP are systematically ranked in the last 2 positions w.r.t. both metrics. Regarding density-based detectors, RS-Hash outperforms STARE but both under-perform compared to the distance-based MCOD.

\subsection{Online vs Offline detectors}
\label{sec:onlineVsOffline}
As a last question in this section we are studying to \emph{what extent online detectors approximate the performance of  offline detectors in real datasets}. In essence, we are interested in comparing the AUC or AP performance of detection models continuously updated in several small windows with batch models build in one big window.

Again we rely on the non parametric Friedman test \cite{friedman} and we reject the null hypothesis that all detectors' performances are equal with a significance level of 5\%. Using the post-hoc Nemenyi test, there is a significant difference when the difference between the average ranks of two detectors is higher than a critical distance (CD) of 4.05 (for 16 detectors on 24 datasets at a significance level of 0.05). Figure \ref{fig:all_ap_ranking} depicts the rank of each detector according to their MAP scores, and Table \ref{onlineVSoffline}
provides the number of wins per detector along with the average difference from the leader detector in each dataset. 

In the following, we are contrasting the rank of online and offline detectors belonging to the same family, namely, tree-based (HST/F and RRCF vs IF and OCRF) or distance/KNN-based (MCOD, CPOD and LEAP vs KNN), density-based  (STARE, and RS-HASH vs LOF), as well as, the stream and batch versions of the two projection-based XSTREAM and LODA.

\subsubsection{Tree based detectors} 
\begin{figure}[!t]
    \subfloat[MAP]{
            \includegraphics[width=1\linewidth]{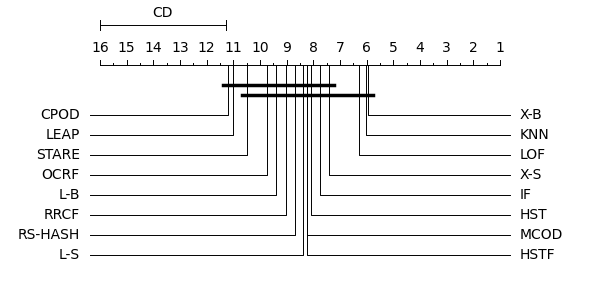}
        \label{fig:all_ap_ranking}
    }
    \hfill
    \subfloat[AUC ROC]{
        \includegraphics[width=1\linewidth]{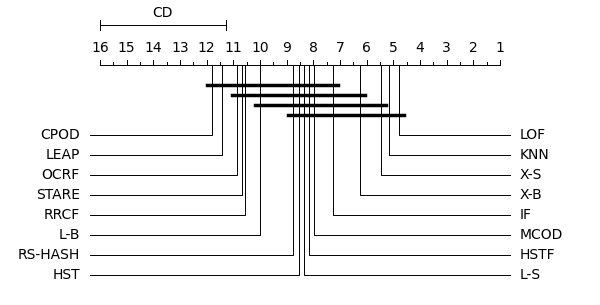}
        \label{fig:all_auc_ranking}
    }
    \caption{{Detectors' ranking using MAP and AUC ROC:  Online detectors seemingly approximate the performance of offline detectors.}}
    \label{fig:detectors_ranking}
\end{figure}

As we can see in Figure \ref{fig:all_ap_ranking}, IF is the best ranked tree-based detector, and fifth overall. Although IF is ranked first, the rank difference between HST(F) and RRCF are less than 0.5 and 1.5 respectively. Surprisingly enough, tree based online detectors not only approximate well the effectiveness of offline ones, but as in the cases of HST(F) and RRCF may outperform offline detectors like OCRF. Despite the most enlighten decision regarding the splitting criteria in trees, OCRF exhibits the worst performance. This behavior can attributed to the fact that to ensure a fair comparison with online detectors, offline detectors are also trained with both normal samples and anomalies (unlike the original OCRF paper \cite{ocrf}).

It is important to notice, that there is no statistically significant difference between any of the tree-based detectors. This picture does not change based on AUC ROC as well. As we can see in Table \ref{onlineVSoffline}, IF, OCRF and HST succeed to get the most wins among tree-based detectors (2 wins each) while HST exhibits the lowest average difference from leader. The rest of the detectors (HSTF, IF, RRCF) have a similar average difference from leader, with OCRF having the highest at (31.4\%).

\subsubsection{Distance/KNN and Density based detectors}
{
As we can see in Figure \ref{fig:detectors_ranking}, offline detectors KNN and LOF are ranked in the first 3 places based on both AUC ROC and MAP. Their rank difference with MCOD is close to 2. As a matter of fact, MCOD and LOF achieve the highest number of wins with 4 respectively, followed by LEAP and KNN with 2 and 1 wins (see Table \ref{onlineVSoffline}). After MCOD, RS-HASH is the second best ranked between online detector while CPOD, STARE and LEAP are consistently ranked in the 3-4 lower places achieving on average the highest performance difference from the leader detector in each dataset (see Table\ref{onlineVSoffline}). Furthermore, RS-Hash and MCOD fail ($AUC < 0.6$) in less datasets compared to CPOD,LEAP,STARE. KNN and LOF show a statistically important difference to CPOD, LEAP and STARE based on AUC ROC and to CPOD and LEAP based on MAP. It is worth mentioning that LOF wins in all three datasets with highest number of features (isolet, letter-recognition, InternetAds) with an AUC above 0.75 while on the other hand every online KNN and Density based detector performs randomly.}

\subsubsection{Projection based detectors}
{
As we can see in Figure \ref{fig:all_ap_ranking}, the batch version of XSTREAM (X-B), is the best performing detector whilst its streaming version is placed fourth. There is no statistically significant difference between the performance of the two XSTREAM versions (i.e., Their rank difference is close to 1). X-B succeeds to win in 2 datasets while X-S does not achieve any win and has 2.5\% less average difference from the corresponding leader (see Table \ref{onlineVSoffline}). Interestingly enough, X-S is ranked third and X-B fourth based on AUC ROC (Figure \ref{fig:all_auc_ranking}). In a nutshell, X-S not only approximates well the performance of X-B (and outperforms it according to AUC ROC), but also outperforms almost every other detector in our benchmark (besides LOF and KNN).}

{As we can see in Figure \ref{fig:all_ap_ranking}, the streaming (L-S) and batch (L-B) versions of LODA are ranked closely (only RRCF separates them). L-S wins in 4 datasets with a lower average difference (28.0\%) from the leader while L-B is systematically outperformed with highest average difference (32.1\%) from the leader in each dataset (see Table \ref{onlineVSoffline}).} Clearly, L-S outperforms L-B w.r.t. both metrics.

\subsubsection{Discussion}
{
The previous experiments demonstrate that online detectors can effectively approximate the performance of offline ones (e.g., X-S, MCOD and RS-HASH vs KNN and LOF or HST/F vs IF) and in some metrics to outperform them (i.e., X-S vs X-B). Distance/KNN and density-based offline detectors are ranked high (top 3), while the optimized MCOD variations like CPOD and LEAP are ranked in the lowest places. Furthermore, projection-based detectors (XSTREAM and LODA) consistently outperform tree-based ones (HST/F, RRCF and OCRF) in either batch or stream modes.}

\begin{table}[!t]
\caption{{Number of wins based on MAP of online vs offline detectors on real datasets and their average difference from winner (ADW): LOF achieves the most wins.}}
\label{onlineVSoffline}
\scalebox{0.75}{
\begin{tabular}{lcccccc}

\hline
\multicolumn{1}{|l|}{Detector} & \multicolumn{1}{c|}{IF}      & \multicolumn{1}{c|}{LOF}    & \multicolumn{1}{c|}{KNN}    & \multicolumn{1}{c|}{X-B}    & \multicolumn{1}{c|}{OCRF}   & \multicolumn{1}{c|}{L-B}    \\ \hline
\multicolumn{1}{|l|}{\#Wins}   & \multicolumn{1}{c|}{2}       & \multicolumn{1}{c|}{4}      & \multicolumn{1}{c|}{1}      & \multicolumn{1}{c|}{2}      & \multicolumn{1}{c|}{2}      & \multicolumn{1}{c|}{0}      \\ \hline
\multicolumn{1}{|l|}{ADW}      & \multicolumn{1}{c|}{26.7\%}  & \multicolumn{1}{c|}{24.4\%} & \multicolumn{1}{c|}{24.4\%} & \multicolumn{1}{c|}{25.8\%} & \multicolumn{1}{c|}{31.4\%} & \multicolumn{1}{c|}{32.1\%} \\ \hline

\multicolumn{7}{c}{Online detectors}                                                                                                                                                                                \\ \hline
\multicolumn{1}{|l|}{Detector} & \multicolumn{1}{c|}{RS-HASH} & \multicolumn{1}{c|}{STARE}  & \multicolumn{1}{c|}{LEAP}   & \multicolumn{1}{c|}{CPOD}   & \multicolumn{1}{c|}{X-S}    & \multicolumn{1}{c|}{HST}    \\ \hline
\multicolumn{1}{|l|}{\#Wins}   & \multicolumn{1}{c|}{0}       & \multicolumn{1}{c|}{0}      & \multicolumn{1}{c|}{2}      & \multicolumn{1}{c|}{0}      & \multicolumn{1}{c|}{0}      & \multicolumn{1}{c|}{2}      \\ \hline
\multicolumn{1}{|l|}{ADW}      & \multicolumn{1}{c|}{31.6\%}  & \multicolumn{1}{c|}{38.0\%} & \multicolumn{1}{c|}{33.3\%} & \multicolumn{1}{c|}{34.9\%} & \multicolumn{1}{c|}{28.3\%} & \multicolumn{1}{c|}{24.9\%} \\ \hline

\multicolumn{7}{c}{Online Detectors}                                                                                                                                                                                \\ \hline
\multicolumn{1}{|l|}{Detector} & \multicolumn{2}{c|}{HSTF}                                  & \multicolumn{1}{c|}{RRCF}   & \multicolumn{1}{c|}{MCOD}   & \multicolumn{2}{c|}{L-S}                                  \\ \hline
\multicolumn{1}{|l|}{\#Wins}   & \multicolumn{2}{c|}{1}                                     & \multicolumn{1}{c|}{0}      & \multicolumn{1}{c|}{4}      & \multicolumn{2}{c|}{4}                                    \\ \hline
\multicolumn{1}{|l|}{ADW}      & \multicolumn{2}{c|}{27.8\%}                                & \multicolumn{1}{c|}{29.3\%} & \multicolumn{1}{c|}{28.6\%} & \multicolumn{2}{c|}{28.0\%}     
\\ \hline

\end{tabular}
}
\end{table}

\section{Detection of Range-based Anomalies}
\label{chapter:range-based}
In this section we evaluate the performance of online detectors (X-S, L-S, HST/F, RRCF, MCOD, RS-HASH, LEAP, CPOD, STARE) over 5 time-series datasets containing a different type of range-based anomalies (see Section~\ref{sec:exathlon}): (a)  $1\_2\_100000\_68$ contains anomalies caused by bursty input until crash; (b) $1\_4\_1000000\_80$ contains anomalies caused by CPU contention; (c) $1\_5\_1000000\_86$ contains anomalies caused by a process failure; (d) $6\_1\_500000\_65$ contains anomalies caused by bursty input without resulting in a crash; and (e) $6\_3\_200000\_76$ contains anomalies caused by stalled input from the sender. As in our previous experiments, we thoroughly tuned the hyper-parameters (see Table~\ref{table:tuned_hyperparameters_exathlon} in Appendix~\ref{app:hyperparameters}) of each detector. 

For range-based anomalies, the selection of window size is more challenging than for point anomalies. This is because the anomaly ranges (and thus the number of anomalies) vary from one window to the other per dataset. In other words, one cannot stratify the datasets contaminated with range-based anomalies as the order of the samples matters. For this reason, we included the window size as a hyper-parameter of each detection algorithm in our cross-validation protocol. In Table \ref{table:tuned_hyperparameters_exathlon} (see Appendix), we report the window size that led to the highest AUC for each detector with a performance higher than the random classifier.
as well as, the window size (64,128,256) per dataset.

\begin{figure*}[!t]%
 \centering
 \subfloat[]{\includegraphics[scale=0.4]{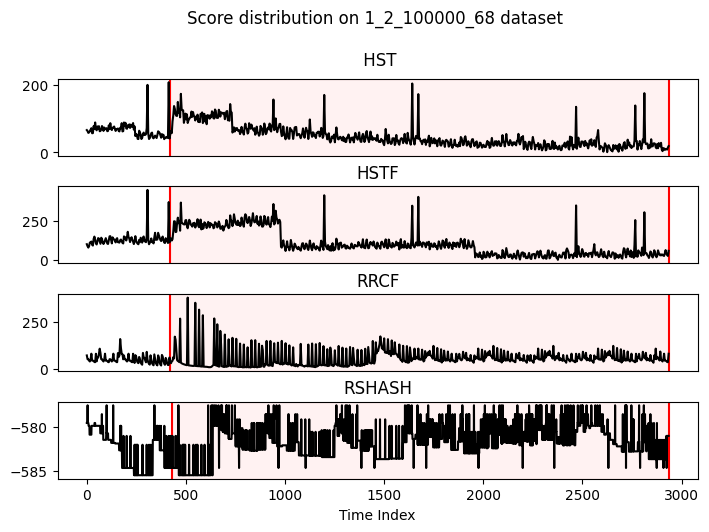}\label{fig:dist_ex12}}%
 \subfloat[]{\includegraphics[scale=0.4]{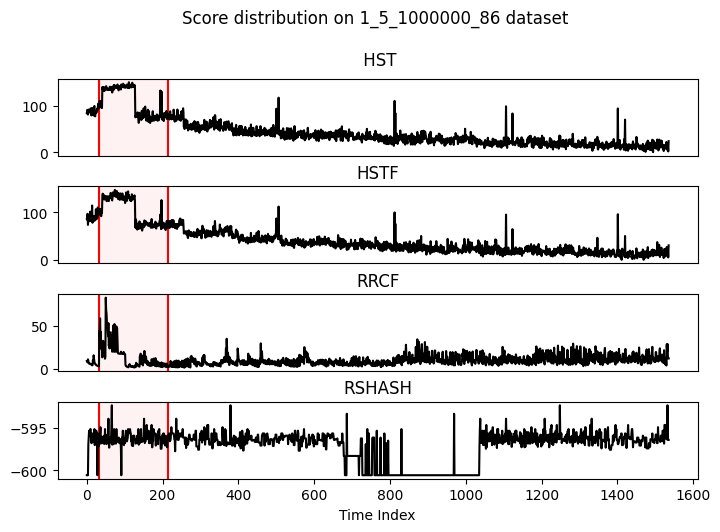}\label{fig:dist_ex15}}\\
 \subfloat[]{\includegraphics[scale=0.4]{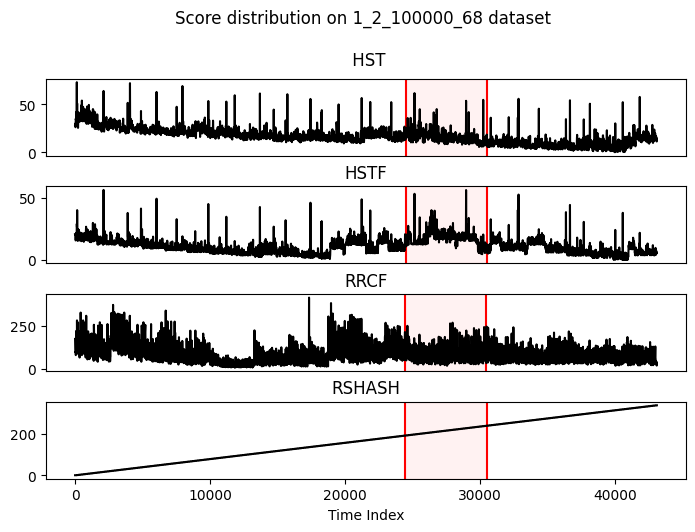}\label{fig:dist_ex14}}%
  \subfloat[]{\includegraphics[scale=0.4]{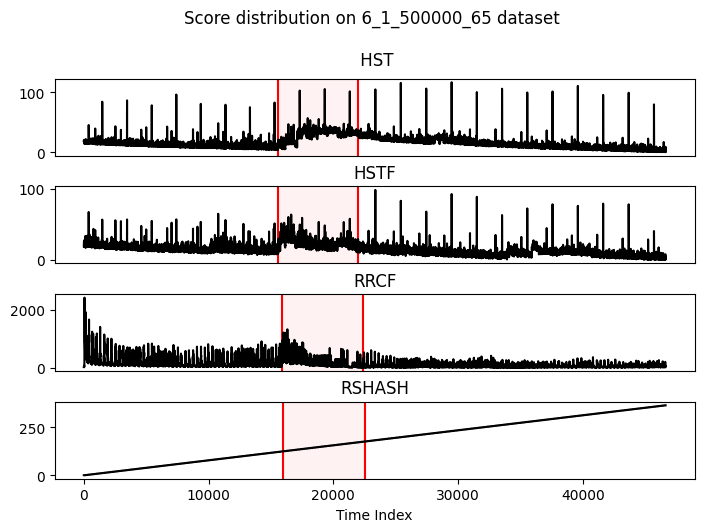}\label{fig:dist_ex61}}\\
   \subfloat[]{\includegraphics[scale=0.4]{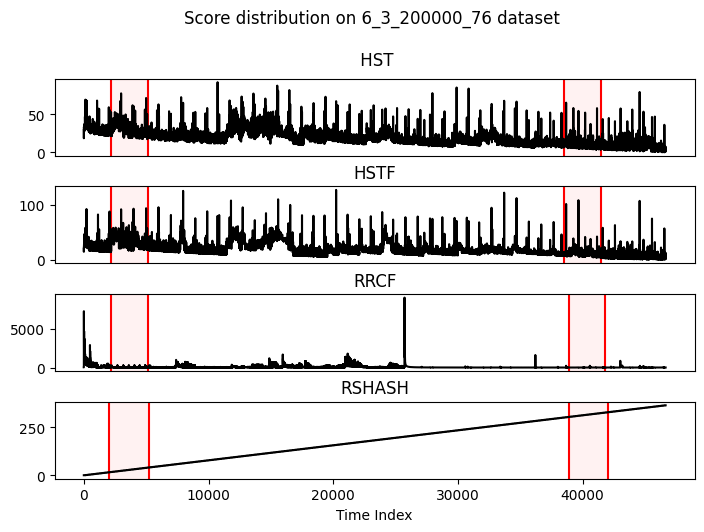}\label{fig:dist_ex63}}
 \caption{Score distribution of HST/F, RRCF and RS-HASH for the 5 datasets from Exathlon benchmark. The red highlighted areas indicate the the anomaly range according to the ground truth.}%
 \label{fig:exathlon_distributions}%
\end{figure*}

Table~\ref{table:exathlon_scores} depicts the MAP and the AUC ROC scores of online detectors in the aforementioned datasets. We can easily observe that the majority of detectors (X-S, L-S, MCOD, CPOD, LEAP, STARE) exhibit a random behavior in all 5 datasets. For this reason, we will focus on the remaining 4 detectors (HST/F, RRCF, RS-HASH). Notably, HSTF and RRCF are the only two detectors that are exploring a \emph{tunable} forgetting mechanism as new samples appear. HST/F outperforms all other detectors in the 5 datasets, while exhibits a close to random performance on $6\_3\_200000\_76$  dataset. The addition of the forgetting mechanism on HST slightly improves its performance in 4 out of 5 datasets as it is able to forget past points for detecting range-based anomalies. We should stress that RS-HASH is more effective than X-S due to its forgetting policy, i.e., it \emph{gradually} reduces the histogram counts rather than assigning zero counts to all bins when the window slides, as in X-S. \emph{Tunable} and \emph{gradual} forgetting mechanisms prove to be more effective for detecting range-based anomalies in several time-frames. Moreover, the range-based anomalies of Exathlon are clustered in the specific time-frames; thus, after few frames, a global forgetting mechanism adjusts to the anomalous samples considering them as normal. The dataset containing anomalies that are caused by CPU contention (6\_3\_2000000\_76), proves to be the most difficult for all these 4 detectors, where all detectors exhibit a random performance. As a matter of fact, this is the dataset containing two different interval anomalies while the application returns back to its normal state after a period of an abnormal behavior. On the other hand, all 4 detectors achieve the highest scores on 1\_2\_100000\_68 dataset, where anomalies are caused by an increase of the input rate of the sender, until the application crashed.

\begin{table*}[!t]
\centering
\caption{MAP and AUC ROC scores of online detectors over the 5 time-series datasets (1\_4\_1000000\_80, 1\_5\_1000000\_86, 1\_2\_100000\_68, 6\_1\_500000\_65, 6\_3\_200000\_76) contained in Exathlon benchmark.}
\label{table:exathlon_scores}
\scalebox{0.6}{
\begin{tabular}{lccccc}
\hline
\multicolumn{1}{|l|}{{\textbf{DATASET}}} & \multicolumn{1}{l|}{\textbf{1\_4\_1000000\_80}} & \multicolumn{1}{l|}{\textbf{1\_2\_100000\_68}} & \multicolumn{1}{l|}{\textbf{1\_5\_1000000\_86}} &
\multicolumn{1}{l|}{\textbf{6\_1\_500000\_65}} &
\multicolumn{1}{l|}{\textbf{6\_3\_200000\_76}} \\ \hline
\multicolumn{6}{c}{\textit{\textbf{MAP}}}                                                                                                                                                            \\ \hline
\multicolumn{1}{|l|}{\textbf{HST}}              & \multicolumn{1}{c|}{0.2695}                     & \multicolumn{1}{c|}{0.8264}                     & \multicolumn{1}{c|}{0.9997}               & \multicolumn{1}{c|}{0.6708}     & \multicolumn{1}{c|}{0.1671} \\ \hline
\multicolumn{1}{|l|}{\textbf{HSTF}}             & \multicolumn{1}{c|}{0.2813}                     & \multicolumn{1}{c|}{0.8038}                     & \multicolumn{1}{c|}{0.9998}                 & \multicolumn{1}{c|}{0.5282}   & \multicolumn{1}{c|}{0.1680} \\ \hline
\multicolumn{1}{|l|}{\textbf{RRCF}}             & \multicolumn{1}{c|}{0.1007}                     & \multicolumn{1}{c|}{0.1880}                     & \multicolumn{1}{c|}{0.9989}                  & \multicolumn{1}{c|}{0.2810}  & \multicolumn{1}{c|}{0.0936} \\ \hline
\multicolumn{1}{|l|}{\textbf{X-S}}              & \multicolumn{1}{c|}{0.1283}                     & \multicolumn{1}{c|}{0.0506}                     & \multicolumn{1}{c|}{0.7824}                & \multicolumn{1}{c|}{0.5000}    & \multicolumn{1}{c|}{0.0936} \\ \hline
\multicolumn{1}{|l|}{\textbf{L-S}}              & \multicolumn{1}{c|}{0.1283}                     & \multicolumn{1}{c|}{0.0506}                     & \multicolumn{1}{c|}{0.7824}                & \multicolumn{1}{c|}{0.5000}    & \multicolumn{1}{c|}{0.0936} \\ \hline
\multicolumn{1}{|l|}{\textbf{MCOD}}             & \multicolumn{1}{c|}{0.1283}                     & \multicolumn{1}{c|}{0.0506}                     & \multicolumn{1}{c|}{0.7881}                 & \multicolumn{1}{c|}{0.5000}   & \multicolumn{1}{c|}{0.0936} \\ \hline
\multicolumn{1}{|l|}{\textbf{CPOD}}             & \multicolumn{1}{c|}{0.1283}                     & \multicolumn{1}{c|}{0.0506}                     & \multicolumn{1}{c|}{0.7881}                 & \multicolumn{1}{c|}{0.5000}   & \multicolumn{1}{c|}{0.0936} \\ \hline
\multicolumn{1}{|l|}{\textbf{LEAP}}             & \multicolumn{1}{c|}{0.1283}                     & \multicolumn{1}{c|}{0.0506}                     & \multicolumn{1}{c|}{0.7881}                  & \multicolumn{1}{c|}{0.5000}  & \multicolumn{1}{c|}{0.0936} \\ \hline
\multicolumn{1}{|l|}{\textbf{STARE}}            & \multicolumn{1}{c|}{0.1283}                     & \multicolumn{1}{c|}{0.0506}                     & \multicolumn{1}{c|}{0.7881}                  & \multicolumn{1}{c|}{0.5000}  & \multicolumn{1}{c|}{0.0936} \\ \hline
\multicolumn{1}{|l|}{\textbf{RS-HASH}}          & \multicolumn{1}{c|}{0.1283}                     & \multicolumn{1}{c|}{0.10}                       & \multicolumn{1}{c|}{0.8789}            & \multicolumn{1}{c|}{0.5000}  & \multicolumn{1}{c|}{0.0936}     \\ \hline
\multicolumn{6}{c}{\textit{\textbf{AUC ROC}}}                                                                                                                                                        \\ \hline
\multicolumn{1}{|l|}{\textbf{HST}}              & \multicolumn{1}{c|}{0.8360}                     & \multicolumn{1}{c|}{0.9839}                     & \multicolumn{1}{c|}{0.9477} &  \multicolumn{1}{c|}{0.8953}     & \multicolumn{1}{c|}{0.5268} \\ \hline
\multicolumn{1}{|l|}{\textbf{HSTF}}             & \multicolumn{1}{c|}{0.8659}                     & \multicolumn{1}{c|}{0.9846}                     & \multicolumn{1}{c|}{0.9754}                 & \multicolumn{1}{c|}{0.8998}   & \multicolumn{1}{c|}{0.5500}  \\ \hline
\multicolumn{1}{|l|}{\textbf{RRCF}}             & \multicolumn{1}{c|}{0.5160}                     & \multicolumn{1}{c|}{0.6764}                     & \multicolumn{1}{c|}{0.8876}               & \multicolumn{1}{c|}{0.7182}  & \multicolumn{1}{c|}{0.5000} \\ \hline
\multicolumn{1}{|l|}{\textbf{X-S}}              & \multicolumn{1}{c|}{0.5000}                     & \multicolumn{1}{c|}{0.5000}                     & \multicolumn{1}{c|}{0.500}                  & \multicolumn{1}{c|}{0.5000} & \multicolumn{1}{c|}{0.5000}  \\ \hline
\multicolumn{1}{|l|}{\textbf{L-S}}              & \multicolumn{1}{c|}{0.5000}                     & \multicolumn{1}{c|}{0.5000}                     & \multicolumn{1}{c|}{0.500}                  & \multicolumn{1}{c|}{0.5000} & \multicolumn{1}{c|}{0.5000} \\ \hline
\multicolumn{1}{|l|}{\textbf{MCOD}}             & \multicolumn{1}{c|}{0.5000}                     & \multicolumn{1}{c|}{0.5000}                     & \multicolumn{1}{c|}{0.5000}               & \multicolumn{1}{c|}{0.5000}     & \multicolumn{1}{c|}{0.5000} \\ \hline
\multicolumn{1}{|l|}{\textbf{CPOD}}             & \multicolumn{1}{c|}{0.5000}                     & \multicolumn{1}{c|}{0.5000}                     & \multicolumn{1}{c|}{0.500}                    & \multicolumn{1}{c|}{0.5000}& \multicolumn{1}{c|}{0.5000}  \\ \hline
\multicolumn{1}{|l|}{\textbf{LEAP}}             & \multicolumn{1}{c|}{0.5000}                     & \multicolumn{1}{c|}{0.5000}                     & \multicolumn{1}{c|}{0.500}                   & \multicolumn{1}{c|}{0.5000}  & \multicolumn{1}{c|}{0.5000} \\ \hline
\multicolumn{1}{|l|}{\textbf{STARE}}            & \multicolumn{1}{c|}{0.5000}                     & \multicolumn{1}{c|}{0.5000}                     & \multicolumn{1}{c|}{0.500}                   & \multicolumn{1}{c|}{0.5000} & \multicolumn{1}{c|}{0.5000}  \\ \hline
\multicolumn{1}{|l|}{\textbf{RS-HASH}}          & \multicolumn{1}{c|}{0.5000}                     & \multicolumn{1}{c|}{0.72}                       & \multicolumn{1}{c|}{0.7376}  & \multicolumn{1}{c|}{0.5000} & \multicolumn{1}{c|}{0.5000}                   \\ \hline
\end{tabular}
}
\end{table*}

Figure~\ref{fig:exathlon_distributions} depicts the scores per sample for each of the 4 detectors (HST/F, RRCF and RS-HASH). RS-HASH has a similar performance across 3 out of 5 datasets where the scores of samples increase as the time passes. As we can see in Figure~\ref{fig:dist_ex12}, HST/F and RRCF are able to highly score anomalies from the beginning of the bursty input where they start to appear. As time passes, the scores of anomalies are getting closer to those of normal samples before the bursty inputs begin. The forgetting mechanism of HST, results in more immediate drop in the scores' range. The same behavior can be observed in the 1\_5\_10000000\_86 dataset where the scores of normal samples, after the anomalies have appeared, are getting smaller over time. Last but not least, datasets 1\_4\_1000000\_80, 6\_1\_500000\_65 and 6\_3\_200000\_76 that do not result in application crash, proves to be the most challenging for all detectors. As shown in Figures~\ref{fig:dist_ex14},~\ref{fig:dist_ex61} and ~\ref{fig:dist_ex63}, RRCF and HST/F are not able to detect the existence of abnormal samples and maintain a high score over a large period of time. This behavior is due to the fact that anomalies appearing in more than 40 consecutive windows, facilitate detectors to adjust to the distribution change. As the application returns to its normal state, normal samples are identified as anomalies until the detectors adjust again to the new distribution change. As a result, data streams processed in high number of consecutive windows featuring anomalies which do not correspond to application crash, prove to be the most challenging for all online detectors.

Unlike point-based anomalies, the vast majority of detectors exhibit a random performance in range-based anomalies. Furthermore, larger intervals of anomalies challenge more all detectors (see Figure~\ref{fig:dist_ex14}). HST/F proved to be the best overall detector in this experiment, exhibiting the highest AUC ROC and MAP scores.

\section{Robustness of Detectors} \label{chapter:robustness}

In this experiment we are assessing the robustness of online \{HST, HSTF, RRCF, MCOD, XSTREAM (X-S), LODA (L-S), {RS-HASH, STARE, CPOD, LEAP}\} and offline \{LOF, KNN$_W$, iForest (IF) , XSTREAM (X-B), LODA (L-B)\} detectors against increasing data and subspace anomaly dimensionality i.e., anomalies hidden in subspaces with more features. To keep constant the anomaly ratio we use twenty synthetic datasets derived from HiCS (see Section \ref{sec:synthetic_datasets}) as follows: for each of the five datasets containing 20 up to 100 features, namely \{HiCS20, HiCS40, HiCS60, HiCS80, HiCS100\}, we generate four variants contaminated with the same number of 2-$d$, 3-$d$, 4-$d$ and 5-$d$ subspace anomalies. To assess the impact of increasing irrelevant features on the detection of subspace anomalies of a given dimensionality, each dataset has four sub-versions, one for 2-$d$, 3-$d$, 4-$d$ and 5-$d$ subspace anomalies. 

\begin{figure*}[!t] 
    \centering
  \subfloat[Online detectors]{
    \includegraphics[width=0.5\linewidth]{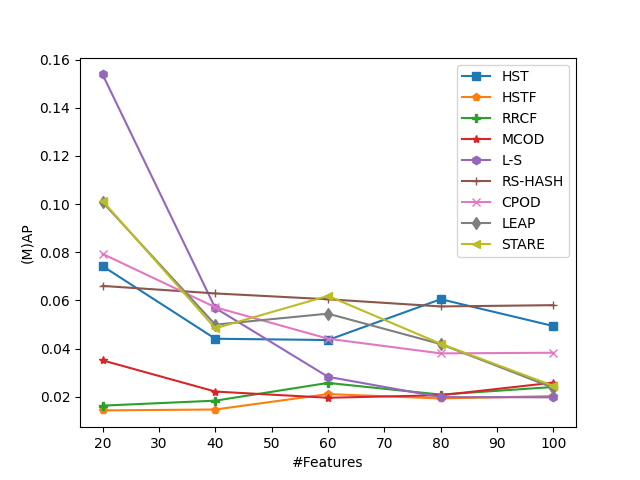}
    \label{fig:online_dim_hics1}
    }
  \subfloat[Offline detectors]{
    \includegraphics[width=0.5\linewidth]{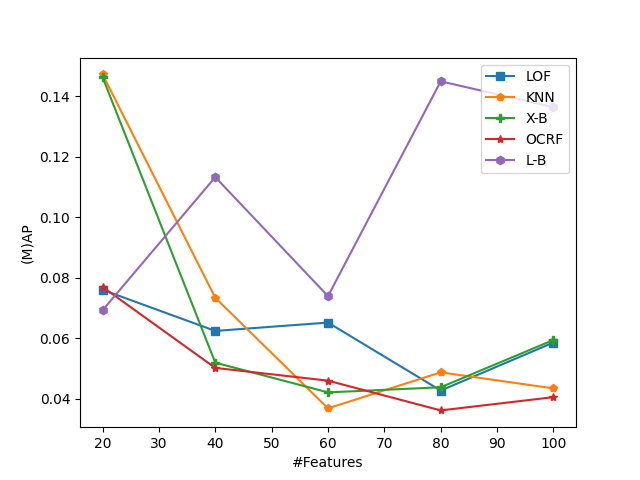}
    \label{fig:offline_dims_hics2}
    }
    \newline
    \subfloat[Online detectors]{
    \includegraphics[width=0.5\linewidth]{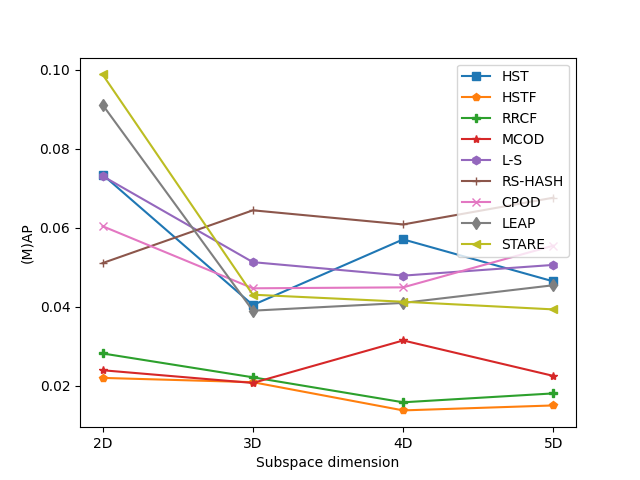}
    \label{fig:online_sub_hics3}
    }
    \subfloat[Offline detectors]{
    \includegraphics[width=0.5\linewidth]{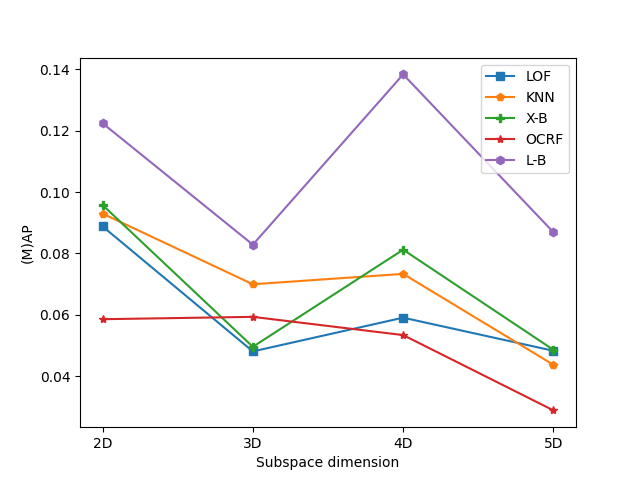}
    \label{fig:offline_sub_hics4}
    }
    \caption{MAP scores over increasing data and subspace dimensionality on synthetic datasets.}
    \label{fig:hics_performances}
\end{figure*}

As we can see in Figure \ref{fig:online_dim_hics1} (a), in synthetic datasets, {projection-based detectors (X-S, L-S) outperform (in terms of MAP) all other online detectors in higher dimensions (80-$d$ and 100-$d$). MCOD is up to one order of magnitude more effective than tree-based detectors (HST/F, RRCF) in 20 and 40 dimensions while in higher dimensions all detectors exhibit a similar effectiveness. CPOD and LEAP seem to better perform than MCOD as data dimensionality increases: in some cases (60-$d$) they even achieve to be two times more effective than MCOD. Moreover, the performance RS-HASH which is a subspace detector, drops as the number of feature increases. HST/F, RRCF and STARE exhibit a similar performance across all dimensions. As expected, distance and density-based detectors like MCOD, CPOD and LEAP underperform as data dimensionality increases.} This is due to the fact that Euclidean distance becomes less effective on higher dimensions \cite{euclidean_effectiveness}. We finally observe that the implemented forgetting mechanism makes HSTF to perform slightly better than HST in detecting subspace anomalies. 

In overall, with the exception of distance-based, online detectors exhibit a \textit{robust} behavior in \textit{increasing} data dimensionality.  This behavior can be attributed to various reasons. First, the random projections enable X-S and L-S to effectively reduce data dimensionality while keeping the distances between samples. This mechanism seems to be less affected by the increasing number of features which are irrelevant to the subspace anomalies of the datasets compared to the random feature splitting employed HST/F and RRCF. {Second, the notion of neighborhood in MCOD, CPOD, and LEAP becomes less meaningful in higher dimensions  \cite{10.5555/645504.656414}.} Third, in our benchmark the successive windows consumed by online detectors are fed  with stratified anomalies over shuffled normal samples (see Section \ref{datasets}). It is known that processing data in sub-samples reduces both \emph{anomaly swamping} and \emph{masking effects} \cite{iforest,sub_sampling}, that may incur when we increase the dataset dimensionality with irrelevant features.

In Figure~\ref{fig:offline_dims_hics2} (b) we can observe that with the exception of OCRF, the effectiveness of offline detectors gets \textit{decreased} while \textit{increasing} data dimensionality. Clearly, in high-dimensional spaces, all pairs of samples become almost equidistant (\emph{distance concentration}), and distance and density-based detectors like KNN$_W$ and LOF struggle to separate separate abnormal from normal samples. In the experiment reported in \cite{high_dim_problem} this effect has been observed at 100-$d$, while in our experiment started earlier at 40-$d$. This is due to the fact that HiCS datasets are contaminated by subspace instead of fullspace anomalies.

IF fails to isolate subspace anomalies, since by increasing the number of features which are irrelevant to the anomalies, the noise in tree structures constructed by uniformly sampling features gets increased. In the experiment reported in \cite{feature_irrelevance_forest} this effect has been observed with the addition of 30 irrelevant features, while in our experiment started earlier even with the addition of 20 irrelevant features due to the contamination of HiCS datasets with subspace anomalies. This is not the case for OCRF, which is the most robust offline detector against increasing data dimensionality. OCRF actually succeeds to obtain better scores on higher dimensions (80-$d$, 100-$d$) compared to lower ones. This is due to the fact, that OCRF relies on an enlighten choice of split features that leads to more consistent splits across increasing dimensions. Also, by increasing the dimensions, the volume of the data goes up, which helps OCRF to perform more accurate splits due to its One Class Gini index splitting criterion.

\begin{table*}[!t]
\centering
\caption{Mean and std. of training and update time (in seconds) of online detectors per window over all datasets: X-S is the fastest detector in terms of update time whereas RRCF is the slowest.}
\label{table:times}
\scalebox{0.7}{
\begin{tabular}{|l|c|c|c|c|c|c|}
\hline
\textbf{Detectors}      & \textbf{X-S}       & \textbf{HST}    & \textbf{HSTF}   & \textbf{RRCF}   & \textbf{MCOD}  \\ \hline
\textbf{Training times} & 32.32 $\pm$ 67.73     & 47.15 $\pm$ 145.88 & 42.9 $\pm$ 146.3   & 59.91 $\pm$ 191.91    & -   \\ \hline
\textbf{Update times}   & 2 $\cdot 10^{-4}$ $\pm$ 25 $\cdot 10^{-5}$ & 1.16 $\pm$ 2.77    & 60.57 $\pm$ 196.31 & 2106.67 $\pm$ 6063.49 & 0.019 $\pm$ 0.013 \\ \hline

  & \textbf{L-S}  & \textbf{RS-HASH} & \textbf{STARE} & \textbf{LEAP} & \textbf{CPOD} \\ \hline
\textbf{Training times} & 1.509 $\pm$ 1.973 & 0.037 $\pm$ 0.003 & {0.023 $\pm$ 0.025} & - & -\\ \hline
\textbf{Update times}  & 0.354 $\pm$ 0.986 & {0.027 $\pm$ 0.057} & {0.021 $\pm$ 0.03} & {0.013 $\pm$ 0.0129} & {0.016 $\pm$ 0.015} \\ \hline
\end{tabular}
}
\end{table*}

We are now turning our attention to the impact of subspace anomaly dimensionality on the effectiveness of detectors. Figure \ref{fig:online_sub_hics3} (c) illustrates the average MAP of online detectors across all HiCS datasets for subspace anomalies of 2-$d$, 3-$d$, 4-$d$ and 5-$d$. {In general, all online detectors exhibit a robust behavior against increasing subspace dimensionality. L-S achieves the higher scores on 3-$d$, 4-$d$ and 5-$d$ subspace anomalies with a performance drop in 2-$d$. LEAP, CPOD and X-S exhibit higher (M)AP scores in 2-$d$ subspace anomalies, while having a relatively stable performance in higher dimensions. Surprisingly enough, the dedicated subspace anomaly detector RS-HASH does not excel in HICS datasets. RS-HASH achieves a better performance in 2-$d$ and 5-$d$ subspace anomalies, while having a lower performance in 3-$d$ and 4-$d$.} MCOD exhibits a better performance than tree-based detectors (HSTF/F, RRCF) due to the fact that it re-computes the actual (L2) distance between samples in sliding windows but also updates both the content and the number of micro-clusters (i.e., deleting old and inserting new). 

Tree-based online detectors exhibit a similar performance across all anomaly subspaces. RRCF is slightly better than HST/F as it updates the tree structure of its model with feature subsets that are more relevant to the subspaces of the anomalies contained in a window instead of only the mass profiles of immutable feature partitions. Clearly, the activation of a forgetting mechanism allows HSTF to better capture 2-$d$ and 3-$d$ subspace anomalies than HST. However, both they face difficulties in finding 4-$d$ and 5-$d$ subspace anomalies. Surprisingly enough, RRCF performance is improved in 4-$d$ subspace anomalies (in HiCS20). Effectiveness of MCOD, HST/F, RRCF and LODA remains robust against increasing subspace dimensionality. As we can see from Figure \ref{fig:offline_sub_hics4} (d), offline detectors exhibit a similar behavior, with ups and downs due to the non deterministic nature of all detectors besides KNN.

\section{Efficiency of Detectors} \label{chapter:efficiency}
In this section we are turning our attention to the efficiency of online detectors. We are focusing on measuring the CPU time of detectors while their memory footprint is left as future work. {Execution times are measured directly in the native language of the respective implementation (C++ for XSTREAM, Matlab for LODA, Python for RS-HASH and JAVA for the rest). Compared to the analytical complexities reported in Table \ref{table:online_detectors_comparison}, the detectors' run-time depends not only on the execution speed of the programming language used, but also on the optimizations implemented at code-level, as well as, on the values of the hyperparameters w.r.t. data distribution in consecutive windows. The actual runtime of all detectors in real datasets lies between the two extremes: X-S as the fastest and RRCF as the slowest detector. For this reason, we rely more on the ranking of detectors' efficiency across all datasets of our testbed rather than their absolute runtimes per dataset. This is in particular useful when studying the trade-off between effectiveness and efficiency of online detectors \{X-S, L-S, HST, HSTF, RRCF, MCOD, RS-HASH, STARE, LEAP, CPOD\} in real datasets (see Section \ref{sec:real_world_datasets}). }

Table~\ref{table:times} depict the mean runtime required for training and updating the model of detectors, respectively across all datasets of our testbed. As training time, we report the total time for model construction while as update time, the average time per window for model update and anomaly detection. To provide a common ground of comparison, the size of windows is the same for all detectors running on the same dataset. {This choice is justified by the results of a preliminary experiment on a subset of the datasets, indicating that a window size of 128 seems to be optimal for most detectors besides HSTF and MCOD (see Figure~\ref{fig:window_sizes} in Appendix).} Only the number of windows vary per dataset. Datasets on the x-axis are ranked according to their dimensionality.

The run-time required to construct the initial model of HST/F is one order of magnitude less than the RRCF mostly in high dimensional datasets. This is due to the fact that HST/F uses a maximum depth (hyper-parameter) to construct small (perfect) binary trees instead of large (full) binary trees, as in the case of RRCF. Similarly, the model updates of HST/F cost one order of magnitude less than RRCF thanks to the constant cost of counter updates in mass profiles, as opposed to tree re-structuring of RRCF. HSTF runtime is slightly greater than HST due to counter decrements for forgetting old samples. 

X-S, exhibits a stable training time, as the dimensionality increases thanks to its sparse random projections, so models are trained on almost a similar number of projections across all datasets. X-S is also the faster detector w.r.t the update time, as its updates have constant complexity (see Table \ref{table:online_detectors_comparison}) {and its implementation benefits from several code-level optimizations. As expected, the two optimized variants of MCOD, LEAP and CPOD, exhibit faster update times (up to $30\%$ faster) compared to MCOD. Despite the fact that LEAP has quadratic time complexity at worst case, it does not update micro-clusters (MCOD) or core points (CPOD) when the window slides; in combination with the minimal probing that is frequently activated, LEAP achieves a faster execution time. RS-HASH and STARE are slightly slower with similar update times. RS-HASH runs on Python which is generally slower compared to JAVA. STARE's update times are highly dependent on the values of its hyperparameters.} Furthermore, we observe that HST/F and RRCF run slower in \{Forestcover, http, InternetAds\} datasets. This is due to the fact that the maximum tree height of their binary trees is proportional to the volume of the dataset ($volume = \#samples * \#features$). In fact, the higher the volume, the higher the trees depth and therefore the more CPU time is needed to construct and update them. In this respect, L-S relies on one of the simple and fast update procedures: it creates k one-dimensional histograms using sparse random projections\footnote{$k$ value is automatically computed during training.} and each histogram is then updated by projecting the training sample onto a vector and then updating the corresponding histogram bin. However, L-S runtime is penalized by the execution speed of the Matlab code.

\begin{figure}[!t] 
    \centering
    \label{sub-fig1:trade_off}
    \includegraphics[width=1.0\linewidth]{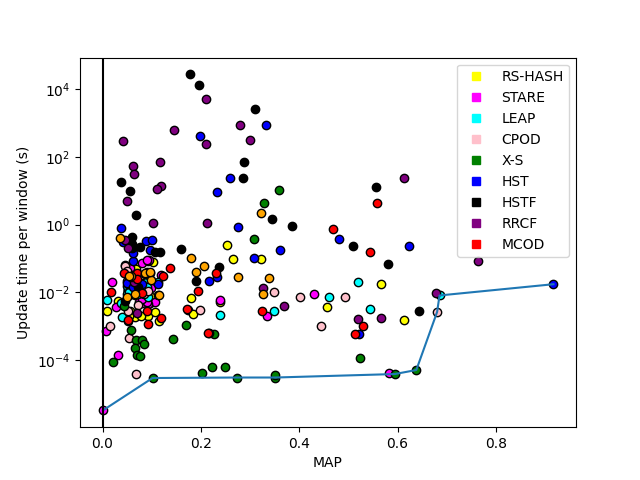}
  \caption{{Pareto frontier of MAP and Update time (per window) for each detector per dataset: X-S dominates in most datasets in terms of both effectiveness and efficiency.}}
  \label{fig:trade_off}
\end{figure}

Figure~\ref{fig:trade_off} illustrates the \emph{Pareto frontier} capturing the trade-off between the update time (per window) and MAP of the ten online detectors \{ X-S, L-S, HST, HSTF, RRCF, RS-HASH, STARE, LEAP, CPOD, MCOD \}. Samples belonging to the Pareto set indicate that the respective detector dominates the remaining ones in the specific dataset that the sample represents. Note that not all datasets necessarily belong to the Pareto set.

We can easily observe that 5 out of 9 of the samples of the Pareto set, belong to X-S. Moreover, 5 more samples related to X-S lie very closely to the Pareto frontier. This fact confirms the leading performance of X-S in terms of both effectiveness and efficiency. The rest 4 samples of the Pareto set belong to STARE, CPOD, LEAP and HST. LEAP and CPOD dominate in the dataset with the highest anomaly ratio (electricity) compared to other detectors with low update times. Note that one of X-S's samples concerns the electricity dataset, as it is the fastest detector with a comparable effectiveness w.r.t. others in this dataset. LEAP dominates in SMTP in terms of efficiency, although its effectiveness is the worst of all. Last but not least, HST dominates in magic-telescope, achieving the highest MAP score w.r.t. detectors with comparable update times.

\section{Meta-learning of XSTREAM's performance} \label{chapter:metalearning}
The experiments in the previous sections showed that X-S is the most effective detector w.r.t. both AUC and MAP metrics while its implementation exhibits the fastest update time. For this reason, we investigate which of the datasets' meta-features discussed in Section \ref{sec:metafeatures} are statistically correlated (using Spearman correlation) with the drop or increase in X-S's efficacy relative to the rest online anomaly detectors. Our meta-learning analysis sheds light on the datasets' characteristics that boost X-S effectiveness or that make detectors to exhibit a similar performance.

\begin{table}[!t]
\caption{{Statistically significant correlations of meta-features with the performance of X-S, 2$^{nd}$ best detector, as well as, their ratio: The scale variance, non-normality of features as well as the existence of subspace anomalies let X-S to outperform the rest detectors.}}
\label{table:correlations}
\centering
\scalebox{0.77}{
\begin{threeparttable}
\begin{tabular}{lccc}
\hline
\multicolumn{1}{|l|}{\textbf{Meta-feature}}                     & \multicolumn{1}{l|}{\textbf{$\rho$ (Ratio)}} & \multicolumn{1}{l|}{\textbf{$\rho$ (X-S)}} & \multicolumn{1}{l|}{\textbf{$\rho$ (2nd Best)}} \\ \hline
\multicolumn{4}{c}{\textit{\textbf{MAP}}}                                                                                                                                                         \\ \hline
\multicolumn{1}{|l|}{\textbf{F4. Mean Kurtosis}}                & \multicolumn{1}{c|}{-0.48*}              & \multicolumn{1}{c|}{-0.32}             & \multicolumn{1}{c|}{0.08}                   \\ \hline
\multicolumn{1}{|l|}{\textbf{F4. SD Kurtosis}}                  & \multicolumn{1}{c|}{-0.46*}              & \multicolumn{1}{c|}{-0.32}             & \multicolumn{1}{c|}{0.04}                   \\ \hline
\multicolumn{1}{|l|}{\textbf{F12. SD Skewness}}                 & \multicolumn{1}{c|}{-0.5*}               & \multicolumn{1}{c|}{-0.38}             & \multicolumn{1}{c|}{0.05}                   \\ \hline
\multicolumn{4}{c}{\textit{\textbf{AUC ROC}}}                                                                                                                                                     \\ \hline
\multicolumn{1}{|l|}{\textbf{G1. \#Samples}}                    & \multicolumn{1}{c|}{0.542**}             & \multicolumn{1}{c|}{0.34}              & \multicolumn{1}{c|}{0.15}                   \\ \hline
\multicolumn{1}{|l|}{\textbf{G2. \#Features}}                   & \multicolumn{1}{c|}{-0.55**}             & \multicolumn{1}{c|}{-0.64***}          & \multicolumn{1}{c|}{-0.31}                  \\ \hline
\multicolumn{1}{|l|}{\textbf{S1. Mean Canonical Corr.}}  & \multicolumn{1}{c|}{-0.488*}             & \multicolumn{1}{c|}{0.21}              & \multicolumn{1}{c|}{0.21}                   \\ \hline
\multicolumn{1}{|l|}{\textbf{S2. Roy’s large root}}             & \multicolumn{1}{c|}{-0.49*}              & \multicolumn{1}{c|}{0.008}             & \multicolumn{1}{c|}{0.33}                   \\ \hline
\multicolumn{1}{|l|}{\textbf{S3. Pillai’s trace}}               & \multicolumn{1}{c|}{-0.49*}              & \multicolumn{1}{c|}{0.008}             & \multicolumn{1}{c|}{0.33}                   \\ \hline
\multicolumn{1}{|l|}{\textbf{S4. Lawly-Hotelling Trace}}        & \multicolumn{1}{c|}{-0.49*}              & \multicolumn{1}{c|}{0.008}             & \multicolumn{1}{c|}{0.33}                   \\ \hline
\multicolumn{1}{|l|}{\textbf{S5. Wilks’ Lambda Value}}          & \multicolumn{1}{c|}{0.49*}               & \multicolumn{1}{c|}{-0.008}            & \multicolumn{1}{c|}{-0.33}                  \\ \hline
\multicolumn{1}{|l|}{\textbf{F3. Mean IQR}}                     & \multicolumn{1}{c|}{0.41*}               & \multicolumn{1}{c|}{0.27}              & \multicolumn{1}{c|}{0.09}                   \\ \hline
\multicolumn{1}{|l|}{\textbf{F3. SD IQR}}                       & \multicolumn{1}{c|}{0.41*}               & \multicolumn{1}{c|}{0.30}              & \multicolumn{1}{c|}{0.10}                   \\ \hline
\multicolumn{1}{|l|}{\textbf{F5. Mean MAD}}                     & \multicolumn{1}{c|}{0.5*}                & \multicolumn{1}{c|}{0.32}              & \multicolumn{1}{c|}{0.09}                   \\ \hline
\multicolumn{1}{|l|}{\textbf{F5. SD MAD}}                       & \multicolumn{1}{c|}{0.48*}               & \multicolumn{1}{c|}{0.33}              & \multicolumn{1}{c|}{0.10}                   \\ \hline
\multicolumn{1}{|l|}{\textbf{F6. Mean MAX}}                     & \multicolumn{1}{c|}{0.49*}               & \multicolumn{1}{c|}{0.31}              & \multicolumn{1}{c|}{0.07}                   \\ \hline
\multicolumn{1}{|l|}{\textbf{F6. SD MAX}}                       & \multicolumn{1}{c|}{0.48*}               & \multicolumn{1}{c|}{0.31}              & \multicolumn{1}{c|}{0.05}                   \\ \hline
\multicolumn{1}{|l|}{\textbf{F7. Mean MEAN}}                    & \multicolumn{1}{c|}{0.57**}              & \multicolumn{1}{c|}{0.39}              & \multicolumn{1}{c|}{0.11}                   \\ \hline
\multicolumn{1}{|l|}{\textbf{F8. Mean Median}}                  & \multicolumn{1}{c|}{0.49*}               & \multicolumn{1}{c|}{0.33}              & \multicolumn{1}{c|}{0.09}                   \\ \hline
\multicolumn{1}{|l|}{\textbf{F8. SD Median}}                    & \multicolumn{1}{c|}{0.43*}               & \multicolumn{1}{c|}{0.35}              & \multicolumn{1}{c|}{0.14}                   \\ \hline
\multicolumn{1}{|l|}{\textbf{F15. \# Statistical Outliers}} & \multicolumn{1}{c|}{-0.45*}              & \multicolumn{1}{c|}{-0.61**}           & \multicolumn{1}{c|}{-0.34}                  \\ \hline
\end{tabular}
       \begin{tablenotes}
            \item[*] For $pvalue \leq 0.05$, **For $pvalue < 0.01$, *** For $pvalue < 0.001$
        \end{tablenotes}
     \end{threeparttable}
}
\end{table}

Table \ref{table:correlations} reports the correlations of every meta-feature found to have statistical significant correlation (i.e., at least $\mathit{pvalue}<0.05$) with the ratio best (i.e., X-S) and second-best online detector (or the best online detector in datasets where X-S loses) in terms of AUC and MAP. The first column of the Table reports the statistical significant correlation values between the meta-features and the ratio between X-S and the second best online detector per dataset. The second and third Table columns report X-S's and second best online detector's raw scores correlations for the meta-features found to have statistical significant correlation with the ratio. 

The number of features ({\tt G2}) is the only general meta-feature with a negative ratio correlation. As data dimensionality increases, the difference between X-S' performance and the second best online detector tend to also decrease. This correlation can be attribute to the fact that in high dimensional datasets, all online detectors score similarly to the Random Classifier. {On the other hand, the performance ratio of X-S with the second best online detector is positively correlated with the number of samples ({\tt G1})}. Meta-features ({\tt F1-F14}) related to the variance of the scale of feature' values are also positively correlated with X-S' performance ratio. As we can see in the second and third columns of Table \ref{table:metafeatures}, these meta-features are more correlated with the raw performance of X-S rather than of the second best detector. Hence, when the ratio increases, this is mainly due to an increase in X-S performance. 

Meta-features such as {\tt F6} (SD Max Value), {\tt F8} (SD Median Value) and {\tt F7} (Mean MEAN Value) are also correlated with the relative  performance of X-S. As discussed in Section~\ref{sec:xstream}, its scoring function aims to assess anomalousness across different features value granularities. As a result, the diversity of the feature' scale boosts X-S's performance compared to the rest of the detectors. In particular, X-S proves to effectively maintain pairwise distances of samples even in datasets whose features exhibit significantly different value scales. 

{
We observe a strong negative correlation of X-S with the Nr. of Statistical Outliers. X-S performance decrease w.r.t. the second best detector as the number of statistical outliers increases. This can be attributed to the fact that the datasets with higher number of statistical outliers are usually of high-dimensionality, so X-S's performance drops as we explained previously.}

{
The ratio between X-S and the second best performing detector is correlated with the meta-features capturing the existence of subspace anomalies ({\tt S1, \tt S2, \tt S3, \tt S4, \tt S5}). There is no significant performance degradation of X-S in subspace anomalies, in construct to the rest of the online detectors, which exhibit a higher correlation. 
This indicates that the density-estimation mechanism of X-S (and L-S) is more robust to subspace anomalies (on real datasets) compared to mechanisms adopted by other detectors.}

Three meta-features are found to be correlated with the relative performance of X-S w.r.t. MAP (Mean/SD Kurtosis, SD Skewness). As the second best online detector is negatively correlated with those meta-features, we observe a negative correlation with the X-S' performance ratio. Higher kurtosis and skewness values reveal denser clusters of samples. To avoid splitting those dense clusters into different bins, X-S relies on a random shifting mechanism, that actually decreases X-S's performance on higher kurtosis and skewness compared to the rest of the detectors. 

{
The pairwise correlations of X-S with online detectors are given in Table~\ref{table:pairwise_correlations_online} of Appendix~\ref{app:meta}. The relative performance of X-S against the two variations of MCOD (CPOD and LEAP) and RS-HASH, is positively correlated with meta-features capturing the variance of features' values (F1-F14). Clearly, features' diversity positively affects X-S's performance. Furthermore, the relative performance of X-S is improved as the number of samples increases compared to RS-HASH and STARE. On the other hand, X-S under-performs on datasets of high dimensionality and anomaly ratio compared to those density-based detectors. Recall that in high-dimensional datasets the majority of detectors exhibit a random behavior. No correlations are observed between X-S and HST, HSTF, as well as, RRCF.}

\section{Lessons Learned} \label{chapter:lessons_learned}
In this section we report the main conclusions drawn from the experiments in our work. 

\emph{The majority of online and offline detectors classify samples in more than half of the datasets with a performance comparable to the Random Classifier.} As expected (see Section~\ref{sec:random}), detectors typically exhibit a good performance when abnormal and normal classes are well separated (high \emph{AND} values) even in datasets of high dimensionality ($>$100). {However, this is not always the case in real datasets contaminated with implanted anomalies. Unlike distance/KNN and density-based detectors calculating samples' distance over all features, the presence of many correlated feature pairs (i.e., multicollinearity) favor stochastic detectors to include less features that are more informative w.r.t. the anomalous class. Unfortunately, random projections of XSTREAM and LODA becomes less effective in most datasets of high dimensionality ($\geq$257, see Table \ref{table:auc_scores}), both detectors exhibit a performance close to the random classifier. }

\emph{Online detectors approximate well and in some cases outperform their offline counterparts.} As shown in Section \ref{sec:onlineVsOffline}, distance-based detectors like MCOD is ranked very close to KNN and LOF, while tree-based detectors like HST approximates well the performance of IF dominating the other member of this algorithmic family, namely, OCRF. Clearly, offline distance/KNN and tree-based detectors are favored by feeding the data into a single large window; KNN and LOF dispose more information regarding the neighborhood of samples, while IF is able to better profile normal samples and thus mitigate the effect of anomaly swamping \cite{iforest}. Finally, the online version of XSTREAM not only outperforms all other online detectors by also is more effective than its offline version.

{
\emph{Tunable and gradual forgetting mechanisms proved to be more effective for range-based anomalies presented on several time-frames.} As shown in Section \ref{sec:exathlon}, only HST/F, RRCF and RS-HASH exhibit a performance better than the random classifier. RRCF and HSTF have a tunable forgetting mechanism and RS-HASH a gradual forgetting mechanism. These properties are essential to detect anomalies especially for longer ranges than global forgetting mechanisms such as XSREAM's, i.e., assigning all bin counts to zero when the window slides. This is due to the fact that the range-based anomalies are clustered; therefore, global forgetting makes the algorithm to rapidly adapt to the new data distribution of anomalies, assigning them scores close to normal samples.}

\emph{Online anomaly detectors exhibit poor performance on the identification of subspace anomalies.} As shown in Section \ref{chapter:robustness}, the majority of online detectors are robust against an increasing subspace or data dimensionality. The best performing strategy was the random projections as performed by L-S. Unlike fullspace anomalies contained in real datasets, anomalies hidden in feature subspaces proved to be a great challenge for all online detectors, as the highest MAP value was 0.16. This observation highlights the need for online detectors that are more effective on identifying subspace anomalies.

\emph{XSTREAM exhibit the best tradeoff between efficiency and effectiveness of continuously updated detection models.} 
As shown in Section \ref{chapter:efficiency}, XSTREAM proves to be both faster w.r.t. updating time and more effective in the majority of the datasets. The different tradeoffs achieved by online detectors can be explained w.r.t. how well their model (bins or histograms in random projections, mass profiles in trees, micro-clusters, core points, etc.) summarizes the data distribution of a stream, as well as, how effective are the incremental updates of models in successive windows. {Finally, we should stress that LODA and RS-HASH share the same updating principles as XSTREAM and the difference to their absolute runtimes is more related to the programming language and the optimizations at code-level. On the other hand, detectors such as LEAP, CPOD, MCOD and STARE seems to run under best cases in the datasets of our testbed. This makes them bypass costly neighbor searches by applying the minimal probing principle (LEAP), explore only cluster centers (MCOD) or cores (CPOD) or skipping stationary regions (STARE).} 

\emph{The non-normality and scale variance of data features favor XSTREAM to outperform the rest detectors on real datasets contaminated with point anomalies.} As shown in Section \ref{chapter:metalearning}, when the scale of feature values exhibit a high dispersion (e.g. high std. of mean values), the difference in AUC performance between XSTREAM and the second best detector is amplified (positive ratio correlation) in favor of XSTREAM (higher raw correlation coefficient than the second best detector). In case of features exhibiting non-normal patterns such as high skewness and kurtosis, XSTREAM's MAP performance decreases (negative raw correlation) but its performance is less affected than the second best detector. Both distributional characteristics imply the existence of dense clusters, concentrated either in the one side of the distribution or around the peak in case of a leptokurtic distribution. It is important that the dense clusters should not be separated by the detectors' partitioning mechanism. XSTREAM handles this issue more effectively than other detectors by employing a shifting mechanism that strives to assign clustered samples on the same bins.

\section{Summary and Future Work} \label{chapter:conclusion_future_work}

In this work, we conducted a large scale experiment to reveal several open questions in the literature. We compared the effectiveness and efficiency of 6 online anomaly detectors along with 5 offline counterparts over 24 real and 1 synthetic dataset with 20 variations. None of the previous empirical studies \cite{offline_benchmark_1, best_offline_2, best_offline_3, offline_benchmark_2, offline_benchmark_3, offline_benchmark_4, offline_benchmark_5} compared the performance of online detectors over a fairly complete collection of datasets used to evaluate the original algorithms. We should note that \cite{best_distance_based} focuses exclusively on  the  efficiency of online proximity-based algorithms without contrasting the effectiveness of algorithms belonging to different families (proximity vs tree vs projection-based). Unlike the aforementioned works, the variations of the synthetic dataset used in our work make it possible to evaluate the behavior of detectors on anomalies visible exclusively on subsets of features.

To fairly compare online and offline algorithms, we relied on both ROC AUC and MAP evaluation metrics, highlighting different aspects of their underlying anomalousness scoring functions. In this respect, we used an evaluation protocol inspired by time-series, namely Forward Chain Cross Validation \cite{rolling_cv}. Rather than using, as in the majority of the empirical studies, the default hyper-parameters provided by the original algorithms we search for optimal hyper-parameter values per dataset. To reveal valuable insights regarding detectors' behavior, we followed a principled methodology to analyse the raw results of our experiments. In particular, we performed a meta-learning of different performance aspects of detectors (e.g., randomness of decisions, trend of outperforming behavior) w.r.t. concrete characteristics of datasets (e.g., ratio of correlated features, feature normality, value range dispersion). 

As future work, we are interested in contrasting the performance of the 'shallow' detection methods studied in this work with deep learning algorithms \cite{Pang2021}, especially for high dimensional datasets that favor a random behavior of detectors. Moreover, we will like to investigate how the behavior of detectors is affected by the anomaly generation method either clustered or scattered across normal samples in datasets.

\section*{Acknowledgements}

The research leading to these results has received funding from the European Research Council under the European Union’s Seventh Framework Programme (FP/2007-2013) /ERC Grant Agreement n. 617393. The research work was also supported by the Hellenic Foundation for Research and Innovation (H.F.R.I.) under the “First Call for H.F.R.I. Research Projects to support Faculty members and Researchers and the procurement of high-cost research equipment grant” (Project Number: 1941).

\bibliographystyle{plain}
\bibliography{bibliography} 

\appendices
\clearpage
\section{Offline Detectors}
\label{app:offline}

Based on the the findings of previous benchmarking efforts \cite{best_offline_1,best_offline_2,best_offline_3} we considered representative offline detectors from three families, namely, distance-based like the weighed KNN$_W$ \cite{knn}, density-based like LOF \cite{lof} and tree-based like IF \cite{iforest} and OCRF \cite{ocrf}. 

\subsection{Weighted K Nearest Neighbor (KNN$_{W}$)}
\label{offline_detectors_knn}
KNN$_{W}$ is a score based variation \cite{weighted_knn} of the distance-based K Nearest Neighbors (KNN) \cite{knn} classifier. An anomaly is a sample that is in substantially higher distance than the rest samples (i.e., they have few neighbors in close distance).

To assign a real-valued score for each sample, it computes a matrix of distances, called \textit{MD}: each row represents a sample and each column represents the distance from another sample. Then, using \textit{MD} it computes the score of a sample \textit{p} as the maximum distance \textit{dist} over its \textit{K} (columns) nearest neighbors as follows: 
\vspace{-0.3cm}
\[
score(p) = max(dist(p, q)) : 1 \leq q \leq K
\]

Higher score indicates more abnormality. The number of nearest neighbors \textit{K} and the distance metric \textit{dist($\cdot$)} are the two hyper-parameters of KNN$_W$. Its effectiveness depends on how carefully \textit{K} is chosen w.r.t. the data characteristics. For a dataset of size $N$, KNN$_{W}$ needs quadratic time $O(N^2)$ to construct \textit{MD} and linear time $O(NK)$ to score each sample.

\subsection{Local Outlier Factor (LOF)} 
\label{offline_detectors_lof}
LOF \cite{lof} is a density-based detector which computes the local density deviation of a sample w.r.t. its neighbors. Essentially, LOF considers an anomaly any sample that lies on a sparse area while its nearest neighbors lie on dense areas.

LOF first computes the reachability distance \textit{rd} between a given sample $p$ and another sample $o$ according to Eq. \ref{eq:rd}, where \textit{dist(p, o)} is the direct distance between the two samples and \textit{k-dist(p,o)} is the distance of $o$ to its k-th nearest neighbor.
\vspace{-0.3cm}
\begin{equation}
    \label{eq:rd}
    \mathit{rd(p, o) = max(dist(p, o), \textnormal{k-}dist(o))}
\end{equation}
\noindent
Then, LOF computes the local reachability density of a sample $p$ as the inverse average reachability distance of $p$ from its k-nearest neighbors \textit{KNN(p)}:
\vspace{-0.3cm}
\begin{equation*}
\mathit{lrd(p) = } \frac{1}{\frac{1}{\vert \mathit{KNN(p)} \vert} \sum_{o \in \mathit{KNN(p)}} \mathit{rd(p, o)}
}.
\end{equation*}
\noindent
The final score of a sample $p$ is computed by comparing to the average local reachability density of its neighbors:
\vspace{-0.3cm}
\begin{equation*}
\label{eq:lof_score}
\mathit{LOF(p)} = \frac{1}{\vert \mathit{KNN(p)} \vert} \sum_{o \in \mathit{KNN(o)}} \frac{\mathit{lrd(o)}}{\mathit{lrd(p)}}
\end{equation*}

The number of nearest neighbors \textit{K} and the distance metric $\mathit{dist(\cdot)}$ are the two hyper-parameters of LOF. The effectiveness of the algorithm strongly depends on the careful selection of \textit{K}. LOF needs quadratic time $O(N^2)$ to compute the distances between each pair of samples and linear time $O(N)$ to score samples, where $N$ is the dataset size.

\subsection{Isolation Forest (IF)} \label{offline_detectors_iforest}
IF is a tree-based ensemble detector \cite{iforest}, which relies on the number of partitions needed to isolate a sample from the rest in a dataset. The less partitions needed to isolate, the more likely a sample is to be an anomaly.  

The algorithm uses a forest of random \textit{Trees} built on samples of the dataset using bootstrapping of size \textit{Max Samples}. For each subsample, it constructs an isolation tree by uniformly selecting \textit{feature}s and their split \textit{value}s as internal nodes. Samples are stored in the leafs, and the \textit{actual height} of the tree could be limited up to a \textit{Max Height}. The length of the path from the tree root to a leaf node, measures how well a sample in that node is isolated from the others: short paths provide evidence of anomalies. The outlyingness score of a sample \textit{p} is then computed by averaging its path length over all Isolation Trees in the forest as follows:
\vspace{-0.3cm}
\begin{equation*}
score(p, n) = 2^{- E(!t(p))/c(n)}
\label{eq:iforest_score}
\end{equation*}
\noindent
where \textit{n} is the \textit{max samples} size, \textit{E(!t(p))} is the average of the actual height \textit{!t(p)} \footnote{When \textit{p} is stored in a leaf of size larger than one, the value of !t(p) is adjusted to c(size).} and \textit{c(n)} is used for normalization: 
\vspace{-0.3cm}
\begin{equation*}
c(n) = 2 (ln(n - 1) + 0.5772156649) - 2 (n - 1) / n
\label{eq:iforest_c}
\end{equation*}
\noindent
The closer to 1 the score is, the more probable the sample to be an anomaly. A score much smaller than 0.5 indicates an inlier. 

The number of \textit{Trees} $t$, the number of samples a tree may contain \textit{Max Samples} $m$, and its \textit{Max Height} $!t$ are the three hyper-parameters of IF. {Max Height} is typically set to $\lceil \mathit{log(n)} \rceil)$ while the variance of scores is usually minimized with ~100 random \textit{Trees}. By avoiding costly distance or density calculation, IF requires linear time $O(tmh)$ to build a forest and linear time $O(tnh)$ to score $n$ samples. 

\subsection{One Class Random Forest (OCRF)} 
\label{offline_detectors_ocrf}
OCRF is a tree-based ensemble detector \cite{ocrf}, extension of the Random Forest. It relies on the adaption of two-class splitting criterion to one-class setting, by assuming an adaptive distribution of anomalies in each node, i.e. assuming the number of anomalies are equal to the number of normal samples in each node. After splitting, one child node captures the maximum amount of samples in a minimal space, while the other the exact opposite. The closer the leaf containing a sample is to the root of the tree, the more likely this sample is to be an anomaly. 

Similarly to IF, it builds a forest of Trees built on samples and features of the dataset using bootstrapping of size Max Samples and Max Features respectively. For each subsample, it constructs a tree by selecting the feature and split value that minimizes the one-class gini improvement proxy. One-class gini improvement proxy is given by the following formula:
\vspace{-0.3cm}
\[I_G^{OC-ad}(t_L, t_R) = \frac{n_{t_L} \gamma n_t \lambda_L}{n_t{_L} + \gamma n_t \lambda_L}  + \frac{n_{t_R} \gamma n_t \lambda_R}{n_t{_R} + \gamma n_t \lambda_R}     \]
\noindent
where, $n_t, n_t{_L}$ and $n_t{_R}$ are the number of samples on the node $t$ and its children nodes ($L,R$) respectively, $\lambda_L = Leb(X_{t_L}/ Leb(X_t))$ and $\lambda_R = Leb(X_{t_R}/ Leb(X_t))$ are the volume proportions of the children nodes and $\gamma$ is a parameter which influences the optimal splits.     

The samples are stored in the leafs, and the \textit{actual height} of the tree could be limited up to a \textit{Max Height}. The length of the path from the tree root to a leaf node, measures how well a sample in that node is isolated from the others: short paths provide evidence of anomalies. The anomalousness score of a sample \textit{p} is then computed by averaging its path length over all trees in the forest as follows:
\vspace{-0.3cm}
\begin{equation*}
\mathit{log_2 score(x)} = - (\sum_{t leaves}{\mathbbm{1}_{\{x\epsilon t\}}d_t + c(n_t)}) / c(n)
\end{equation*}
\noindent
where \textit{$d_t$} is the \textit{depth} of node \textit{t}, and $c(n) = 2H(n-1)-2(n-1)/n$, !t(i) being the harmonic number.

The closer to 1 the score is, the more probable the sample to be an anomaly. A score much smaller than 0.5 indicates an inlier. 

The number of \textit{Trees} $t$, the number of samples a tree may contain \textit{Max Samples} $m$, the number of features for those samples \textit{Max Features} $f$ and its \textit{Max Height} $!t$ are the three hyper-parameters of OCRF. Similarly to IF, \textit{Max Height} is typically set to $\lceil \mathit{log}(n) \rceil$ while the variance scores is usually minimized with ~100 random \textit{Trees}. OCRF requires linear time $O(tmfh)$ to build a forest and linear time $O(tnh)$ to score $n$ samples.

\begingroup
\def\arraystretch{1.5}
\begin{table}[!t]
\centering
\caption{The values of the varying and fixed hyper-parameters}
\label{table:hyperparams_table}
\scalebox{0.55}{
\begin{tabular}{| c | c | c |} 
\hline
\textbf{Hyper-parameter} & \textbf{Value Range} & \textbf{Detectors} \\ [0.5ex] 
\hline
\multicolumn{3}{@{}c}{\textbf{Experimental setting parameters}}   \\
\hline
train size & 50\%  & ALL \\
\hline
max samples & 40\% & ALL \\
\hline
val size & 10\% & ALL \\
\hline
test size & 40\% & ALL \\
\hline
Window size & 128 & ALL \\
\hline
Window slide & 64 & ALL \\
\hline
\multicolumn{3}{@{}c}{\textbf{Common detectors' hyper-parameters}}  \\ \hline
\multirow{2}{*}{Window type} & sliding & RRCF,MCOD \\ & tumbling & X-S,L-S,HST/F  \\
\hline
metric & Euclidean & MCOD, LOF, KNN$_W$  \\
\hline
max depth & $\lceil$(log2(max samples) $\rceil$ & HST, HSTF, RRCF, IF, OCRF  \\
\hline
max leaves & 5000 & HST, HSTF, RRCF, IF, OCRF  \\
\hline
\multicolumn{3}{@{}c}{\textbf{Varying hyper-parameters}}   \\
\hline
trees & 25, 50, 100 & HST, HSTF, RRCF, IF, OCRF \\ 
\hline
forget threshold & 64, 128, 256, 512, max samples & HSTF, RRCF \\
\hline
max distance & \{0.1 up to (95th - 5th percentile)\}$_{x100}$ & MCOD \\
\hline
min neighbors & \{1 up to 64\}$_{x100}$ & MCOD, {STARE, CPOD, LEAP} \\
\hline
neighbors & 5, 10, 15, 20, 30, 50, 100, 150 & LOF, KNN$_W$ \\
\hline
chains & 25, 50, 100 & XSTREAM \\ 
\hline
projection-size (k) & 25, 50, 100 & XSTREAM \\ 
\hline
chain depth & 5, 10, 15 & XSTREAM \\
\hline
{hash tables} & {4, 8, 12} & {RS-HASH } \\
\hline
{iterations} &{25, 50, 100} & {RS-HASH} \\
\hline
\end{tabular}
}
\end{table}
\endgroup

\section{Hyper-parameters}
\label{app:hyperparameters}

We are interested in assessing the effectiveness of a detector under optimal conditions. In this respect, we distinguish between hyper-parameters whose values can be set independently of the datasets and those for which we need to estimate their values per dataset\footnote{In most experimental studies \cite{best_offline_1,best_offline_2,best_offline_3}, detectors were executed using the default values of their  hyper-parameters as \textit{"recommended by their authors"}.}. 

\begin{figure}[!t]
\center\includegraphics[width=0.85\linewidth]{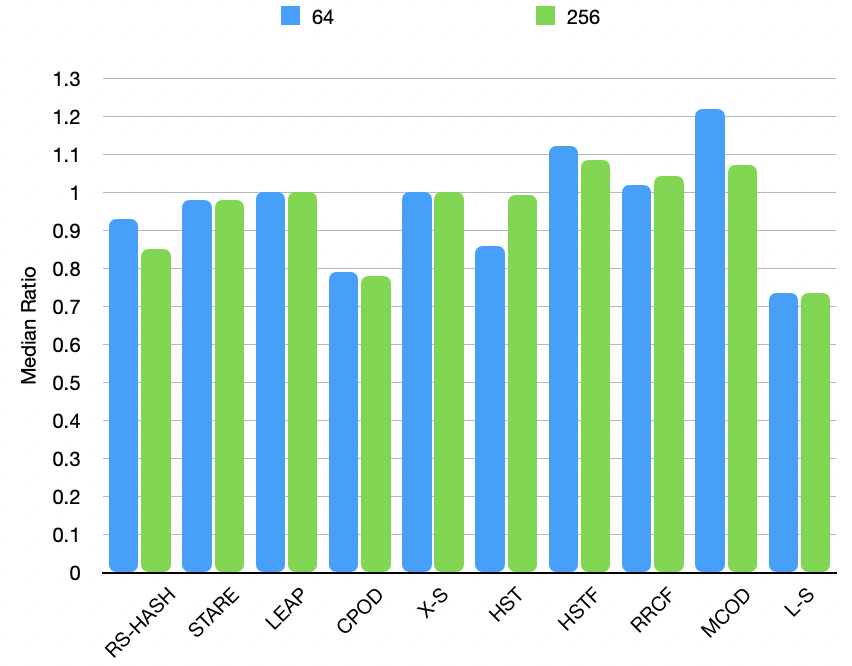}
\caption{Median Ratio of MAP scores on different window sizes (64,256) with 128 as baseline using (Diabetes, Isolet, InternetAds, Wilt). 
}
\label{fig:window_sizes}
\vspace{-0.5cm}
\end{figure}

Table \ref{table:hyperparams_table} presents the fixed values of various hyper-parameters we set in our experiments (e.g., training/testing splitting of datasets, window size and slide) as well as the hyper-parameters of detectors along with the range of values we tested per dataset using random search\footnote{RS does not require to compute the gradient of the problem to be optimized and hence be used on functions that are not continuous or differentiable \cite{rs_optimization_book}.} (RS) \cite{rs_optimization}. The employed ranges empirically ensure the existence of at least an optimal value per detector and dataset. To respect the isolation requirement \cite{cv_importance_1, cv_importance_2}, the optimization phase should take place on an additional validation partition, with samples of the training partition that remain unseen by testing. According to the Tables~\ref{tuned_hyperparameters_online}-\ref{tuned_hyperparameters_offline}, the optimal hyperparameters of each detector per dataset are different in most cases to the default values originally proposed. 

\begin{figure}[!t]
\center\includegraphics[width=0.9\linewidth]{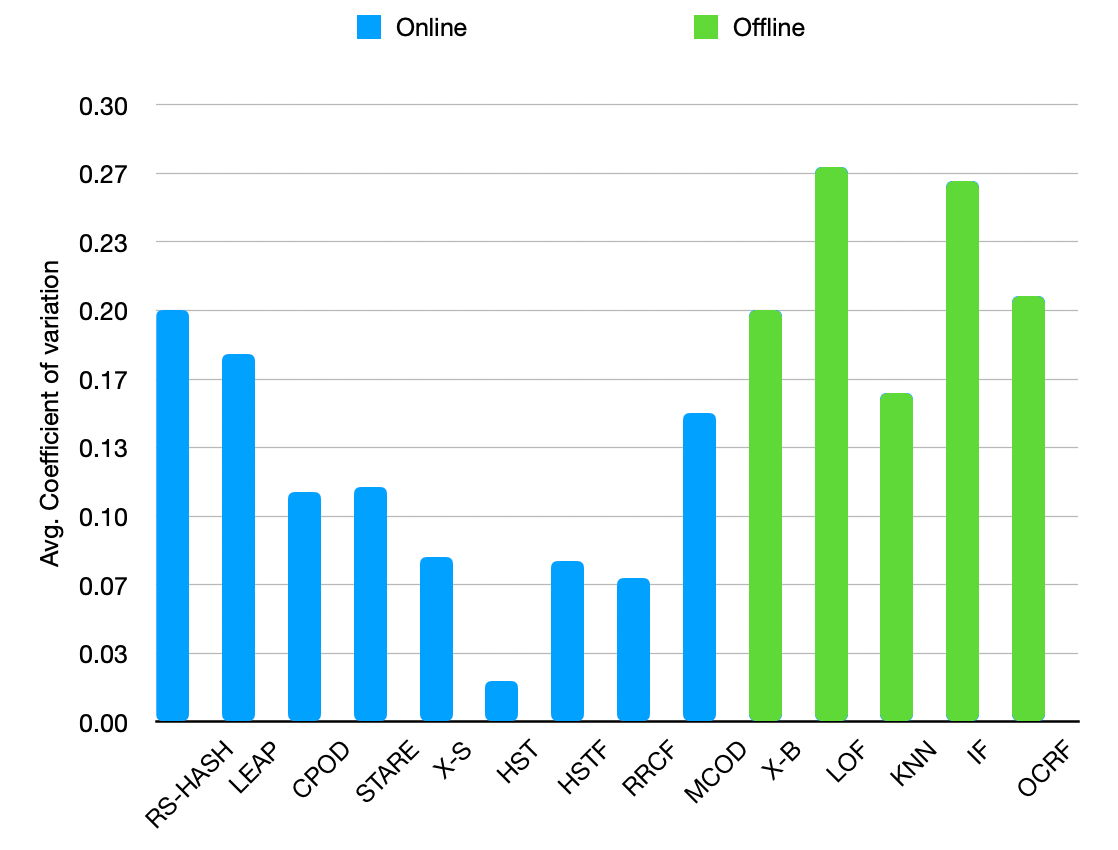}
\caption{Average Coefficient of variation across all datasets of MAP scores of the models built with different hyper-parameters: Online detectors appear to be less sensitive.
}
\label{fig:std_online_hyperparameters}
\vspace{-0.5cm}
\end{figure}

We are additionally interested in assessing \emph{how sensitive detectors are to the tuning of their hyper-parameters}. In this respect, we are computing the average coefficient of variation\footnote{https://en.wikipedia.org/wiki/Coefficient\_of\_variation} of MAP scores of the models tuned with different hyper-parameters across all datasets. {As we can see in Figure \ref{fig:std_online_hyperparameters}, for most online detectors (besides MCOD, LEAP and RS-HASH) the average coefficient is below or close to 0.10. Recall that the tinny performance differences between models are mostly due to the non deterministic nature of the detectors.} On the other hand, offline detectors' average coefficient of variation lies on average around 0.2 and above, revealing that no matter their anomalousness criteria, they are a lot more sensitive to their hyper-parameters tuning. {As we can see from Table~\ref{tuned_hyperparameters_online}, distance-based detectors' hyper-parameters, differ in every dataset, which reveals the need for good tuning according to the characteristics of each dataset (value range, average distance etc.). An interesting observation is that MCOD and CPOD succeeds to have a smaller average coefficient of variation w.r.t. offline detectors (while LEAP is also close). This is attributed to the fact that they exhibit a similar performance to the Random Classifier in most configurations where the two hyper-parameter values are not close to the optimal ones that to a lower coefficient of variation than expected.} On the contrary, tuning does not significantly improve the effectiveness of tree-based and projection-based online detectors.

Regarding hyper-parameters with fixed values across all datasets, we are finally interested in investigating the impact of the window size on detectors' effectiveness. To this end, we tested our detectors on four representative datasets w.r.t. the number of samples/features: Isolet (many samples/features), Wilt (many samples/few features), InternetAds (few samples/many features) and Diabetes (few samples/features). Figure \ref{fig:window_sizes} depicts the median ratio of the MAP scores with \emph{windows size (ws) = 128} as baseline that we use in our experiments. In terms of the window type, we used sliding windows on RRCF, MCOD, LEAP, CPOD, STARE and RS-HASH and tumbling windows on the remaining detectors. According to our experiments, we found that the previously mentioned detectors are performing better on sliding windows compared to tumbling. Note that the window slide indicates the number of past points to forget from the models built. HST exhibits a slightly better performance on \emph{ws} = 128 especially on the datasets with few samples. L-S has a major performance boost using \emph{ws} = 128 on datasets with low samples, while its performance remains stable on the rest. RRCF performs better on low sample datasets with a window size of 64, while HSTF has a performance boost on datasets with many features using \emph{ws} = 256. {RS-HASH is performing better using \emph{ws} = 128 on all cases besides on datasets with many samples and few features. CPOD, LEAP and STARE exhibit better effectiveness using \emph{ws} = 128 in all cases.} X-S is stable throughout all window sizes.

To ensure a fair comparison of detectors in the remaining of our work, we avoid varying the window size per dataset. In this respect, we set the windows size to 128 as it proves to be optimal in most pairs of datasets and detectors. Moreover, we did not choose 256 (or greater) window size, because we need to ensure at least one full window for testing after the training phase; note that the dataset with the lowest number of samples in our benchmark is Ionosphere with 351 samples.

\begin{table*}[!t]
\centering
\caption{Optimal hyper-parameter values of each online detector per dataset after tuning}
\label{tuned_hyperparameters_online}
\scalebox{0.5}{
\begin{tabular}{|l|c|c|c|c|c|c|c|c|c|}
\hline
\textbf{DATASETS}         & \textbf{RS-HASH}     & \textbf{STARE} & \textbf{LEAP}                & \textbf{CPOD}       & \textbf{XSTREAM}   & \textbf{HST}        & \textbf{HSTF}               & \textbf{RRCF}               & \textbf{MCOD}              \\ \hline
\textbf{adult}            & iters=25             & R=29           & maxDistance=3.915   & maxDistance=3.90    & k=100              & trees = 100         & trees= 25                   & trees = 25                  & maxDistance=1.081          \\ \hline
\textbf{}                 & h\_tables=4          & k=57           & minNeighborCount=24          & minNeighborCount=59 & chains=25 &                     & forgetThreshold=64          & forget= max samples         & minNeighbor=50             \\ \hline
\textbf{ALOI}             & iters=100            & R=11           & maxDistance=0.112            & maxDistance=0.127   & k=100              & trees=25            & trees= 25                   & trees= 25                   & maxDistance=1.427          \\ \hline
\textbf{}                 & h\_tables=4          & k=31           & minNeighborCount=28          & minNeighborCount=55 & chains=100         &                     & forgetThreshold=256         & forgetThreshold=256         & minNeighbor=30             \\ \hline
\textbf{Annthyroid}       & iters=25             & R=27           & maxDistance=2.483            & maxDistance=2.363   & k=25               & trees = 50          & trees = 25                  & trees = 100                 & maxDistance=0.694          \\ \hline
\textbf{}                 & h\_tables=8          & k=8            & minNeighborCount=48 & minNeighborCount=56 & chains=100         &                     & forget= max samples         & forget= max samples         & minNeighbor=59             \\ \hline
\textbf{Arrhythmia}       & iters=50             & R=8            & maxDistance=0.163            & maxDistance=1.969   & k=50               & trees = 25          & trees = 25                  & trees = 50                  & maxDistance=3.427 \\ \hline
\textbf{}                 & h\_tables=8          & k=56           & minNeighborCount=47          & minNeighborCount=5  & chains=100         &                     & forget= 128                 & forget= 128                 & minNeighbor=49             \\ \hline
\textbf{breast-w}         & iters=25             & R=4            & maxDistance=0.989            & maxDistance=3.807   & k=100              & \textbf{trees = 50} & trees = 100                 & trees = 25                  & maxDistance=2.495          \\ \hline
\textbf{}                 & h\_tables=4          & k=7            & minNeighborCount=34          & minNeighborCount=36 & chains=25          &                     & forget= 256                 & forget= max samples         & minNeighbor=40             \\ \hline
\textbf{cardiotocography} & iters=100            & R=15           & maxDistance=1.668            & maxDistance=0.347   & k=50               & trees = 100         & trees = 25                  & trees = 100                 & maxDistance=1.081          \\ \hline
\textbf{}                 & h\_tables=8          & k=63           & minNeighborCount=2           & minNeighborCount=32 & chains=50          &                     & forget= 512                 & forget= max samples         & minNeighbor=50             \\ \hline
\textbf{diabetes}         & iters=50    & R=22           & maxDistance=3.094            & maxDistance=2.130   & k=100              & trees = 100         & trees = 25                  & trees = 25                  & maxDistance=3.427          \\ \hline
\textbf{}                 & h\_tables=4          & k=42           & minNeighborCount=18          & minNeighborCount=6  & chains=50          &                     & forget = 64                 & forget = 512                & minNeighbor=49             \\ \hline
\textbf{Electricity}      & iters=25             & R=3            & maxDistance=0.777            & maxDistance=0.178   & k=50               & trees=25            & trees= 25                   & trees= 25                   & maxDistance=2.113          \\ \hline
\textbf{}                 & h\_tables=4          & k=16           & minNeighborCount=54          & minNeighborCount=40 & chains=25          &                     & forgetThreshold=64          & forgetThreshold=10000       & minNeighbor=35             \\ \hline
\textbf{forestcover}      & iters=25             & R=6            & maxDistance=3.398            & maxDistance=1.523   & k=25               & trees=25            & trees= 100                  & trees= 100                  & maxDistance=1.391          \\ \hline
\textbf{}                 & h\_tables=4          & k=24           & minNeighborCount=12          & minNeighbor=42      & chains=50          &                     & forgetThreshold=max samples & forgetThreshold=max samples & minNeighbor=34             \\ \hline
\textbf{http}             & iters=25             & R=1            & maxDistance=2.429            & maxDistance=3.396   & k=50               & trees=50            & trees= 100        & trees= 100                  & maxDistance=3.360          \\ \hline
\textbf{}                 & h\_tables=4 & k=33           & minNeighborCount=56          & minNeighbor=3       & chains=100         &                     & forgetThreshold=max samples & forgetThreshold=max samples & minNeighbor=4              \\ \hline
\textbf{Hypothyroid}      & iters=50             & R=5            & maxDistance=0.989            & maxDistance=0.347   & k=100              & trees=100           & trees= 50                   & trees= 50                   & maxDistance=0.741          \\ \hline
\textbf{}                 & h\_tables=12         & k=59           & minNeighborCount=34          & minNeighborCount=32 & chains=50          &                     & forgetThreshold=max samples & forgetThreshold=max samples & minNeighbor=22             \\ \hline
\textbf{InternetAds}      & iters=100            & R=22           & maxDistance=3.112            & maxDistance=2.595   & k=25               & trees=50            & trees= 100                  & trees= 50                   & maxDistance=0.146          \\ \hline
\textbf{}                 & h\_tables=8          & k=27           & minNeighborCount=60          & minNeighborCount=14 & chains=25          &                     & forgetThreshold=256         & forgetThreshold=512         & minNeighbor=18             \\ \hline
\textbf{Ionosphere}       & iters=100            & R=13           & maxDistance=2.985            & maxDistance=3.010   & k=100              & trees=100           & trees= 50                   & trees= 100                  & maxDistance=0.375          \\ \hline
\textbf{}                 & h\_tables=4          & k=21           & minNeighborCount=18          & minNeighborCount=7  & chains=25          &                     & forgetThreshold=64          & forgetThreshold=64          & minNeighbor=13             \\ \hline
\textbf{isolet}           & iters=50             & R=12           & maxDistance=0.163            & maxDistance=2.130   & k=25               & trees = 100         & trees= 25                   & trees= 50                   & maxDistance=1.281          \\ \hline
\textbf{}                 & h\_tables=4          & k=1            & minNeighborCount=47          & minNeighborCount=6  & chains=25          &                     & forgetThreshold=64          & forgetThreshold=256         & minNeighbor=50             \\ \hline
\textbf{letter}           & iters=50             & R=15           & maxDistance=3.398            & maxDistance=1.495   & k=25               & trees = 100         & trees= 25                   & trees= 50                   & maxDistance=2.373          \\ \hline
\textbf{}                 & h\_tables=4          & k=3            & minNeighborCount=12          & minNeighborCount=16 & chains=25          &                     & forgetThreshold=max samples & forgetThreshold=max samples & minNeighbor=53             \\ \hline
\textbf{madelon}          & iters=100            & R=16           & maxDistance=3.094            & maxDistance=2.146   & k=25               & trees = 25          & trees= 100                  & trees= 25                   & maxDistance=3.427          \\ \hline
\textbf{}                 & h\_tables=12         & k=7            & minNeighborCount=18          & minNeighborCount=38 & chains=25          &                     & forgetThreshold=64          & forgetThreshold=256         & minNeighbor=49             \\ \hline
\textbf{magic}            & iters=50             & R=1            & maxDistance=2.183            & maxDistance=3.289   & k=25               & trees = 50          & trees= 25                   & trees= 100                  & maxDistance=3.427          \\ \hline
\textbf{}                 & h\_tables=4          & k=33           & minNeighborCount=35          & minNeighborCount=44 & chains=50          &                     & forgetThreshold=max samples & forgetThreshold=max samples & minNeighbor=49             \\ \hline
\textbf{MNIST}            & iters=100            & R=5            & maxDistance=0.163            & maxDistance=1.969   & k=50               & trees=50            & trees= 50                   & trees= 100                  & maxDistance=1.559          \\ \hline
\textbf{}                 & h\_tables=12         & k=59           & minNeighborCount=47          & minNeighborCount=5  & chains=100         &                     & forgetThreshold=128         & forgetThreshold=max samples & minNeighbor=46             \\ \hline
\textbf{PageBlocks}       & iters=50             & R=3            & maxDistance=0.777            & maxDistance=1.992   & k=100              & trees=25            & trees= 100                  & trees= 100                  & maxDistance=3.427          \\ \hline
\textbf{}                 & h\_tables=8          & k=12           & minNeighborCount=54          & minNeighborCount=60 & chains=100         &                     & forgetThreshold=64          & forgetThreshold=max samples & minNeighbor=49             \\ \hline
\textbf{pendigits}        & iters=50             & R=29           & maxDistance=3.968            & maxDistance=3.010   & k=100              & trees = 50          & trees = 50                  & trees = 50                  & maxDistance=1.081          \\ \hline
\textbf{}                 & h\_tables=12         & k=32           & minNeighborCount=63          & minNeighborCount=7  & chains=100         &                     & forget= max samples         & forget= max samples         & minNeighbor=50             \\ \hline
\textbf{Pima}             & iters=50             & R=3            & maxDistance=0.163            & maxDistance=2.130   & k=50               & trees=25            & trees= 50                   & trees= 50                   & maxDistance=1.081          \\ \hline
\textbf{}                 & h\_tables=4          & k=16           & minNeighborCount=47          & minNeighborCount=6  & chains=25          &                     & forgetThreshold=64          & forgetThreshold=max samples & minNeighbor=50             \\ \hline
\textbf{Smtp}             & iters=25             & R=1            & maxDistance=2.375            & maxDistance=3.285   & k=50               & trees=50            & trees= 25                   & trees= 100                  & maxDistance=3.427          \\ \hline
\textbf{}                 & h\_tables=4          & k=2            & minNeighborCount=28          & minNeighborCount=37 & chains=100         &                     & forgetThreshold=128         & forgetThreshold=64          & minNeighbor=49             \\ \hline
\textbf{WDBC}             & iters=50             & R=10           & maxDistance=0.777            & maxDistance=1.862   & k=100              & trees=25            & trees= 50                   & trees= 25                   & maxDistance=0.152          \\ \hline
\textbf{}                 & h\_tables=4          & k=3            & minNeighborCount=54          & minNeighborCount=17 & chains=50          &                     & forgetThreshold=512         & forgetThreshold=256         & minNeighbor=41             \\ \hline
\textbf{wilt}             & iters=25             & R=22           & maxDistance=2.944            & maxDistance=3.807   & k=50               & trees = 100         & trees = 50                  & trees = 50                  & maxDistance=1.281          \\ \hline
                          & h\_tables=4          & k=42           & minNeighborCount=29          & minNeighborCount=36 & chains=25          &                     & forget= 512                 & forget= 512                 & minNeighbor=50             \\ \hline
\end{tabular}
}
\end{table*}

\begin{table*}[!t]
\centering
\caption{Optimal hyper-parameter values of online detectors HST/F, RRCF and RS-Hash per Exathlon dataset after tuning (only if the AUC ROC score of the detectors is greater than 0.5)}
\label{table:tuned_hyperparameters_exathlon}
\scalebox{0.7}{
\begin{tabular}{|l|c|c|c|c|}
\hline
\textbf{DATASETS}          & \textbf{HST}   & \textbf{HSTF}       & \textbf{RRCF}       & \textbf{RSHASH} \\ \hline
\textbf{1\_4\_1000000\_80} & trees=50       & trees=100           & trees=100           & -               \\ \hline
                           & windowSize=256 & windowSize=256      & windowSize=128      & -               \\ \hline
                           &                & forgetThreshold=256 & forgetThreshold=256 & -               \\ \hline
\textbf{1\_2\_100000\_68}  & trees=50       & trees=50            & trees=25            & iters=100       \\ \hline
                           & windowSize=128 & windowSize=128      & windowSize=64       & windowSize=128  \\ \hline
                           &                & forgetThreshold=128 & forgetThreshold=128 & h\_tables=2     \\ \hline
\textbf{1\_5\_1000000\_86} & trees=100      & trees=50            & trees=100           & iters=50        \\ \hline
                           & windowSize=64  & windowSize=64       & windowSize=128      & windowSize=256  \\ \hline
                           &                & forgetThreshold=64  & forgetThreshold=256 & h\_tables=4     \\ \hline
\textbf{6\_1\_500000\_65}  & trees=100      & trees=100           & trees=100           & -               \\ \hline
                           & windowSize=128 & windowSize=128      & windowSize=256      & -               \\ \hline
                           &                & forgetThreshold=256 & forgetThreshold=256 & -               \\ \hline
\textbf{6\_3\_200000\_76}  & trees=25       & trees=100           & -                   & -               \\ \hline
                           & windowSize=256 & windowSize=128      & -                   & -               \\ \hline
                           &                & forgetThreshold=512 & -                   & -               \\ \hline
\end{tabular}
}
\end{table*}

\begin{figure}[!t] 
    \centering
  \subfloat[AP]{
    \includegraphics[width=1\linewidth]{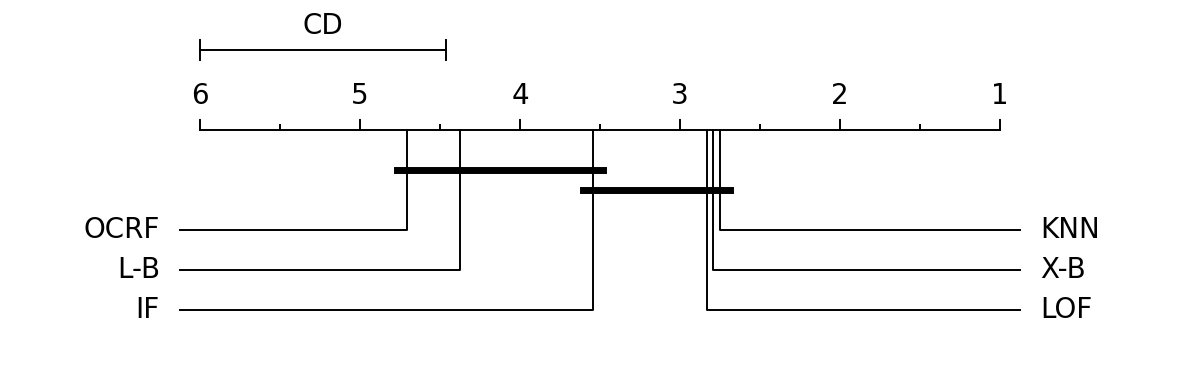}
    \label{fig:ap_offline_rank}
    }
    \hfill
  \subfloat[AUC ROC]{
    \includegraphics[width=1\linewidth]{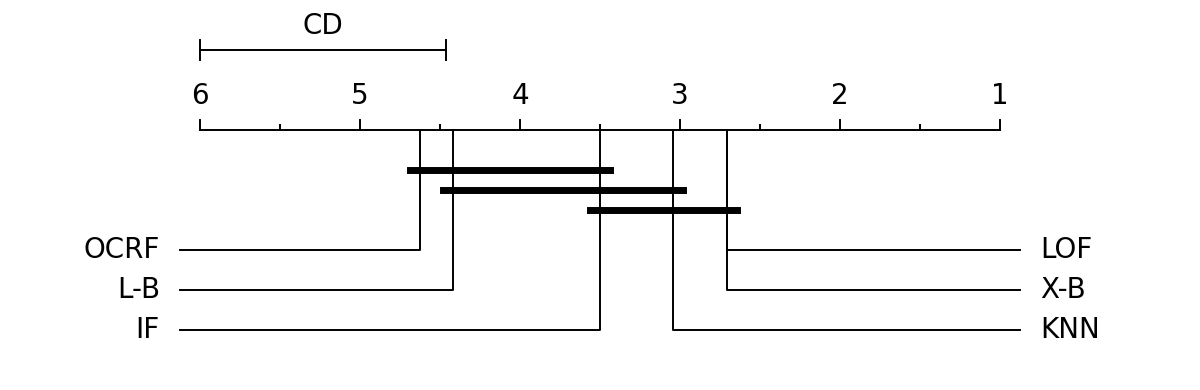}
    \label{fig:auc_offline_rank}
    }
    \caption{Offline detectors' ranking}
\end{figure}

\begin{table*}[!t]
\caption{Optimal hyper-parameter values of each offline detector per dataset after tuning}
\label{tuned_hyperparameters_offline}
\centering
\scalebox{0.7}{
\begin{tabular}{|l|c|c|c|c|c|}
\hline
                          & \multicolumn{1}{l|}{\textbf{IF}} & \multicolumn{1}{l|}{\textbf{KNN}} & \multicolumn{1}{l|}{\textbf{LOF}} & \multicolumn{1}{l|}{\textbf{X-B}} & \multicolumn{1}{l|}{\textbf{OCRF}} \\ \hline
\textbf{adult}            & trees=50                              & neighbors=15                      & neighbors=150                     & chains=25, depth=10                               & trees=25                           \\ \hline
\textbf{}                 &                                       &                                   &                                   & k=100                                   &                                    \\ \hline
\textbf{ALOI}             & trees=50                              & neighbors=5                       & neighbors=5                       & chains=50, depth=5                               & trees=50                           \\ \hline
\textbf{}                 &                                       &                                   &                                   & k=100                                   &                                    \\ \hline
\textbf{Annthyroid}       & trees=100                             & neighbors=5                       & neighbors=5                       & chains=25, depth=15                               & trees=50                           \\ \hline
\textbf{}                 &                                       &                                   &                                   & k=50                                    &                                    \\ \hline
\textbf{Arrhythmia}       & trees=100                             & neighbors=30                      & neighbors=10                      & chains=50, depth=10                               & trees=25                           \\ \hline
\textbf{}                 &                                       &                                   &                                   & k=50                                    &                                    \\ \hline
\textbf{breast-w}         & trees=50                              & neighbors=20                      & neighbors=150                     & chains=25, depth=10                               & trees=50                           \\ \hline
\textbf{}                 &                                       &                                   &                                   & k=25                                    &                                    \\ \hline
\textbf{cardiotocography} & trees=100                             & neighbors=100                     & neighbors=5                       & chains=100, depth=5                              & trees=50                           \\ \hline
\textbf{}                 &                                       &                                   &                                   & k=50                                    &                                    \\ \hline
\textbf{diabetes}         & trees=100                             & neighbors=5                       & neighbors=5                       & chains=100, depth=10                              & trees=100                          \\ \hline
\textbf{}                 &                                       &                                   &                                   & k=100                                   &                                    \\ \hline
\textbf{Electricity}      & trees=25                              & neighbors=100                     & neighbors=150                     & chains=25, depth=10                               & trees=25                           \\ \hline
\textbf{}                 &                                       &                                   &                                   & k=25                                    &                                    \\ \hline
\textbf{forestcover}      & trees=25                              & neighbors=30                      & neighbors=150                     & chains=50, depth=10                               & trees=100                          \\ \hline
\textbf{}                 &                                       &                                   &                                   & k=25                                    &                                    \\ \hline
\textbf{http}             & trees=100                             & neighbors=5                       & neighbors=50                      & chains=100, depth=10                              & trees=50                           \\ \hline
\textbf{}                 &                                       &                                   &                                   & k=50                                    &                                    \\ \hline
\textbf{Hypothyroid}      & trees=50                              & neighbors=5                       & neighbors=10                      & chains=25, depth=10                               & trees=50                           \\ \hline
\textbf{}                 &                                       &                                   &                                   & k=50                                    &                                    \\ \hline
\textbf{InternetAds}      & trees=100                             & neighbors=5                       & neighbors=15                      & chains=100, depth=15                              & trees=100                          \\ \hline
\textbf{}                 &                                       &                                   &                                   & k=100                                   &                                    \\ \hline
\textbf{Ionosphere}       & trees=100                             & neighbors=100                     & neighbors=100                     & chains=50, depth=10                               & trees=100                          \\ \hline
\textbf{}                 &                                       &                                   &                                   & k=25                                    &                                    \\ \hline
\textbf{isolet}           & trees=100                             & neighbors=10                      & neighbors=30                      & chains=50, depth=10                               & trees=25                           \\ \hline
\textbf{}                 &                                       &                                   &                                   & k=50                                    &                                    \\ \hline
\textbf{letter-recognition}           & trees=25                              & neighbors=10                      & neighbors=30                      & chains=25, depth=15                               & trees=50                           \\ \hline
\textbf{}                 &                                       &                                   &                                   & k=50                                    &                                    \\ \hline
\textbf{madelon}          & trees=100                             & neighbors=5                       & neighbors=10                      & chains=50, depth=10                               & trees=25                           \\ \hline
\textbf{}                 &                                       &                                   &                                   & k=25                                    &                                    \\ \hline
\textbf{magic-telescope}            & trees=50                              & neighbors=5                       & neighbors=50                      & chains=100, depth=10                              & trees=25                           \\ \hline
\textbf{}                 &                                       &                                   &                                   & k=100                                   &                                    \\ \hline
\textbf{MNIST}            & trees=25                              & neighbors=30                      & neighbors=150                     & chains=50, depth=10                               & trees=25                           \\ \hline
\textbf{}                 &                                       &                                   &                                   & k=50                                    &                                    \\ \hline
\textbf{PageBlocks}       & trees=25                              & neighbors=5                       & neighbors=30                      & chains=50, depth=15                               & trees=100                          \\ \hline
\textbf{}                 &                                       &                                   &                                   & k=25                                    &                                    \\ \hline
\textbf{pendigits}        & trees=50                              & neighbors=5                       & neighbors=5                       & chains=100, depth=10                              & trees=25                           \\ \hline
\textbf{}                 &                                       &                                   &                                   & k=100                                   &                                    \\ \hline
\textbf{Pima}             & trees=50                              & neighbors=5                       & neighbors=5                       & chains=25, depth=10                               & trees=50                           \\ \hline
\textbf{}                 &                                       &                                   &                                   & k=500                                   &                                    \\ \hline
\textbf{Smtp}             & trees=25                              & neighbors=50                      & neighbors=150                     & chains=100, depth=10                              & trees=50                           \\ \hline
\textbf{}                 &                                       &                                   &                                   & k=25                                    &                                    \\ \hline
\textbf{WDBC}             & trees=50                              & neighbors=100                     & neighbors=10                      & chains=50, depth=5                               & trees=25                           \\ \hline
\textbf{}                 &                                       &                                   &                                   & k=100                                   &                                    \\ \hline
\textbf{wilt}             & trees=50                              & neighbors=5                       & neighbors=10                      & chains=50, depth=15                               & trees=50                           \\ \hline
                          &                                       &                                   &                                   & k=25                                    &                                    \\ \hline
\end{tabular}
}
\end{table*}

\section{AUC/MAP of Detectors}
\label{sec:aucmapscores}

For the shake of completeness, we are also \emph{investigating the performance ranking of online detectors}. Table~\ref{table:offline_ap} depicts the total number of wins of offline detectors using MAP. X-B and LOF outperform the remaining detectors in 7 datasets respectively. This places LOF the best performing proximity-based detector and X-B the best ensemble-based detector. KNN is best performing in 3 datasets along with OCRF and IF. The worst performing detector is L-B, managing to lead only in 1 dataset. It is worth mentioning that KNN has the lowest average difference from the leader along with LOF at 9.8\%. There are no major changes in Table~\ref{table:offline_auc}, which illustrates the number of wins using AUC ROC. LOF achieves to get more wins in AUC ROC, being the best detector, leading in 9 out of 24 datasets compared to X-B getting 2 less wins than before. OCRF dropped to just one win, as well as having the highest average difference from leader. It is worth mentioning that OCRF does not manage to make any split in some datasets, due to the high range of features, which leads to an infinite volume ($> 10^{308}$)

As we did on the previous section (Section~\ref{sec:online_ranking}) we use the non parametric Friedman test \cite{friedman}, in order determine if there are any significant difference between the average ranks of the detectors. With a $pvalue < 0.000$ we reject the null hypothesis with a confidence level of 5\% that all detectors' performances are the same. Subsequently, we use the post-hoc Nemenyi test, in order to compare the detectors in pairs. There is a significant difference when the difference between the average ranks of two detectors is higher than a critical distance (CD) of 1.539 (for 6 detectors on 25 datasets at a significance level a = 0.05) Figure \ref{fig:ap_offline_rank} illustrates the ranking of detectors w.r.t. MAP scores. KNN and X-B are both ranked first (rank tie) followed closely by LOF. KNN is ranked higher than LOF, while having almost half of the wins, due to its low average difference from leader. We observe that there is a statistically significant difference between the first three detectors (LOF,KNN,X-B) and both OCRF and L-B which come last. As we can Figure \ref{fig:auc_offline_rank}, there are no drastic changes on ranking when using AUC ROC. LOF is first as expected from the previous results (9/24 wins) followed by KNN and X-B. There is a statistically significant difference between the two proximity-based detectors (LOF, KNN) and both L-B and OCRF, while IF and X-B have a statistical significant difference only with OCRF.

In overall, we observe that the two proximity-based detectors (LOF, KNN) ,as well as X-B, outperform all other detectors, with L-B and OCRF exhibiting the worst performance. LOF is performing better when using AUC ROC, while the rest of the detectors remain stable w.r.t both metrics.

\begin{table*}[!t]
\caption{(M)AP Scores of online and offline detectors per dataset}
\label{table:map_scores}

\scalebox{0.55}{
\begin{tabular}{|l|c|c|c|c|c|c|c|c|c|c|c|c|c|c|c|c|}
\hline
\textbf{DATASETS}           & \textbf{RS-HASH} & \textbf{STARE} & \textbf{LEAP}   & \textbf{CPOD} & \textbf{X-S} & \textbf{HST}   & \textbf{HSTF}  & \textbf{RRCF} & \textbf{MCOD}  & \textbf{L-S}   & \textbf{IF}     & \textbf{LOF}    & \textbf{KNN}    & \textbf{X-B}    & \textbf{OCRF}  & \textbf{L-B} \\ \hline
\textbf{http}               & 0.179            & 0.0068         & \textbf{0.5445} & 0.195         & 0.17         & 0.309          & 0.345          & 0.28          & 0.122          & 0.18           & 0.1674          & 0.0422          & 0.013           & 0.203           & 0.0058         & 0.0072       \\ \hline
\textbf{smtp}               & 0.183            & 0.0007         & 0.4599          & 0.4449        & 0.3505       & 0.089          & 0.077          & 0.11          & 0.119          & 0.086          & \textbf{0.7924} & 0.0825          & 0.4767          & 0.3887          & 0.4208         & 0.2267       \\ \hline
\textbf{wilt}               & 0.0539           & 0.08           & 0.054           & 0.054         & 0.0755       & 0.275          & 0.16           & 0.05          & 0.214          & \textbf{0.339} & 0.0478          & 0.0994          & 0.0792          & 0.0524          & 0.0374         & 0.0409       \\ \hline
\textbf{adult}              & 0.2393           & 0.3351         & 0.2394          & 0.2394        & 0.4015       & 0.359          & 0.333          & 0.197         & \textbf{0.52}  & 0.469 & 0.3424          & 0.3592          & 0.4106          & 0.35            & 0.421          & 0.3036       \\ \hline
\textbf{Diabetes}           & 0.0653           & 0.093          & 0.0508          & 0.0508        & 0.0582       & 0.623          & 0.51           & 0.613         & \textbf{0.678} & 0.544          & 0.0743          & 0.077           & 0.0668          & 0.066           & 0.0542         & 0.0659       \\ \hline
\textbf{electricity}        & 0.5667           & 0.5818         & \textbf{0.6853} & 0.6804        & 0.6381       & 0.216          & 0.19           & 0.212         & 0.216          & 0.195          & 0.5238          & 0.4964          & 0.5053          & 0.525           & 0.494          & 0.6309       \\ \hline
\textbf{pima}               & 0.4571           & 0.4303         & 0.349           & 0.349         & 0.5242       & 0.096          & 0.106          & 0.102         & 0.091          & 0.099          & \textbf{0.5353} & 0.4884          & 0.4861          & 0.5226          & 0.5108         & 0.4174       \\ \hline
\textbf{breast-w}           & 0.0928           & 0.107          & 0.1133          & 0.1126        & 0.2489       & 0.081          & 0.112          & 0.06          & 0.064          & 0.071          & 0.2576          & 0.2095          & 0.2983          & \textbf{0.37}   & 0.1568         & 0.1472       \\ \hline
\textbf{forestcover}        & 0.0096           & 0.0264         & 0.0096          & 0.015         & 0.143        & \textbf{0.48}  & 0.285          & 0.145         & 0.018          & 0.191          & 0.0858          & 0.0238          & 0.0677          & 0.168           & 0.0138         & 0.0917       \\ \hline
\textbf{magic-telescope}    & 0.2652           & 0.0797         & 0.0716          & 0.0716        & 0.3514       & \textbf{0.917} & 0.643          & 0.764         & 0.53           & 0.327          & 0.3616          & 0.342           & 0.4135          & 0.3587          & 0.2551         & 0.303        \\ \hline
\textbf{PageBlocks}         & 0.1031           & 0.0901         & 0.0501          & 0.0501        & 0.2273       & 0.361          & 0.385          & 0.21          & 0.092          & 0.207          & 0.4432          & \textbf{0.4728} & 0.1496          & 0.2518          & 0.2896         & 0.1592       \\ \hline
\textbf{pendigits}          & 0.055            & 0.0736         & 0.05            & 0.05          & 0.2025       & 0.061          & 0.072          & 0.046         & 0.052          & 0.05           & 0.1265          & \textbf{0.6784} & 0.5642          & 0.1553          & 0.0652         & 0.0573       \\ \hline
\textbf{Cardiotocography}   & 0.2533           & 0.0802         & 0.0511          & 0.0511        & 0.2217       & 0.522          & \textbf{0.581} & 0.557         & 0.566          & 0.513          & 0.2993          & 0.1728          & 0.1593          & 0.2             & 0.0519         & 0.2146       \\ \hline
\textbf{ALOI}               & 0.032            & 0.0306         & 0.0376          & 0.0402        & 0.0534       & 0.328          & 0.199          & 0.178         & 0.37           & \textbf{0.558} & 0.0388          & 0.1114          & 0.1249          & 0.067           & 0.0404         & 0.0421       \\ \hline
\textbf{AnnThyroid}         & 0.0683           & 0.0547         & 0.0509          & 0.0509        & 0.0686       & 0.044          & 0.042          & 0.055         & 0.071          & \textbf{0.097} & 0.0485          & 0.0833          & 0.0803          & 0.097           & 0.0594         & 0.0529       \\ \hline
\textbf{hypothyroid}        & 0.0654           & 0.0601         & 0.0517          & 0.0517        & 0.065        & 0.233          & 0.287          & 0.211         & \textbf{0.325} & 0.322          & 0.0602          & 0.108           & 0.0826          & 0.063           & 0.0526         & 0.054        \\ \hline
\textbf{WDBC}               & 0.6123           & 0.0558         & 0.5197          & 0.4929        & 0.5943       & 0.1            & 0.116          & 0.117         & 0.086          & 0.096          & 0.4217          & 0.6302          & 0.4866          & \textbf{0.9375} & 0.3256         & 0.6492       \\ \hline
\textbf{Ionosphere}         & 0.078            & 0.1182         & 0.0697          & 0.0703        & 0.0707       & 0.233          & 0.236          & 0.119         & 0.171          & \textbf{0.275} & 0.0555          & 0.0614          & 0.0621          & 0.0757          & 0.0543         & 0.0612       \\ \hline
\textbf{MNIST}              & 0.3231           & 0.0983         & 0.092           & 0.092         & 0.274        & 0.049          & 0.045          & 0.051         & 0.08           & 0.114          & 0.297           & 0.3385          & 0.3925          & 0.2786          & \textbf{0.394} & 0.2597       \\ \hline
\textbf{Arrhythmia}         & 0.0956           & 0.0806         & 0.0509          & 0.0509        & 0.0682       & 0.307          & 0.26           & 0.31          & \textbf{0.326} & 0.23           & 0.0497          & 0.0611          & 0.0926          & 0.1153          & 0.051          & 0.0627       \\ \hline
\textbf{madelon}            & 0.0982           & 0.0909         & 0.0909          & 0.0909        & 0.1017       & 0.059          & 0.06           & 0.064         & 0.058          & 0.053          & 0.0967          & 0.1185          & 0.1235          & 0.1041          & \textbf{0.41}  & 0.101        \\ \hline
\textbf{isolet}             & 0.1059           & 0.0926         & 0.0798          & 0.0798        & 0.0796       & 0.037          & 0.038          & 0.041         & 0.043          & 0.036          & 0.1263          & \textbf{0.2327} & 0.2148          & 0.1104          & 0.08           & 0.0899       \\ \hline
\textbf{letter-recognition} & 0.115            & 0.087          & 0.0848          & 0.0848        & 0.0848       & 0.062          & 0.06           & 0.062         & 0.071          & 0.057          & 0.1117          & \textbf{0.3067} & 0.2428          & 0.1105          & 0.085          & 0.0952       \\ \hline
\textbf{InternetAds}        & 0.1739           & 0.0196         & 0.0457          & 0.0452        & 0.0213       & 0.062          & 0.068          & 0.3           & 0.138          & 0.066          & 0.1086          & 0.4373          & \textbf{0.4431} & 0.119           & 0.02           & 0.0506       \\ \hline
\end{tabular}
}
\end{table*}

\begin{table*}[!t]
\caption{AUC Scores of online and offline detectors per dataset}
\label{table:auc_scores}
\centering
\scalebox{0.55}{
\begin{tabular}{|l|c|c|c|c|c|c|c|c|c|c|c|c|c|c|c|c|}
\hline
\textbf{DATASETS}           & \textbf{RS-HASH} & \textbf{STARE} & \textbf{LEAP}   & \textbf{CPOD} & \textbf{X-S}   & \textbf{HST}   & \textbf{HSTF} & \textbf{RRCF} & \textbf{MCOD}  & \textbf{L-S}  & \textbf{IF}     & \textbf{LOF}    & \textbf{KNN}    & \textbf{X-B}   & \textbf{OCRF}   & \textbf{L-B} \\ \hline
\textbf{http}               & 0.98             & 0.6985         & \textbf{0.9979} & 0.99          & 0.983          & 0.996          & 0.993         & 0.995         & 0.997          & 0.4926        & 0.9898          & 0.3542          & 0.1075          & 0.99           & 0.5002          & 0.638        \\ \hline
\textbf{smtp}               & 0.79             & 0.8006         & 0.9004          & 0.9112        & \textbf{0.958} & 0.894          & 0.91          & 0.732         & 0.848          & 0.854         & 0.9569          & 0.9266          & 0.9137          & 0.8505         & 0.8981          & 0.8329       \\ \hline
\textbf{wilt}               & 0.5              & 0.6491         & 0.5011          & 0.5011        & 0.6769         & 0.417          & 0.393         & 0.516         & 0.616          & 0.671         & 0.4605          & \textbf{0.7308} & 0.6991          & 0.4909         & 0.4666          & 0.3793       \\ \hline
\textbf{adult}              & 0.5              & 0.6015         & 0.5003          & 0.5003        & 0.6364         & 0.572          & 0.523         & 0.535         & 0.574          & 0.4677        & 0.5936          & 0.609           & 0.6022          & 0.58           & \textbf{0.6892} & 0.5395       \\ \hline
\textbf{Diabetes}           & 0.5517           & 0.5609         & 0.5             & 0.5           & 0.5597         & 0.531          & 0.548         & 0.519         & 0.527          & 0.5459        & 0.5692          & \textbf{0.5821} & 0.5694          & 0.567          & 0.4871          & 0.5243       \\ \hline
\textbf{electricity}        & 0.5              & 0.5136         & \textbf{0.6236} & 0.6217        & 0.5818         & 0.611          & 0.484         & 0.573         & 0.619          & 0.4651        & 0.4642          & 0.4134          & 0.4248          & 0.463          & 0.448           & 0.5786       \\ \hline
\textbf{pima}               & 0.6247           & 0.5827         & 0.5             & 0.5           & 0.6981         & 0.534          & 0.596         & 0.535         & \textbf{0.764} & 0.61          & 0.6845          & 0.6528          & 0.6346          & 0.6658         & 0.6561          & 0.5957       \\ \hline
\textbf{breast-w}           & 0.7284           & 0.5922         & 0.747           & 0.741         & 0.8088         & 0.767          & 0.759         & 0.824         & 0.803          & 0.79          & 0.8317          & \textbf{0.8372} & 0.8269          & 0.833          & 0.7305          & 0.7326       \\ \hline
\textbf{forestcover}        & 0.5              & 0.7844         & 0.5             & 0.5           & 0.955          & \textbf{0.988} & 0.934         & 0.703         & 0.722          & 0.7928        & 0.9304          & 0.6222          & 0.8602          & 0.96           & 0.6175          & 0.8723       \\ \hline
\textbf{magic-telescope}    & 0.6438           & 0.5586         & 0.5001          & 0.5001        & 0.7494         & 0.629          & 0.661         & 0.616         & 0.693          & 0.6564        & \textbf{0.7883} & 0.7617          & 0.7726          & 0.7539         & 0.6907          & 0.6879       \\ \hline
\textbf{PageBlocks}         & 0.7153           & 0.6369         & 0.5019          & 0.5019        & 0.8524         & 0.785          & 0.832         & 0.695         & 0.57           & 0.678         & \textbf{0.9148} & 0.8674          & 0.5533          & 0.8308         & 0.8468          & 0.5084       \\ \hline
\textbf{pendigits}          & 0.5246           & 0.618          & 0.5             & 0.5           & 0.7347         & 0.517          & 0.518         & 0.725         & 0.689          & 0.587         & 0.7279          & \textbf{0.9152} & 0.8915          & 0.7036         & 0.5894          & 0.5386       \\ \hline
\textbf{Cardiotocography}   & \textbf{0.8359}  & 0.6044         & 0.5118          & 0.5118        & 0.8287         & 0.766          & 0.765         & 0.667         & 0.798          & 0.823         & 0.7936          & 0.7286          & 0.7583          & 0.815          & 0.5192          & 0.8089       \\ \hline
\textbf{ALOI}               & 0.5164           & 0.5012         & 0.5222          & 0.523         & 0.5516         & 0.535          & 0.533         & 0.518         & 0.532          & 0.5339        & 0.5406          & \textbf{0.7492} & 0.7023          & 0.543          & 0.5032          & 0.5264       \\ \hline
\textbf{AnnThyroid}         & 0.5444           & 0.5335         & 0.5077          & 0.5077        & 0.5587         & 0.529          & 0.531         & 0.547         & 0.572          & 0.5663        & 0.4713          & 0.5818          & \textbf{0.5912} & 0.577          & 0.5254          & 0.5078       \\ \hline
\textbf{hypothyroid}        & 0.5477           & 0.5228         & 0.5159          & 0.5157        & 0.5573         & 0.546          & 0.544         & 0.533         & 0.54           & 0.5211        & 0.5377          & \textbf{0.6144} & 0.6031          & 0.535          & 0.4988          & 0.5124       \\ \hline
\textbf{WDBC}               & 0.8961           & 0.5859         & 0.9136          & 0.8894        & 0.9294         & 0.997          & 0.981         & 0.701         & 0.907          & 0.962         & 0.8336          & 0.9322          & 0.9168          & \textbf{0.997} & 0.8722          & 0.9236       \\ \hline
\textbf{Ionosphere}         & 0.5787           & 0.5821         & 0.6009          & 0.6109        & 0.5665         & 0.38           & 0.34          & 0.492         & 0.622          & \textbf{0.71} & 0.4811          & 0.5394          & 0.542           & 0.5591         & 0.4926          & 0.5698       \\ \hline
\textbf{MNIST}              & 0.7551           & 0.5215         & 0.4998          & 0.4998        & 0.8054         & 0.827          & \textbf{0.84} & 0.667         & 0.5            & 0.7315        & 0.8177          & 0.8028          & 0.836           & 0.7979         & 0.84            & 0.7479       \\ \hline
\textbf{Arrhythmia}         & \textbf{0.6422}  & 0.5354         & 0.5             & 0.5           & 0.5429         & 0.49           & 0.553         & 0.483         & 0.501          & 0.5123        & 0.4675          & 0.5166          & 0.559           & 0.5993         & 0.5             & 0.5326       \\ \hline
\textbf{madelon}            & 0.5367           & 0.5            & 0.5             & 0.5           & 0.5396         & 0.508          & 0.524         & 0.508         & 0.502          & 0.5098        & 0.5001          & 0.5474          & \textbf{0.5669} & 0.5075         & 0.5             & 0.5281       \\ \hline
\textbf{isolet\_sampled}    & 0.5796           & 0.5338         & 0.501           & 0.501         & 0.5            & 0.527          & 0.469         & 0.533         & 0.48           & 0.5287        & 0.5942          & \textbf{0.7737} & 0.7074          & 0.5409         & 0.5486          & 0.5168       \\ \hline
\textbf{letter-recognition} & 0.573            & 0.5            & 0.5             & 0.5           & 0.5            & 0.541          & 0.548         & 0.507         & 0.503          & 0.5079        & 0.5777          & \textbf{0.8175} & 0.7425          & 0.5688         & 0.492           & 0.5416       \\ \hline
\textbf{InternetAds}        & 0.6013           & 0.5            & 0.6765          & 0.6701        & 0.5            & 0.649          & 0.689         & 0.492         & 0.659          & 0.684         & 0.584           & \textbf{0.9286} & 0.8255          & 0.53           & 0.5             & 0.5616       \\ \hline
\end{tabular}
}
\end{table*}

\begin{table}[!t]
\centering
\caption{Offline detectors' number of wins (using AP Scores) and average AP difference from the winner per dataset}
\label{table:offline_ap}
\begin{tabular}{|l|c|c|}
\hline
\textbf{Detector}& \multicolumn{1}{l|}{\textbf{\#Wins}} & \multicolumn{1}{l|}{\textbf{Avg. Difference from leader}} \\ \hline
\textbf{IF} & 3 & 13.5\%\\ \hline
\textbf{LOF} & 6 & 11.2\%\\ \hline
\textbf{KNN} & 3 & 11.2\%\\ \hline
\textbf{X-B} & 8 & 9.6\%\\ \hline
\textbf{OCRF}& 3 & 18.2\%\\ \hline
\textbf{L-B}& 1 & 18.9\%\\ \hline
\end{tabular}
\end{table}

\begin{table}[!t]
\centering
\caption{Offline detectors' number of wins (using AUC Scores) and average AUC ROC difference from the winner per datase}
\label{table:offline_auc}
\begin{tabular}{|l|c|c|}
\hline
\textbf{Detector}  & \multicolumn{1}{l|}{\textbf{\#Wins}} & \multicolumn{1}{l|}{\textbf{Avg. Difference from leader}} \\ \hline
\textbf{IF} & 2 & 13.1\%\\ \hline
\textbf{LOF} & 6 & 10.2\%\\ \hline
\textbf{KNN} & 3 & 12.7\%\\ \hline
\textbf{X-B} & 11 & 7.7\%\\ \hline
\textbf{OCRF}& 1 & 20.2\%\\ \hline
\textbf{L-B}& 1 & 19\%\\ \hline
\end{tabular}
\end{table}

\section{Meta-Features}
\label{app:meta}

\begin{table}[!t]
\caption{Statistically significant correlations between the ratio of X-S divided by the respective online detector and meta-features. We report only pairs that had a statistically significant correlation at a significance level of 0.05.}
\label{table:pairwise_correlations_online}
\centering
\scalebox{0.8}{
\begin{threeparttable}
\begin{tabular}{cc}
\hline
\multicolumn{1}{|c|}{\textbf{Meta-feature}} & \multicolumn{1}{c|}{\textbf{$\rho$ Ratio}} \\ \hline
\multicolumn{2}{c}{\textbf{RS-HASH}}                                              \\ \hline
\multicolumn{1}{|c|}{G1}                    & \multicolumn{1}{c|}{0.45*}          \\ \hline
\multicolumn{1}{|c|}{G2}                    & \multicolumn{1}{c|}{-0.62**}        \\ \hline
\multicolumn{1}{|c|}{S1}                    & \multicolumn{1}{c|}{-0.48*}         \\ \hline
\multicolumn{1}{|c|}{F3 (MEAN)}             & \multicolumn{1}{c|}{0.42*}          \\ \hline
\multicolumn{1}{|c|}{S4}                    & \multicolumn{1}{c|}{-0.52**}        \\ \hline
\multicolumn{1}{|c|}{F5 (MEAN)}             & \multicolumn{1}{c|}{0.49*}          \\ \hline
\multicolumn{1}{|c|}{F5 (SD)}               & \multicolumn{1}{c|}{0.45*}          \\ \hline
\multicolumn{1}{|c|}{F6 (MEAN)}             & \multicolumn{1}{c|}{0.44*}          \\ \hline
\multicolumn{1}{|c|}{F7 (MEAN)}             & \multicolumn{1}{c|}{0.58**}         \\ \hline
\multicolumn{1}{|c|}{F8 (MEAN)}             & \multicolumn{1}{c|}{0.47*}          \\ \hline
\multicolumn{1}{|c|}{F9 (MEAN)}             & \multicolumn{1}{c|}{0.44*}          \\ \hline
\multicolumn{1}{|c|}{F15}                   & \multicolumn{1}{c|}{-0.63**}        \\ \hline
\multicolumn{1}{|c|}{S3}                    & \multicolumn{1}{c|}{-0.52**}        \\ \hline
\multicolumn{1}{|c|}{S2}                    & \multicolumn{1}{c|}{-0.52**}        \\ \hline
\multicolumn{1}{|c|}{S5}                    & \multicolumn{1}{c|}{0.52**}         \\ \hline
\multicolumn{2}{c}{\textbf{STARE}}                                                \\ \hline
\multicolumn{1}{|c|}{G1}                    & \multicolumn{1}{c|}{0.52**}         \\ \hline
\multicolumn{1}{|c|}{G3}                    & \multicolumn{1}{c|}{-0.50*}         \\ \hline
\multicolumn{2}{c}{\textbf{LEAP}}                                                 \\ \hline
\multicolumn{1}{|c|}{F1 (MEAN)}             & \multicolumn{1}{c|}{0.69**}         \\ \hline
\multicolumn{1}{|c|}{F1 (SD)}               & \multicolumn{1}{c|}{0.65**}         \\ \hline
\multicolumn{1}{|c|}{F2 (MEAN)}             & \multicolumn{1}{c|}{0.60**}         \\ \hline
\multicolumn{1}{|c|}{F2 (SD)}               & \multicolumn{1}{c|}{0.59**}         \\ \hline
\multicolumn{1}{|c|}{FR1}                   & \multicolumn{1}{c|}{0.54**}         \\ \hline
\multicolumn{1}{|c|}{F3 (MEAN)}             & \multicolumn{1}{c|}{0.70**}         \\ \hline
\multicolumn{1}{|c|}{F3 (SD)}               & \multicolumn{1}{c|}{0.67**}         \\ \hline
\multicolumn{1}{|c|}{F5 (MEAN)}             & \multicolumn{1}{c|}{0.67**}         \\ \hline
\multicolumn{1}{|c|}{F5 (SD)}               & \multicolumn{1}{c|}{0.68**}         \\ \hline
\multicolumn{1}{|c|}{F6 (MEAN)}             & \multicolumn{1}{c|}{0.65**}         \\ \hline
\multicolumn{1}{|c|}{F6 (SD)}               & \multicolumn{1}{c|}{0.62**}         \\ \hline
\multicolumn{1}{|c|}{F7 (MEAN)}             & \multicolumn{1}{c|}{0.60**}         \\ \hline
\multicolumn{1}{|c|}{F7 (SD)}               & \multicolumn{1}{c|}{0.63**}         \\ \hline
\multicolumn{1}{|c|}{F8 (MEAN)}             & \multicolumn{1}{c|}{0.62**}         \\ \hline
\multicolumn{1}{|c|}{F8 (SD)}               & \multicolumn{1}{c|}{0.68**}         \\ \hline
\multicolumn{1}{|c|}{F10 (MEAN)}            & \multicolumn{1}{c|}{0.61**}         \\ \hline
\multicolumn{1}{|c|}{F10 (SD)}              & \multicolumn{1}{c|}{0.54**}         \\ \hline
\multicolumn{1}{|c|}{F11 (MEAN)}            & \multicolumn{1}{c|}{0.64**}         \\ \hline
\multicolumn{1}{|c|}{F11 (SD)}              & \multicolumn{1}{c|}{0.58**}         \\ \hline
\multicolumn{1}{|c|}{F14 (MEAN)}            & \multicolumn{1}{c|}{0.62**}         \\ \hline
\multicolumn{1}{|c|}{F14 (SD)}              & \multicolumn{1}{c|}{0.58**}         \\ \hline
\multicolumn{2}{c}{\textbf{CPOD}}                                                 \\ \hline
\multicolumn{1}{|c|}{F1 (MEAN)}             & \multicolumn{1}{c|}{0.69**}         \\ \hline
\multicolumn{1}{|c|}{F1 (SD)}               & \multicolumn{1}{c|}{0.65**}         \\ \hline
\multicolumn{1}{|c|}{F2 (MEAN)}             & \multicolumn{1}{c|}{0.60**}         \\ \hline
\multicolumn{1}{|c|}{F2 (SD)}               & \multicolumn{1}{c|}{0.59**}         \\ \hline
\multicolumn{1}{|c|}{FR1}                   & \multicolumn{1}{c|}{0.54**}         \\ \hline
\multicolumn{1}{|c|}{F3 (MEAN)}             & \multicolumn{1}{c|}{0.70**}         \\ \hline
\multicolumn{1}{|c|}{F3 (SD)}               & \multicolumn{1}{c|}{0.67**}         \\ \hline
\multicolumn{1}{|c|}{F5 (MEAN)}             & \multicolumn{1}{c|}{0.67**}         \\ \hline
\multicolumn{1}{|c|}{F5 (SD)}               & \multicolumn{1}{c|}{0.68**}         \\ \hline
\multicolumn{1}{|c|}{F6 (MEAN)}             & \multicolumn{1}{c|}{0.65**}         \\ \hline
\multicolumn{1}{|c|}{F6 (SD)}               & \multicolumn{1}{c|}{0.62**}         \\ \hline
\multicolumn{1}{|c|}{F7 (MEAN)}             & \multicolumn{1}{c|}{0.60**}         \\ \hline
\multicolumn{1}{|c|}{F7 (SD)}               & \multicolumn{1}{c|}{0.63**}         \\ \hline
\multicolumn{1}{|c|}{F8 (MEAN)}             & \multicolumn{1}{c|}{0.62**}         \\ \hline
\multicolumn{1}{|c|}{F8 (SD)}               & \multicolumn{1}{c|}{0.68**}         \\ \hline
\multicolumn{1}{|c|}{F10 (MEAN)}            & \multicolumn{1}{c|}{0.61**}         \\ \hline
\multicolumn{1}{|c|}{F10 (SD)}              & \multicolumn{1}{c|}{0.54**}         \\ \hline
\multicolumn{1}{|c|}{F11 (MEAN)}            & \multicolumn{1}{c|}{0.64**}         \\ \hline
\multicolumn{1}{|c|}{F11 (SD)}              & \multicolumn{1}{c|}{0.58**}         \\ \hline
\multicolumn{1}{|c|}{F14 (MEAN)}            & \multicolumn{1}{c|}{0.62**}         \\ \hline
\multicolumn{1}{|c|}{F14 (SD)}              & \multicolumn{1}{c|}{0.58**}         \\ \hline
\multicolumn{2}{c}{\textbf{MCOD}}                                                 \\ \hline
\multicolumn{1}{|c|}{F4 (MEAN)}             & \multicolumn{1}{c|}{-0.43*}         \\ \hline
\multicolumn{1}{|c|}{F4 (SD)}               & \multicolumn{1}{c|}{-0.45*}         \\ \hline
\multicolumn{1}{|c|}{F12 (SD)}              & \multicolumn{1}{c|}{-0.46*}         \\ \hline
\multicolumn{2}{c}{\textbf{L-S}}                                                  \\ \hline
\multicolumn{1}{|c|}{F12(SD)}               & \multicolumn{1}{c|}{-0.47*}         \\ \hline
\end{tabular}
       \begin{tablenotes}
            \item[*] For $pvalue \leq 0.05$
            \item[**] For $pvalue < 0.01$
            \item[***] For $pvalue < 0.001$
        \end{tablenotes}
     \end{threeparttable}
}
\end{table}

\end{document}